%% file: main.tex
\documentclass{naturep}
\usepackage[utf8]{inputenc}
\usepackage{lineno}
\usepackage{setspace}
\usepackage{color}
\usepackage{graphicx}
\usepackage{booktabs}
\usepackage{amssymb}
\usepackage{amsmath}
\usepackage{hyperref}
\usepackage{adjustbox}
\usepackage{subcaption}
\usepackage{tablefootnote}
\usepackage[margin=0.6in]{geometry}
\usepackage{verbatim}
\usepackage{caption}
\usepackage{paralist}
\usepackage{cleveref}
\usepackage{multirow}
\usepackage{tabularx} 
\usepackage{array} 
\usepackage{makecell}
\usepackage{xurl}

\usepackage{booktabs}
\usepackage[table]{xcolor}

\usepackage{xspace}
\newcommand{\ours}{\textsc{SEAL}\xspace}

\newcommand{\methodlong}{\textbf{S}patial \textbf{E}xpression-\textbf{A}ligned \textbf{L}earning\xspace}
\newcommand{\data}{\textsc{MAPLE}\xspace}
\newcommand{\datasetshort}{MAPLE-790k\xspace}
\newcommand{\datasetlong}{\textbf{M}orphology \textbf{A}ligned Dataset with \textbf{P}aired \textbf{L}ocal \textbf{E}xpression\xspace}

\newcommand{\github}{\url{https://github.com/mahmoodlab/SEAL/}\xspace}

\usepackage[all]{nowidow}

\newcommand\Heading[1]{
  \noindent\textbf{\Large{#1}}
}

\newcommand\heading[1]{
  \noindent\textbf{\large{#1}}
}

\newcommand\hheading[1]{
  \noindent\textbf{#1}
}

\newcommand{\vect}[1]{\mathbf{#1}}
\newcommand{\mat}[1]{\mathbf{#1}}

\newcommand{\ngenes}{G}
\newcommand{\nsamples}{N}
\newcommand{\embdim}{d_i}

\newcommand{\imgx}[1]{\vect{x}_{#1}^{(p)}}
\newcommand{\imgz}[1]{\vect{z}_{#1}^{(p)}}
\newcommand{\genex}[1]{\vect{x}_{#1}^{(g)}}
\newcommand{\genez}[1]{\vect{z}_{#1}^{(g)}}
\newcommand{\geney}[1]{\vect{\hat{y}}_{#1}^{(g)}}
\newcommand{\imgy}[1]{\vect{\hat{y}}_{#1}^{(p)}}

\newcommand{\genemat}{\mat{X}^{(g)}}

\newcommand{\imgenc}[1]{f(#1)_{\theta}}

\newcommand{\geneenc}[1]{g(#1)_{enc, \phi}}
\newcommand{\genedec}[1]{g(#1)_{dec, \phi}}
\newcommand{\downstream}[1]{t(#1)_{\gamma}}

\title{\raggedright{\textbf{Towards Spatial Transcriptomics-driven Pathology Foundation Models}}} 

\author{\raggedright{
Konstantin Hemker$^{\ast, 1,2}$, 
Andrew H. Song$^{\ast, 1,3,4}$, 
Cristina Almagro-Pérez$^{1,3,4,5}$, 
Guillaume Jaume$^{1,3,4}$, 
Sophia J. Wagner$^{1,3,4}$,
Anurag Vaidya$^{1,3,4,5}$,
Nikola Simidjievski$^{2,6}$,
Mateja Jamnik$^{2}$, and 
Faisal Mahmood$^{\dagger,1,3,4,7}$
}}
\date{}
    
\makeatletter
\let\saved@includegraphics\includegraphics
\AtBeginDocument{\let\includegraphics\saved@includegraphics}

\makeatother

\begin{document}
\maketitle
\begin{affiliations}
\item Department of Pathology, Mass General Brigham, Harvard Medical School, Boston, MA, USA
\item Department of Computer Science \& Technology, University of Cambridge, Cambridge, UK
 \item Cancer Program, Broad Institute of Harvard and MIT, Cambridge, MA, USA
 \item Data Science Program, Dana-Farber Cancer Institute, Boston, MA, USA
\item Harvard-MIT Division of Health Sciences and Technology, Massachusetts Institute of Technology, Cambridge, MA, USA
\item Télécom Paris, Institut Polytechnique de Paris, Paris, France
\item Harvard Data Science Initiative, Harvard University, Cambridge, MA, USA
\item[$^{\ast}$] Equal contribution\\
\textbf{$\dagger$ Corresponding author}: Faisal Mahmood (FaisalMahmood@bwh.harvard.edu)
\end{affiliations}

\clearpage
\Heading{Abstract}

\input{sections/0-abstract}


\clearpage
\begin{spacing}{1.35}

\Heading{Introduction}

\input{sections/1-introduction}

\Heading{Results}
\input{sections/2-results}

\Heading{Discussion}
\input{sections/3-discussion}

\clearpage

\Heading{Online Methods}

\input{supp/0-supp}
\input{supp/1-additional}

\setcounter{figure}{0}
\renewcommand{\figurename}{Extended Data Figure}
\input{supp/2-figures}

\clearpage
\begin{nolinenumbers}
\setcounter{table}{0}
\renewcommand{\tablename}{Extended Data Table}
\begin{spacing}{0.9}

\input{tables/01-data_hyperparameters}

\input{tables/02-maple_test_results}

\input{tables/03-hestbench_results}
\input{tables/04-data_efficiency}
\input{tables/05-slide_level_results}

\end{spacing}
\end{nolinenumbers}

\clearpage
\begin{nolinenumbers}
\Heading{References}
\bibliographystyle{nature}
\bibliography{main}
\end{nolinenumbers}

\newpage
\setcounter{figure}{0}
\renewcommand{\figurename}{\textbf{Extended Data Figure}}

\setcounter{table}{0}
\renewcommand{\tablename}{\textbf{Extended Data Table}}

\end{spacing}
\end{document}

%% file: sections/0-abstract.tex
\begin{spacing}{1.2}
\noindent
\textbf{Spatial transcriptomics (ST) provides spatially resolved measurements of gene expression, enabling characterization of the molecular landscape of human tissue beyond histological assessment as well as localized readouts that can be aligned with morphology. Concurrently, the success of multimodal foundation models that integrate vision with complementary modalities suggests that morphomolecular coupling between local expression and morphology can be systematically used to improve histological representations themselves. We introduce \methodlong (\ours), a vision–omics self-supervised learning framework that infuses localized molecular information into pathology vision encoders. Rather than training new encoders from scratch, \ours is designed as a parameter-efficient vision-omics finetuning method that can be flexibly applied to widely used pathology foundation models. We instantiate \ours by training on over 700,000 paired gene expression spot–tissue region examples spanning tumor and normal samples from 14 organs. Tested across 38 slide-level and 15 patch-level downstream tasks, \ours provides a drop-in replacement for pathology foundation models that consistently improves performance over widely used vision-only and ST prediction baselines on slide-level molecular status, pathway activity, and treatment response prediction, as well as patch-level gene expression prediction tasks. Additionally, \ours encoders exhibit robust domain generalization on out-of-distribution evaluations and enable new cross-modal capabilities such as gene-to-image retrieval. Our work proposes a general framework for ST-guided finetuning of pathology foundation models, showing that augmenting existing models with localized molecular supervision is an effective and practical step for improving visual representations and expanding their cross-modal utility. 
The code is available at \github.
}

\end{spacing}

%% file: sections/1-introduction.tex
Computational analysis of whole-slide images (WSIs) relies on models that can encode regions of interest (ROIs or ``patches'') or entire tissue slides into low-dimensional, informative embeddings\cite{song2023artificial, shmatko2022artificial, van2021deep, campanella2025clinical}. Recently, pathology foundation models (FMs), trained either on large histopathology image collections (\emph{vision-only})\cite{chen_Generalpurpose_2024UNI, wang2022transformer, zimmermann_Virchow2_2024Virchow2a, hoptimus0, filiot2024phikon, neidlinger2025benchmarking} or on paired histopathology images and captions (\emph{vision-language})\cite{huang2023visual, lu_Visuallanguage_2024CONCH, xiang_Vision_2025MUSK, wang_Pathology_2024CHIEF, ding_Multimodal_2024TITAN, shaikovski_PRISM_2024PRISM, xu_Wholeslide_2024GigaPath}, have improved performance for diagnostic and prognostication tasks. Concurrently, several studies have extended these frameworks to incorporate molecular signals, especially bulk RNA sequencing\cite{jaume2024transcriptomics, jaume2024multistain, vaidya_Moleculardriven_2025THREADS}. However, such bulk readouts lack the spatial resolution required to link fine-grained morphological patterns to their molecular correlates. As a result, current multimodal approaches underutilize spatially resolved morphology–molecular pairs that could support more nuanced tissue representations.

\emph{Spatial transcriptomics} (ST) directly addresses this limitation by providing spatially resolved gene expression profiles that can be registered to the underlying histomorphology\cite{staahl2016visualization, moses2022museum, marx2021method, tian2023expanding, jaume_HEST1k_2024HEST-1k}. 
While the earlier studies primarily remained within the ST domain, based on the analysis of spatially variable genes followed by qualitative correlation with underlying morphology\cite{staahl2016visualization, rao2021exploring, mo_Tumour_2024b}, recent works have focused on integrative analysis of both domains.
Specifically, the local integration of morphology and molecular state, enabled by the co-registration of two modalities, has been leveraged to produce image representations optimized to predict gene expression profiles from ROIs\cite{he_Integrating_2020HisToGene, xie_Spatially_2023BLEEP, lee_PathOmCLIP_2024PathOmCLIP, han_Unified_2024Umpire, chung_Accurate_2024Triplex, chen2025visual, zhang_Inferring_2024iStar, schroeder2025scaling, huang2025scalable, gindra2025large}. 
However, it has not been established how and to what extent we can build general-purpose ST-guided pathology FMs that improve representation quality beyond ST prediction tasks.

We hypothesize that cohesive vision–omics alignment, trained on diverse paired morphology image patch and ST datasets\cite{moses2022museum, jaume_HEST1k_2024HEST-1k, chen_STimage1K4M_2024STimage-1K4M}, can enhance image-based downstream clinical tasks, for which pathology FMs already play a crucial role, whilst further enabling new tasks such as cross-modal prediction or retrieval\cite{echle2021deep}. With a spectrum of pathology FMs already exhibiting strong downstream performance and the paired morphology-ST pretraining datasets being smaller than typical FM pretraining datasets due to sequencing costs, we identify two crucial requirements for successful vision-omics foundation models. 
First, the model needs to leverage the already strong morphological representations and finetune existing pathology FMs in a data-efficient manner. Second, the vision-omics alignment recipe needs to be generalizable enough to cater to a range of pathology FMs that harbor different architectures, training data distributions, and pretraining methods.

In this context, we introduce \methodlong (\ours), a self-supervised vision-omics learning framework for ST-guided pathology FMs, designed to augment prior learned morphological knowledge with localized molecular information, using a computationally efficient finetuning scheme. 
Concretely, \ours finetunes widely-used pathology FMs by integrating their patch embeddings with the corresponding Visium ST profiles. Our training objective combines a \emph{contrastive loss}, which aligns vision and omics embeddings in a shared space, with a \emph{reconstruction loss} that enables accurate prediction of transcriptomic profiles from image patches alone. To mitigate catastrophic forgetting of pretrained features\cite{ramasesh_Effect_2021catastrophic_forgetting}, \ours uses low-rank adaptation matrices on top of the original backbone weights, preserving morphological representations while integrating molecular signal. This yields general-purpose patch-level FMs with improved vision embeddings as gene-to-image and image-to-gene retrieval capabilities. We train \ours on MAPLE (Morphology Aligned Dataset with Paired Local Expression), a large-scale repository of over 720K paired histology patches with ST spots from diverse human tissue types. 

We demonstrate the effectiveness and generalizability of \ours finetuning recipe by deploying it on five pathology FMs: CONCH\cite{lu_Visuallanguage_2024CONCH}, H-Optimus-mini\cite{filiot_Distilling_2025hoptimus-mini}, Phikon-v2\cite{filiot2024phikon}, UNI-v2-h\cite{chen_Generalpurpose_2024UNI}, and Virchow-v2\cite{zimmermann_Virchow2_2024Virchow2a}. In extensive benchmarking across 38 slide-level and 15 patch-level downstream tasks, \ours vision-omics encoders consistently outperform their original non-finetuned counterparts on predicting across spatial scales such as slide-level molecular status, pathway activation, and morphological classes, as well as patch-level gene expression prediction.
Furthermore, the \mbox{\ours-finetuned} encoders show strong domain generalization on out-of-distribution expression prediction and gene-to-image retrieval, underscoring the benefits of augmenting existing FMs with complementary molecular information to enhance performance in spatial biology tasks.  

%% file: sections/2-results.tex
\begin{figure*}
\centering 
\includegraphics[width=\textwidth]{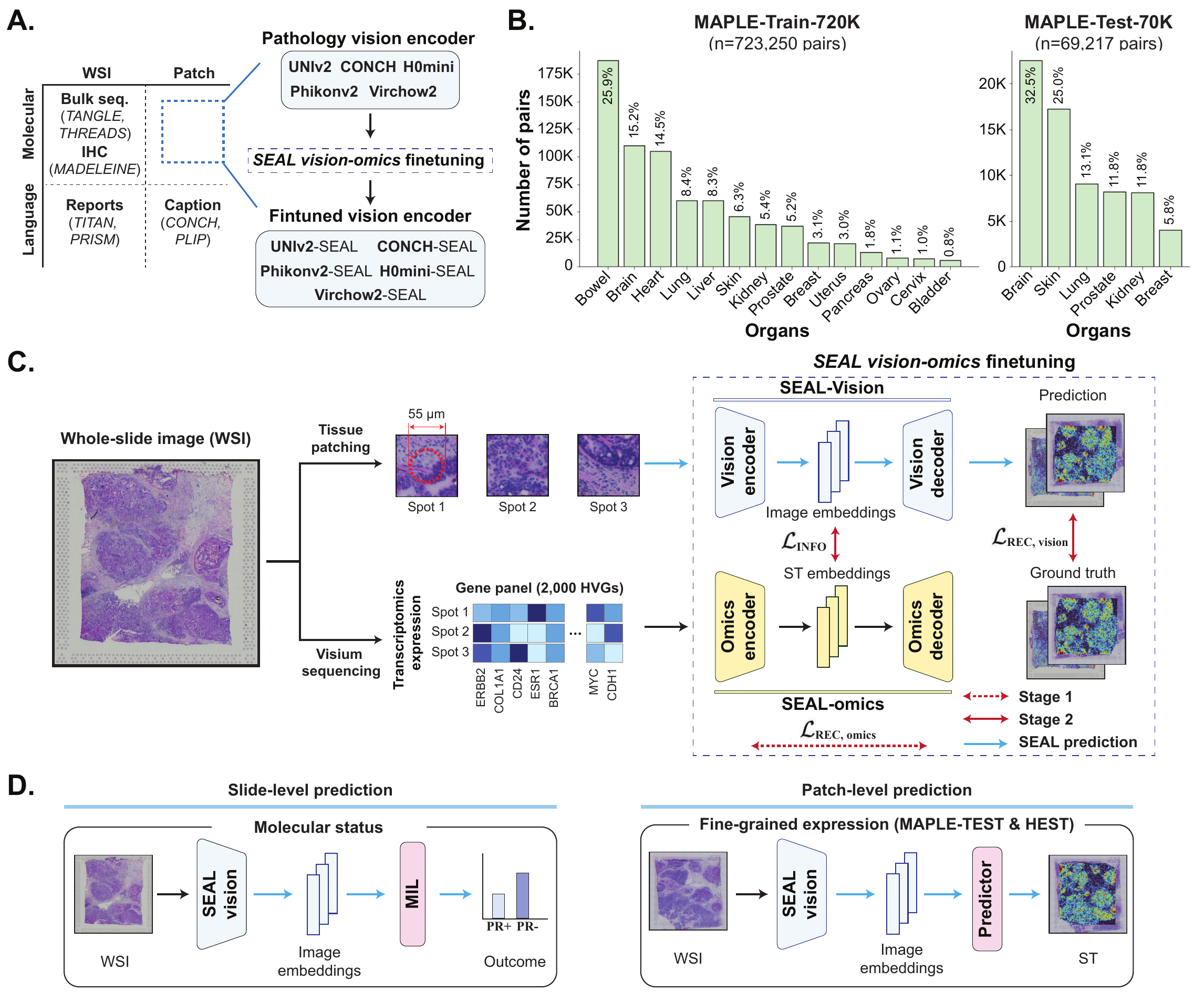}
\caption{\textbf{Overview of \ours.} \textbf{(A.)} Despite progress in multimodal foundation models for pathology, self-supervised learning using fine-grained vision–omics pretraining remains unexplored. \ours closes this gap by applying \textit{vision-omics} finetuning to pretrained patch encoders, yielding vision embeddings that better encode local molecular information.
\textbf{(B.)} Distribution of histology image patches and corresponding spatial transcriptomics (ST) expression profiles in \mbox{MAPLE-Train-720k} and \mbox{MAPLE-Test-70K}, used for training and evaluating \ours. \mbox{MAPLE} includes cancerous (42.8\%), diseased (19.4\%), healthy (32.2\%), and treated (5.6\%) samples.
\textbf{(C.)} Overview of \ours dual-pretraining strategy. First, an ST autoencoder is trained (\ours-omics) on an auxiliary transcriptomics reconstruction task ($\mathcal{L}_{\text{REC, omics}}$). Second, the vision encoder (\ours-vision), equipped with low-rank adapters, is trained to simultaneously align with the embedding space of \ours-omics ($\mathcal{L}_{\text{INFO}}$) and reconstruct the original gene panel ($\mathcal{L}_{\text{REC,vision}}$).
\textbf{(D.)} \ours-vision can be applied to a range of downstream tasks, such as molecular outcome prediction from whole-slide images and gene expression prediction. ST: Spatial Transcriptomics, HVG: Highly variable genes.}
\label{fig:main}
\end{figure*}

The main innovation of \ours~is to identify a molecular-guided finetuning recipe that is universally applicable to a range of existing Vision Transformer (ViT)-based\cite{dosovitskiy2020image} pathology FMs of different model sizes, pre-training methods, and data distributions (\textbf{Figure~\ref{fig:main}A}). This addresses the critical gap not yet filled by the different families of multimodal pathology FMs on a combination of different spatial scales (patches and WSIs) and complementary modalities (language and omics)\cite{huang2023visual, lu_Visuallanguage_2024CONCH, jaume2024transcriptomics, jaume2024multistain, xu_Wholeslide_2024GigaPath, ding_Multimodal_2024TITAN, shaikovski_PRISM_2024PRISM, vaidya_Moleculardriven_2025THREADS}. 

The utilization of existing pathology foundation models for \ours~is driven by the challenges present in curating large-scale training data of image patches and corresponding ST. Despite the concurrent efforts in aggregating and standardizing the publicly available slides and ST pairs\cite{jaume_HEST1k_2024HEST-1k, chen_STimage1K4M_2024STimage-1K4M, moses2022museum}, the image patches and ST pairs within each slide are relatively homogeneous due to non-negligible spatial autocorrelation\cite{yan2025categorization}, and thus cannot be treated as independent and identically distributed pairs as commonly done in vision-language datasets\cite{radford2021learning}. Furthermore, with the paired pretraining dataset size not on the scale typically required for training the vision foundation models, the initialization of the vision encoder with weights from the existing pathology foundation models presents an appealing solution to data scarcity and spatial homogeneity.
This approach is further corroborated by recent multimodal pathology foundation models\cite{lu_Visuallanguage_2024CONCH, ding_Multimodal_2024TITAN, xu_Wholeslide_2024GigaPath} demonstrating the success of the two-step approach of first pretraining the vision encoder only with histomorphology images and then subsequently finetuning with a complementary modality guidance.

We train and evaluate \ours, on \data (\datasetlong), a data repository of paired H\&E morphology image patches ($224 \times 224$ pixels) and corresponding Visium ST spots, each of which aggregates transcripts from multiple cells within 55$\mu m$ diameter spots, comprised of HEST\cite{jaume_HEST1k_2024HEST-1k}, public data\cite{mo_Tumour_2024b}, and internal data at Mass General Brigham (MGB) (\textbf{Figure~\ref{fig:main}B, Extended Data~\Cref{tbl:data_overview_organ,tbl:data_overview_disease_state,tbl:train_data_sources}}). The dataset is further divided into \mbox{\data-Train-720K} (\mbox{14 organs}) for model development and \mbox{\data-Test-70K} (\mbox{six organs}) for independent evaluation, stratified by patient cohort.

\ours training proceeds in two stages (\textbf{Figure~\ref{fig:main}C}). In the first stage, we train a transcriptomic encoder–decoder using a variational autoencoder (VAE), with a scale-invariant loss that mitigates the effects of high sparsity and heterogeneous variance in gene expression profiles. This transcriptomics-only pretraining produces a stable molecular latent space that is amenable to subsequent alignment with tissue morphology\cite{han_Unified_2024Umpire}. In the second stage, we align the \textit{transcriptomics} and \textit{histomorphology} embeddings produced by the transcriptomic encoder and the pretrained vision encoder, respectively, using a contrastive loss that maximizes the joint similarity of matched pairs. We couple this with an image-to-gene reconstruction loss that enforces the prediction of transcriptomic profiles from image patches. To reuse pretrained weights and limit catastrophic forgetting, we apply low-rank adaptation (LoRA)\cite{hu2022lora} to a subset of the vision encoder. The \ours-finetuned vision encoders can then be applied to downstream tasks (\textbf{Figure~\ref{fig:main}D}). Additional information on all training stages is provided in the \textbf{Online Methods, SEAL Pipeline} section. 

Previous work on morphomolecular alignment with ST has largely focused on predicting gene expression from morphological image patches, with limited assessment of whether these models improve clinically relevant downstream tasks such as slide-level outcome prediction. These approaches typically follow two directions. First, the ST predictors are fitted on patch embeddings with a reconstruction objective\cite{he_Integrating_2020STNet, he_Integrating_2020HisToGene, zeng2022spatial_Hist2ST, lee_PathOmCLIP_2024PathOmCLIP, wang_FmH2ST_2025FmH2ST, nonchev_DeepSpot_2025DeepSpot}. Second, retrieval-based methods\cite{xie_Spatially_2023BLEEP, han_Unified_2024Umpire, min_Multimodal_2024mclSTExt, lin2024STAlign,  chen2025visual, wang_HECLIP_2025HECLIP, gindra2025large} employ contrastive loss to align between morphology and ST embedding space by finetuning either the vision or gene encoder. The retrieval-based ST prediction amounts to averaging the closest ST embeddings in the reference database to the query morphology embedding. We hypothesize that either of these approaches is limited in its ability to generalize to clinically-relevant and out-of-distribution tasks, which are predominantly at the slide-level, as we further show in ablation experiments. 
In contrast, \ours employs a multi-objective loss function, which provides a direct signal that guides the vision encoder to learn features that are not just aligned (via $\mathcal{L}_{\text{INFO}}$), but also predictive of gene expression (via $\mathcal{L}_{\text{REC, vision}}$), resulting in aligned manifolds that can be leveraged for slide-level tasks.

\begin{figure*}
    \centering
    \includegraphics[width=\textwidth]{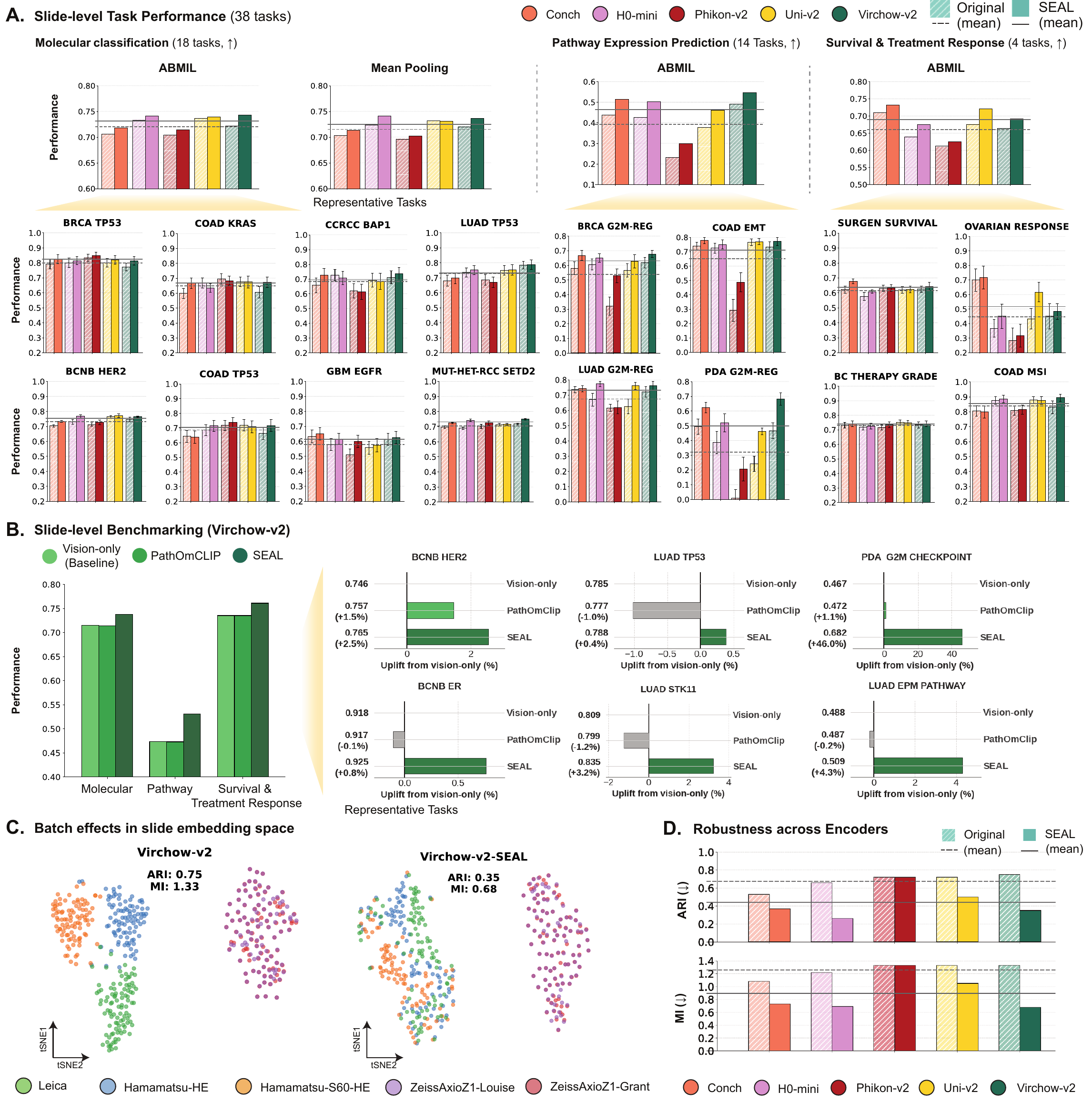}
    \caption{\textbf{Overview of \ours slide-level performance.}
     \textbf{(A.)} Evaluation of five foundation models with and without SEAL on slide-level evaluations across 38 tasks covering molecular classification, pathway expression prediction, and marker and response prediction. The patch features from the baseline encoders and \ours encoders are aggregated using attention-based multiple instance learning (ABMIL).
     \textbf{(B)}. Slide-level performance comparison between Virchow-v2 finetuned with the \ours or PathOmCLIP recipes, along with non-finetuned Virchow-v2 (Baseline) on the full set of tasks with ABMIL aggregation: molecular status prediction, pathway expression, and survival \& treatment response prediction tasks.
     \textbf{(C)}. t-SNE visualization comparing the robustness of Virchow-v2 and \ours counterpart to the batch effects. Each slide was scanned with five different scanners, resulting in subtle differences in color, contrast, and resolution. Each dot represents the mean-pooled slide embedding for 100 identical slides captured using five commonly used WSI scanners. Stronger clusters indicate that the encoder is more prone to batch effects.
     \textbf{(D)}. ARI and MI scores across five baseline encoders and \ours counterparts, using the same setup as (C). Lower ARI and MI correspond to weaker cluster formation. ARI: Adjusted Random Index, MI: Mutual Information.
    }
    \label{fig:slide_performance}
\end{figure*}

\heading{\ours~improves slide-level downstream task performance}

We first investigate whether \ours models improve slide-level classification tasks over baseline pathology FMs. Specifically, we evaluate \ours models on molecular status prediction (18 tasks), pathway expression (14 tasks), and marker \& treatment response prediction tasks (4 tasks).
The task categories were selected based on our hypothesis that molecular finetuning of the vision encoders would be beneficial for molecular outcome prediction or clinical tasks where complementary molecular information is critical.

We construct a slide-level embedding that serves as input to the slide-level predictor by aggregating (or pooling) the patch features extracted with the original or \ours vision encoders. We use two widely-used approaches: attention-based pooling\cite{ilse2018attention, lu2021data, campanella2025clinical} and mean pooling\cite{zaheer2017deep} to ensure \ours is robust to different aggregation strategies.
As for the vision encoders, we study five widely-used FMs for pathology, namely, CONCH\cite{lu_Visuallanguage_2024CONCH}, H0-mini\cite{filiot_Distilling_2025hoptimus-mini}, Phikon-v2\cite{filiot2024phikon}, UNI-v2\cite{chen_Generalpurpose_2024UNI}, and Virchow-v2\cite{zimmermann_Virchow2_2024Virchow2a}. This choice covers multiple ViT sizes from \textit{ViT-Base} (86M parameters, H0-mini \& CONCH), to \textit{ViT-Large} (307M parameters, Phikon-v2) and \textit{ViT-Huge} (632M parameters, UNI-v2 and Virchow-v2), as well as multiple  pretraining recipes: \textit{vision-language alignment} in CONCH, \textit{distillation} in H0-mini, and \textit{DINOv2} in Phikon-v2, UNI-v2, Virchow-v2. Each encoder is finetuned with the same \ours recipe with minor encoder-specific adaptations (\textbf{Extended Data Table~\ref{tbl:seal_hyperparameters}}).

We observe that all five pathology FMs benefit from \ours (\textbf{Figure~\ref{fig:slide_performance}A}). In molecular subtyping for gene mutation prediction tasks, we observe that, on average across encoders and all 18 tasks, the \ours~encoders achieve a 1.5\% improvement compared to the vision-only models (\textbf{Extended Data Figure~\ref{edf:molecular_performance}, \ref{edf:pathway_other_performance}}, \textbf{Extended Data~\Cref{tab:bc_therapy_er_status_results}-\ref{tab:mut-het-rcc_SETD2_mutation_results}}), with Virchow-v2-\ours seeing the largest relative improvement (3.1\%) and UNI-v2-\ours seeing the smallest (0.6\%) across molecular tasks. Notably, we also observe a minor improvement in performance on two immunohistochemistry (IHC) status prediction tasks (ER and PR status) on additional experiments run for Uni-v2 and Virchow-v2 (+0.3\% and +1.0\% respectively, \textbf{Extended Data Figure~\ref{edf:morphological_tasks}}). 
The same trends are observed with mean pooling, where we observe a 1.3\% increase on average (\textbf{Extended Data Figure~\ref{edf:molecular_performance}B}).
We also compare Virchow-v2-\ours with Virchow-v2 finetuned with PathOmCLIP recipe\cite{lee_PathOmCLIP_2024PathOmCLIP}, which finetunes the vision encoder with ST to enhance patch-level gene expression performance (\textbf{Figure~\ref{fig:slide_performance}B}). 
The superior performance of Virchow-v2-\ours on the slide-level tasks suggests that even with ST guidance, the finetuning recipes need to be carefully optimized to ensure improved performance across spatial scales.
In summary, the consistent slide-level improvement, regardless of the aggregation approaches, underscores the benefit of integrating localized molecular signals for superior slide-level representation quality.

We further evaluate \ours on morphological subtyping tasks to understand whether molecular finetuning can enhance performance for tasks where clinical outcomes are defined from morphological criteria. Specifically, we identify a set of challenging grading and subtyping tasks to compare the Virchow-v2 and UNI-v2 encoders and their \mbox{\ours-finetuned} counterparts. We observe that \ours mostly improves on slide-level morphological tasks as well, albeit by a smaller margin than the molecular slide-level tasks (\textbf{Extended Data Figure~\ref{edf:morphological_tasks}}).
We conjecture that the vision-only DINOv2 pretraining on more than 100 million patches, employed by both Virchow-v2 and UNI-v2 (\textbf{Extended Data Table~\ref{tbl:vision_encoder_overview}}), already yields strong morphological representations leading to saturated performance on morphology-oriented tasks, thus offering diminished benefit when finetuned with the 700k morphology-ST pairs. 

Beyond predictive performance on slide-level endpoints, we investigate the robustness of \ours representations to batch effects arising from variations in data acquisition, a persistent challenge in computational pathology\cite{tellez2019quantifying,stacke2020measuring,howard2021impact,ghaffari2022adversarial}. To this end, we use the ICAIRD dataset~\cite{InHwaUm2024icaird} comprising 100 renal cell carcinoma slides digitized with five distinct scanners, which introduces non-biological domain shifts in color, contrast, and resolution (\textbf{Extended Data \Cref{edf:robustness}A}). Visualizing the mean-pooled slide embeddings across different scanners reveals that the baseline FMs form more distinct clusters compared to \ours embeddings from the same set of slides (\textbf{\Cref{fig:slide_performance}C}). This suggests that \ours is able to further suppress the visual artifacts and thereby batch effects with the multi-modal finetuning. This is in line with the previous findings on the superior robustness of the multi-modal pathology FMs over uni-modal pathology FMs\cite{komen2025towards}.
Additionally, we measure the clustering tendency by scanner domain using the Adjusted Rand Index (ARI) and Mutual Information (MI), where a lower score indicates weaker cluster formation and would indicate higher robustness. \ours achieves a significantly better ARI and MI for four encoders, except for Phikon-v2. On average, the ARI score drops from 0.67 to 0.44 and the MI from 1.25 to 0.9, indicating that \ours embeddings are more robust to scanning artifacts (\textbf{\Cref{fig:slide_performance}D, Extended Data \Cref{edf:robustness}B}).

\begin{figure*}
\centering
\includegraphics[width=\textwidth]{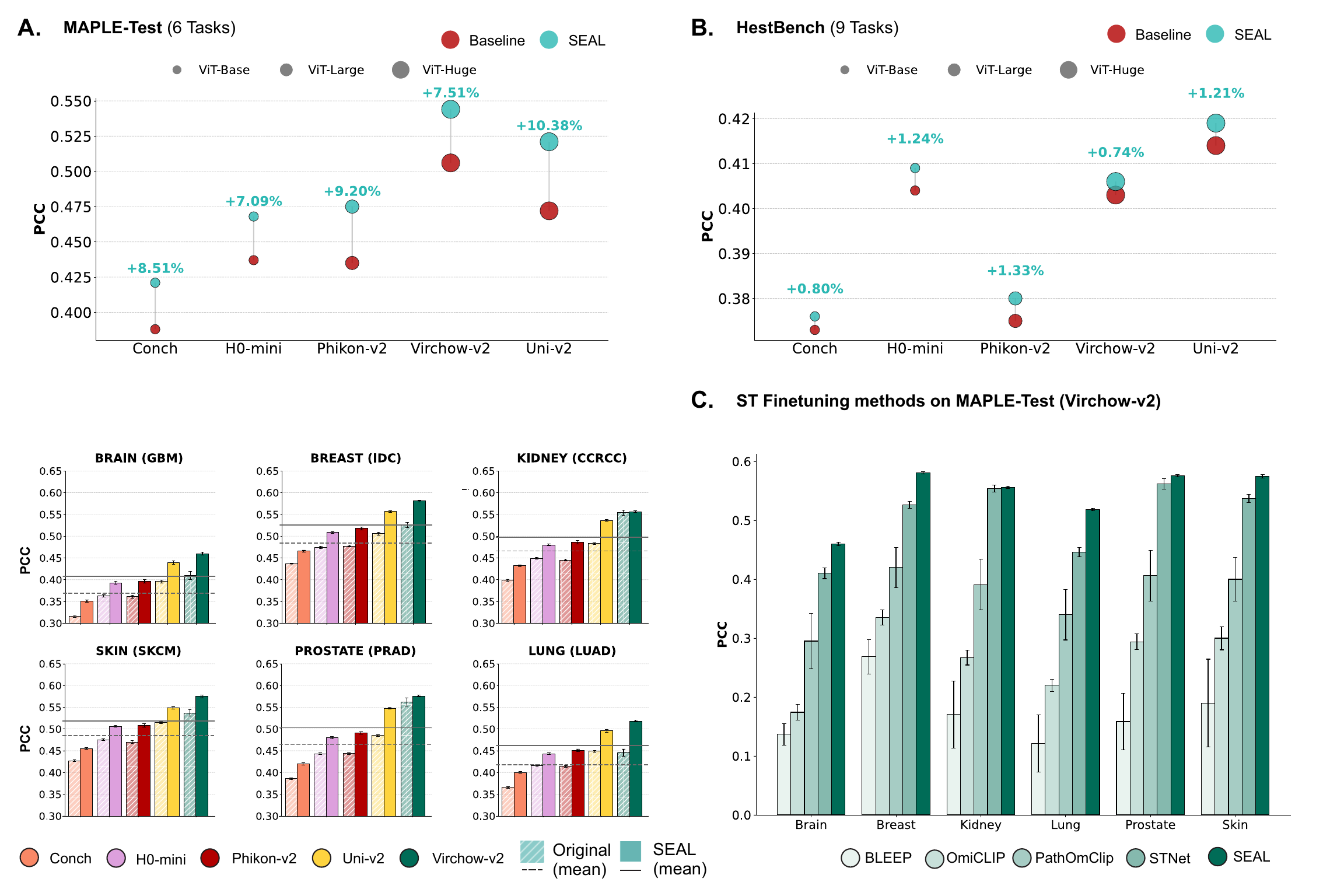}
\caption{\textbf{Overview of \ours patch-level task performance.}
\textbf{(A.)} (top) Average patch-based ST prediction performance on $\data$-Test dataset (6 organs) across five foundation models for pathology (Conch, H0-mini, Phikon-v2, Virchow-v2, Uni-v2), evaluated using Pearson correlation coefficient (PCC). The circle size indicates the relative number of trainable parameters in each model, from ViT-Base (86M parameters) to ViT-Huge (680M).
(bottom) Organ-level predictive performance of five foundation models and their SEAL variants on the $\data$-Test dataset (six organs).
\textbf{(B.)} Average performance of five pathology foundation models on HEST-Bench (9 histology-to–gene expression tasks) and corresponding absolute improvements obtained with \ours relative to the original models.
\textbf{(C.)} Performance comparison between Virchow-v2 SEAL encoder and other widely-used histology-to-ST prediction baselines on \data-Test. All baselines were also trained using the Virchow-v2 encoder, except OmiCLIP, which uses its corresponding encoder.   
}
\label{fig:patch_performance}
\end{figure*} 

\heading{\ours~improves patch-level gene expression prediction performance}

Encouraged by the performance uplift with \ours on slide-level tasks, we next assess the quality of patch-level representations, which directly determines the quality of slide-level representations\cite{ding_Multimodal_2024TITAN, vaidya_Moleculardriven_2025THREADS, campanella2025clinical, neidlinger2025benchmarking}. Specifically, we focus on patch-level cross-modal gene expression prediction as a direct measure of morphomolecular representation quality, similar to the slide-level molecular tasks in previous analyses.

\hheading{Comparison with vision-only models.}
\ours-finetuned encoder is evaluated on two benchmarks for predicting highly variable genes (HVGs) from image patch embeddings, namely \data-Test-70k (six organs) and HESTBench\cite{jaume_HEST1k_2024HEST-1k} (nine organs). Specifically, we fit a linear model on the frozen patch embedding (i.e., linear probe) to predict the expression levels of 10 HVGs (\data-Test) and 50 HVGs for each organ (HEST-Bench). We use the Pearson correlation coefficient (PCC) to compare the true and predicted expression. Additional information on the linear probing can be found in the \textbf{Online Methods, Linear Probing} section.

When evaluated on \data-Test, we observe a consistent and significant performance increase of \ours-finetuned across all vision encoders in the six test organs: Brain, Breast, Kidney, Lung, Prostate, and Skin (Wilcoxon signed-rank test, p$<$0.001, \textbf{Figure \ref{fig:patch_performance}A}). Across the five encoders, Virchow-v2 showed the largest average performance gain (+17.3\%), whereas H0-mini, the least improved model, still gained +7.3\% on average (\textbf{Extended Data~\Cref{tab:prostate_results,tab:brain_results,tab:breast_results,tab:kidney_results,tab:skin_results,tab:lung_results}}). The same trend is observed when evaluated with mean squared error (MSE), where \ours-finetuned encoders consistently lead to lower MSE (lower is better), with Virchow-v2 benefiting the most (11\% lower MSE). Additional evaluation on the HESTBench further highlights the benefit of \ours-finetuning (\textbf{Figure~\ref{fig:patch_performance}B, Extended Data~\Cref{tbl:hestbench}}), where all five encoders improve overall performance (evaluated using PCC), with Phikon-v2 improving the most with the average of 1.33\%. We attribute the lower performance gain on HESTBench to the significant distributional shift of this test benchmark compared to the \ours data. While \ours-Test assesses in-distribution generalization to holdout patients from the same source cohorts, HESTBench is composed of entirely separate study cohorts and ST technology (Xenium), thus representing a more challenging out-of-distribution task. Notably, the results also show improvements on some of the higher resolution Xenium tasks in this benchmark (IDC, PAAD, SKCM, COAD, LUAD) despite the absence of Xenium training data during finetuning, indicating \ours's potential to generalize beyond Visium tasks. These results collectively demonstrate that even powerful pathology vision encoders can be further enhanced via simple finetuning with molecular guidance.

\hheading{Comparison with ST prediction models.}
We further compare the predictive performance of \ours against widely-used patch-level ST prediction methods (\textbf{Figure~\ref{fig:patch_performance}C}). Specifically, in addition to linear probing evaluation on embeddings from both \ours-vision encoders and frozen original vision encoders (i.e., STNet\cite{he_Integrating_2020STNet}), we implement BLEEP\cite{xie_Spatially_2023BLEEP}, OmiCLIP\cite{chen2025visual}, and PathOmCLIP\cite{lee_PathOmCLIP_2024PathOmCLIP}. For a fair comparison, we use Virchow-v2 as the vision encoder for all frameworks, except for OmiCLIP for which we use the original encoder. When evaluated on \data-Test, we observe that the \ours vision encoders lead to the highest performance, followed by linear probe on the original vision encoders (STNet). Interestingly, prediction-based approaches (\ours, STNet, and PathOMCLIP) consistently provide better performance than the retrieval-based approaches (OmiCLIP and BLEEP). 
This implies that explicitly learning a predictive mapping of morphology to the gene expression is more generalizable and expressive than retrieving nearest ``neighbors'' from a reference dataset. The large gap is particularly surprising since some samples in the reference database used for retrieval are sourced from the same study cohorts as those in \data-Train. 
The strong performance of STNet, a simple linear probe on the frozen pathology FM embeddings, underscores that fitting simple models on top of large and extensively trained vision encoders can outperform more complex finetuning frameworks.
The lower performance of PathOmCLIP can be attributed to model overfitting induced by the reconstruction task, which may overwrite meaningful morphological features. This trend is consistent with our observations on slide-level prediction tasks for PathOmCLIP (\textbf{Figure~\ref{fig:slide_performance}B}).
Meanwhile, the \ours training paradigm outperforms all the ST finetuning baselines across all six test organs with an average PCC improvement of 2.4\%, with expression prediction on kidney genes leading to the smallest (0.4\%) and lung leading to the largest (16.1\%) improvement compared to the next-best model. Additional information on the linear probing can be found in the \textbf{Online Methods, Stage III - Evaluation} section.

\heading{\ours~design considerations}

We next investigate how different components of \ours pretraining shape downstream performance, comparing the effect of multiple pretraining stages, finetuning strategy, sample efficiency, and reconstruction objective. The ablation analyses are performed on \data-Test (patch-level) and BCNB cohort (slide-level) with Virchow-v2-\ours and UNI-v2-\ours.

\begin{figure*}
\centering
\includegraphics[width=\textwidth]{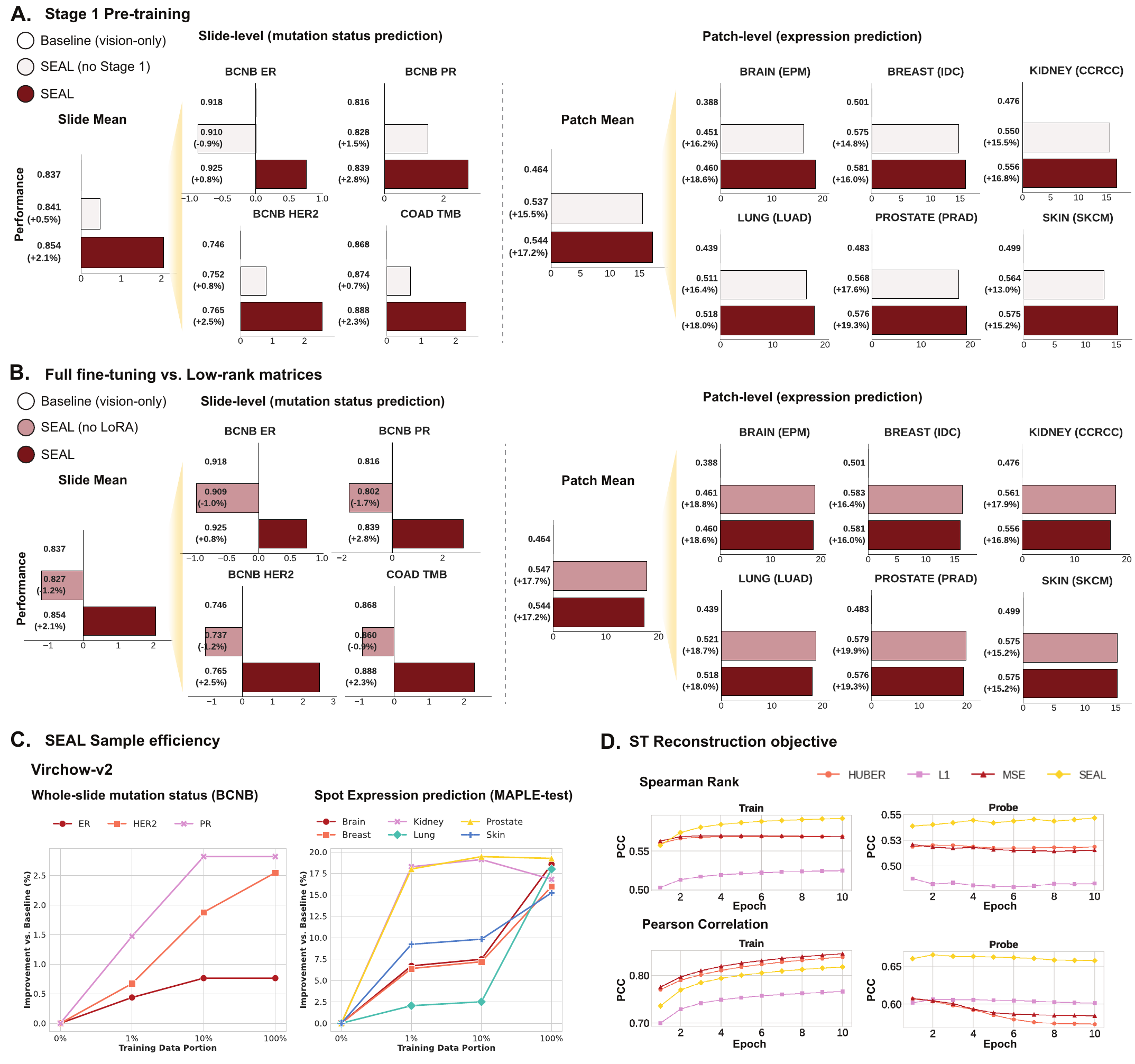}
\caption{\textbf{Investigation of $\ours$ design considerations.} 
\textbf{(A.)} Effect of Stage I pretraining (joint pretraining of transcriptomics encoder and decoder) in Virchow-v2 with $\ours$, showing a representative set of slide- and patch-level tasks.
\textbf{(B.)} Effect of conducting full model finetuning compared to learning low-rank matrices (LoRA) in \ours.
\textbf{(C.)} Effect of data size on \data-Test and the BCNB dataset. Different percentage of \data-Train (0, 1, 10, and 100\%) was used for \ours-finetuning. The y-axis shows the percentage improvement relative to the probe of the vision-only Virchow-v2 embeddings.
\textbf{(D.)} Effect of replacing \ours reconstruction objective ($\mathcal{L}_{\text{REC, vision}}$) by a HUBER, L1, or MSE objective. Performance evaluated using Spearman rank and Pearson correlation coefficient.
}
\label{edf:data_efficiency_pretraining}
\end{figure*}

First, we assess \ours without Stage 1 pretraining to isolate the contribution of first learning a structured transcriptomic embedding space before aligning it with the vision embeddings. We observe that omitting Stage 1 results in an average performance decrease of 1.48\% at the patch level and 1.57\% at the slide level (\textbf{Figure~\ref{edf:data_efficiency_pretraining}A}). This indicates that learning a structured embedding space for each modality prior to cross-modal alignment facilitates more effective coupling, highlighting the importance of Stage 1 in \ours. These results are consistent with reports in vision–language models that emphasize the value of structured unimodal embeddings before alignment\cite{lu_Visuallanguage_2024CONCH, ding_Multimodal_2024TITAN}. 

We further investigate the impact of the vision encoder finetuning recipe, replacing the parameter-efficient LoRA finetuning by full model finetuning (i.e., finetuning every parameter of the vision encoder). Our analysis shows the trade-off between optimization for the training task (patch-level prediction) and the preservation of general-purpose features for downstream slide-level tasks (\textbf{Figure~\ref{edf:data_efficiency_pretraining}B}). While full finetuning leads to marginally higher patch-level expression prediction performance (+0.5\%), it also incurs a notable loss in the predictive performance for slide-level tasks (-1.2\%). In contrast, the LoRA-based \ours increases the slide-level task performance while achieving patch-level performance comparable with full finetuning. This demonstrates that \ours can balance efficient adaptation of molecular information and preservation of the pretrained morphological features. 

We additionally explore varying the portion of the \data available for \ours-finetuning. Specifically, we sample 0, 1, 10, and 100\% of \data-Train, ensuring equal organ distribution as the original, with 0\% and 100\% representing non-finetuned and \ours-finetuned vision encoders. Based on the Virchow-v2 encoder, we observe the increasing performance trend as more ST data is available at both the patch-level and the slide-level tasks (\textbf{Figure~\ref{edf:data_efficiency_pretraining}C, \Cref{tbl:patch_Virchow-v2_data_efficiency,tbl:slide_Virchow-v2_data_efficiency}}), with the same trend observed for UNI-v2 encoder (\textbf{\Cref{tbl:patch_Uni-v2_data_efficiency,tbl:slide_Uni-v2_data_efficiency}}). 
This indicates the importance of having a large pretraining dataset, as previously established in foundation modeling literature\cite{chen_Generalpurpose_2024UNI, ding_Multimodal_2024TITAN}. 

We next conduct an ablation study to validate the design of the gene reconstruction objective introduced in \ours. A primary challenge of ST data is its high sparsity and varying distribution across genes\cite{bintayyash2021non, danaher2022advances}, which may persist even after normalization and preprocessing. Consequently, common regression objectives, such as Mean Squared Error (MSE), L1, or Huber loss, become biased towards highly variable genes and may neglect genes with lower variance. The \ours reconstruction objective introduces a scale-invariant loss term to address these challenges (\textbf{Online Methods, Stage I -- Self-Supervised ST representations}). To validate this choice, we compared the performance of identical models trained with our objective against those trained using only Huber, L1, or MSE (\textbf{Figure~\ref{edf:data_efficiency_pretraining}D}). The \ours objective achieves a higher and more stable Spearman rank and Pearson correlation on both validation sets throughout pre-training.

\begin{figure*}
\centering 
\includegraphics[width=\textwidth]{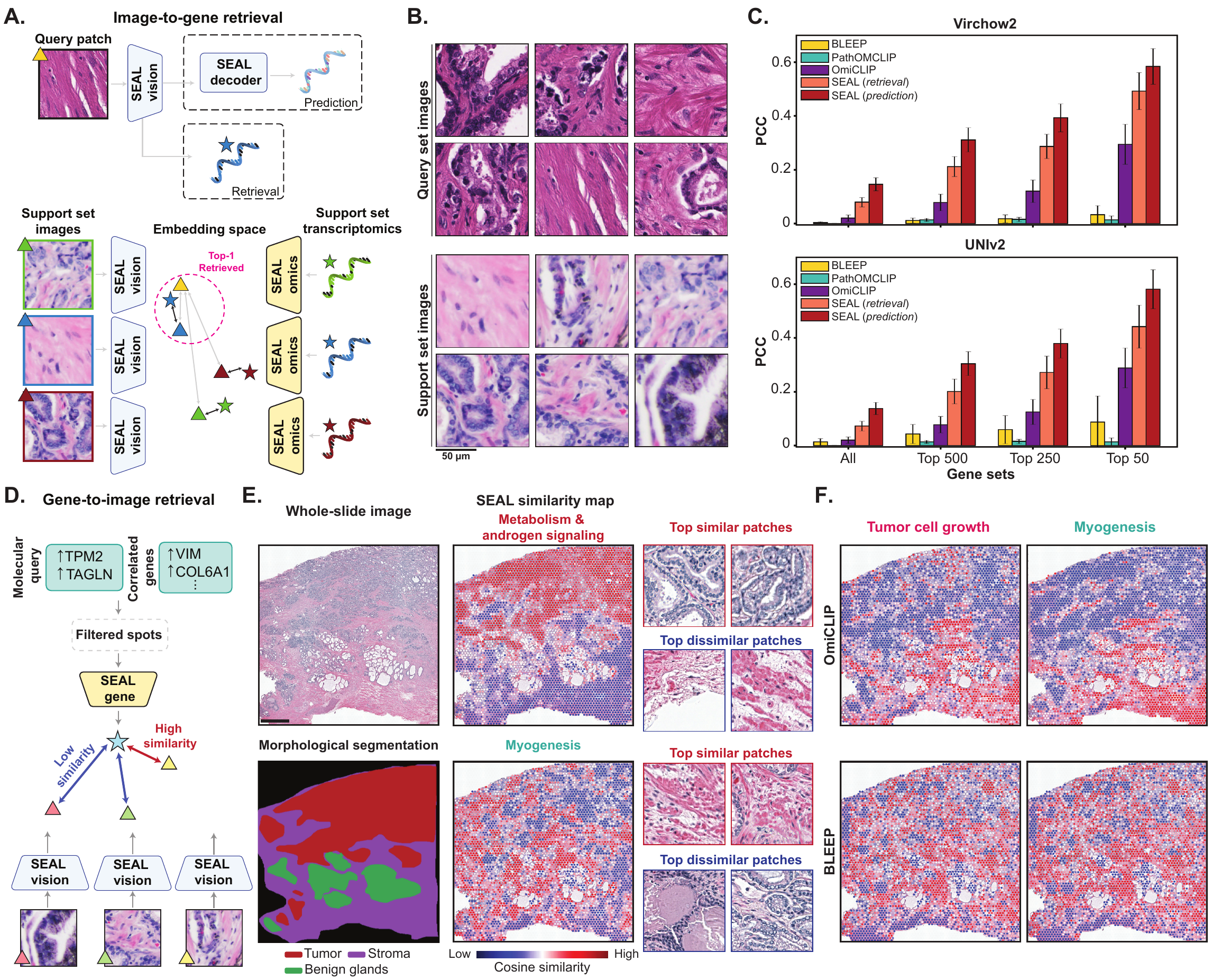}
\caption{\textbf{Cross-modal retrieval with $\ours$.} $\ours$ can perform cross-modal retrieval (image-to-gene, gene-to-image) leveraging the aligned vision and gene expression embedding space. 
\textbf{(A.)} Schematic overview of image-to-gene retrieval with $\ours$. Given a query image, we retrieve the $K$-nearest images from the support (training) set and average their corresponding transcriptomic embedding to become the predicted expression. For comparison, the direct image-to-gene predictive performance using the \ours decoder is also shown. 
\textbf{(B.)} Examples of the training (support) and test (query) set from internal prostate cancer samples. The data was split based on two separate cohorts that exhibit clear visual differences, representing an out-of-distribution prediction task.
\textbf{(C.)} Retrieval performance of $\ours$ and baselines evaluated with PCC. All models use the same Virchow-v2 as the initial patch encoder, except for OmiCLIP with its own pretrained weights. For $\ours$, we evaluate both retrieval (taking the k-nearest image embeddings and averaging the corresponding gene expression) and direct prediction (\ours decoder). 
\textbf{(D.)} A schematic of gene-to-image retrieval with $\ours$. 
Given a gene or pathway of interest, a molecular query is formulated where important genes are highly expressed, while average expression is used for all other genes. The molecular query embedding is obtained using the \ours-omics encoder, and the cosine similarity with image patches across the slide of interest is calculated. 
\textbf{(E.)} Example of gene-to-image retrieval with $\ours$ in prostate cancer. We build a custom query with a set of genes from the \textit{metabolism \& androgen signaling} and \textit{myogenesis} pathways. The cosine similarity map illustrates the similarity of each image patch to the query, with red (blue) indicating high (low) similarity.
\textbf{(F.)} The cosine similarity map for two different molecular queries (\textit{Tumor cell growth} and \textit{Myogenesis} pathways) using BLEEP and OmniCLIP.
PCC: Pearson Correlation Coefficient. The scale bar is $1\,mm$ unless specified otherwise. 
}
\label{fig:inhouse}
\end{figure*}

\heading{\ours learns a representative morphology-omics embedding space}

We next investigate the quality of the vision and omics embedding space alignment from \ours-finetuning. While the strong performance on patch- and slide-level tasks suggests linear separability in some tasks, we seek to evaluate the encoder quality beyond linear probing. We therefore evaluate the cross-modal retrieval performance between localized morphology and transcriptomics by identifying the closest ST expression (or image patch) based on a query image patch (or ST expression). This setting, also referred to as zero-shot cross-modal inference \cite{jia2021scaling, radford2021learning, huang2023visual, lu_Visuallanguage_2024CONCH}, allows direct evaluation of the embedding space alignment without fitting an additional linear model. Additionally, it enables investigating the relationship between key pathways and morphological features. 

First, we assess image-to-gene retrieval, where for each image patch query we retrieve the 50 nearest ST embeddings from the support (training) set and average them, following retrieval-based strategies\cite{xie_Spatially_2023BLEEP, chen2025visual} (\textbf{Figure~\ref{fig:inhouse}A}). For this analysis, we curate two in-house prostate cancer cohorts: a query cohort of 94,142 Visium spots (eight patients, 20 histological images) and a support cohort of 31,192 Visium spots (six patients, eight histological images). The two cohorts were processed at distinct sequencing cores, leading to visible staining differences due to divergent tissue processing protocols (\textbf{Figure~\ref{fig:inhouse}B}) and introducing additional batch effects. Retrieval performance is quantified with PCC across progressively larger HVG panels.

The comparison of \ours retrieval against other retrieval-based approaches, namely OmiCLIP and BLEEP, reveals several insights (\textbf{Figure~\ref{fig:inhouse}C}). Specifically, \ours on average, achieves a 3.1x higher PCC compared to \mbox{OmiCLIP} across gene sets, which highlights the importance of a domain-specific vision encoder, as \mbox{OmiCLIP} omits both vision-only pretraining on histopathology slides and uses a general-purpose language model to encode highly expressed genes. Additionally, BLEEP performance is substantially worse even with the same vision encoder (Virchow2) as \ours (\mbox{\ours: 0.146} vs. \mbox{BLEEP: 0.003} for \textit{All} genes). This suggests that \ours finetuning, which adds an additional ST reconstruction objective on top of the contrastive objective, creates a joint embedding space that is more robust to batch effects arising from different tissue processing protocols for the query and support sets. In addition, the \ours decoder performs better than \ours-retrieval in a zero-shot setting. This suggests that retrieval-based approaches are less effective for ST prediction as the performance solely depends on the quality of the learned embedding space, as also previously observed (\textbf{Figure~\ref{fig:patch_performance}C}). These observations are replicated with UNI-v2 encoder, highlighting the robustness of cross-modal alignment for \ours.

Building on the strong image-to-gene retrieval performance of \ours, we next examine gene-to-image retrieval (\textbf{Figure~\ref{fig:inhouse}D}). While image-to-gene retrieval can be framed as a standard ST prediction task and quantified with PCC, there is no analogous, well-defined notion of “predicting” images from ST, making gene-to-image retrieval harder to evaluate quantitatively. Instead, our goal is to qualitatively inspect which morphological patterns or WSI regions are most strongly retrieved (i.e., activated) by a given molecular query. To provide more interpretable and functionally meaningful queries than single-spot gene expression, we construct customized transcriptomic queries derived from sets of highly expressed genes associated with specific pathway activation\cite{liberzon2015molecular} (\textbf{Online Methods, Stage III -- Evaluation (Retrieval)} section).

For a prostate cancer slide from $\data$-Test, annotated with three different morphological classes (tumor, stroma, and benign glands), we construct $\ours$ similarity map by visualizing the cosine similarity between the transcriptomics query and all image patch embeddings (\textbf{Figure~\ref{fig:inhouse}E}). Based on the WSI visualization and examples of top similar patches, we observe that for \textit{androgen signaling} and \textit{myogenesis} function-related transcriptomics queries, the tumoral and stromal regions are most activated (red), consistent with the literature\cite{liberzon2015molecular,ma2023prostate,fujita2019role}. In contrast, the top dissimilar patches show the opposite trend, with stromal and adipose/tumoral patterns, respectively. 

Finally, we performed a qualitative comparison of \ours-finetuned Virchow-v2 with BLEEP and OmiCLIP by visualizing the pathway activations for the same prostate cancer sample (\textbf{Figure~\ref{fig:inhouse}F}). We observe that, despite the functionally opposite nature of the \textit{tumor cell growth} and \textit{myogenesis} pathways, which are expected to show more activation in tumor and stromal regions, respectively, almost identical regions are activated for both BLEEP and OmiCLIP. This again suggests that the less structured vision-omics embedding spaces for these approaches do not exhibit natural morphomolecular properties, in contrast to that of $\ours$.

%% file: sections/3-discussion.tex
We introduce \methodlong (\ours), a vision-omics self-supervised learning framework, which shows that the localized morphomolecular information from paired spatial transcriptomics (ST) and hematoxylin and eosin (H\&E) image patches can enhance existing pathology vision foundation models (FMs) for a wide range of downstream tasks. Given any state-of-the-art FMs for pathology, $\ours$ infuses localized transcriptomics expression information through a sample-efficient finetuning process that simultaneously aligns image and ST embeddings and predicts ST expression profiles from image embeddings (\textbf{\Cref{fig:main}}). \ours encoders, finetuned on a large pretraining cohort that combines public and in-house ST-morphology pairs, lead to significantly improved performance over their non-finetuned counterparts, as evaluated across diverse organs with 38 slide-level molecular and morphological outcome prediction tasks (\textbf{\Cref{fig:slide_performance}}) and 15 patch-level ST prediction tasks (\textbf{\Cref{fig:patch_performance}}). $\ours$ finetuning is agnostic to the vision encoder, with consistent results obtained over five popular FMs of different sizes and pretraining data distributions, thereby presenting a highly data- and parameter-efficient finetuning method (\textbf{\Cref{edf:data_efficiency_pretraining}}). As an added benefit, the multi-modal finetuning renders the $\ours$ embeddings more robust to batch effects. Moreover, $\ours$-finetuning establishes a geometrically more meaningful joint embedding space compared to other contrastive alignment approaches, as shown by the superior cross-modal retrieval performance of \ours and more interpretable gene-to-image pathway analysis (\textbf{\Cref{fig:inhouse}}). 

Our design choices distinguish \ours from prior multimodal alignment paradigms in three ways. 
First, we introduce a robust transcriptomics encoder on a fixed-size gene panel prior to the cross-modal alignment. We introduce a scale-invariant reconstruction objective to pretrain a variational autoencoder which balances absolute error with gene-level correlations. This encoder design explicitly handles the high dimensionality, sparsity, and noise of ST data by using normalizing flows to create a more expressive latent model, allowing it to capture complex data distributions and filter measurement noise (\textbf{Online Methods -- Stage I}). 
Second, we employ Low-Rank Adaptation (LoRA) matrices to finetune the vision encoders. LoRA not only provides parameter efficiency but also prevents catastrophic forgetting of the rich prior information of the vision encoders, which we observed with regular finetuning. This allows the model to preserve the morphological information while additionally integrating transcriptomic signals (\textbf{Online Methods -- Stage II}). 
Third, \ours uses a multi-objective training framework to ensure that the resulting vision-omics embeddings are both aligned with one another and directly predictive of gene expression.

In this sense, \ours molecular alignment of a pathology FM is conceptually analogous to post-training alignment in large language models. By exposing vision encoders to localized transcriptomic supervision, \ours reshapes the joint embedding space so that clinically relevant morphomolecular patterns become easier to retrieve across diseases. This suggests that the training of foundation models for pathology should not remain purely vision-only. We envision molecular alignment to be adopted as a systematic post-training stage, applied not only to public resources but also scaled with proprietary ST collections wherever available. A notable consequence of this alignment is that generalizable gains in downstream performance and robustness arise as emergent properties of the training objective, rather than from task-specific finetuning.

Despite the strong performance of the \ours models, several limitations remain. While we demonstrate \ours's efficacy on a selection of five major pathology FMs, this does not cover all publicly available models. \ours is finetuned using Visium data, which captures expression at a spot resolution with multiple cells, rather than at single-cell resolution. However, we anticipate that the same finetuning principles are transferable to technologies such as Xenium or VisiumHD, as indicated by the moderate performance improvements on the HESTBench Xenium tasks. With the increasing availability of higher-resolution ST, we expect that applying \ours can unlock more granular insights into cell-level morphomolecular links. Finally, we use a variational autoencoder with an MLP structure for the transcriptomics encoder to handle relatively small pretraining datasets, instead of heavier transcriptomics FMs trained predominantly on single cells\cite{theodoris2023transfer, cui_ScGPT_2024scGPT}. Community-driven efforts on curating larger-scale pretraining datasets\cite{jaume_HEST1k_2024HEST-1k, chen_STimage1K4M_2024STimage-1K4M, moses2022museum} will allow $\ours$ to leverage both pathology and transcriptomics FMs in the near future.  

In conclusion, the robustness of \ours across different model architectures, sizes, and pretraining strategies suggests that ST-guided finetuning is a generalized capability rather than a dataset-specific effect. Furthermore, \ours establishes a scalable and parameter-efficient paradigm for the next chapter of vision-omics multimodal FM development in computational pathology. We anticipate that this work can be leveraged in further multimodal settings to build multi-scale representations of human tissue and disease. 

%% file: supp/0-supp.tex
\heading{Datasets \& Processing}

\hheading{Training Data.} Our training data, \data-Train, consists of $N=723,250$ spot-image pairs stemming from 353 human diseased (64.2\%), healthy (30.3\%), and treated (5.5\%) patient samples across 14 organs (detailed splits and study sources can be found in \textbf{Extended Data Table~\ref{tbl:data_overview_organ}-~\ref{tbl:train_data_sources}}). This corresponds to the majority of human Visium (98.4\%) and Visium HD (1.6\%) samples from the Hest-1k dataset\cite{jaume_HEST1k_2024HEST-1k}, preprocessed to align ST gene panels and normalize cross-slide effects (see \textbf{Transcriptomics Data Processing}). We removed the STv1 samples from this dataset to ensure a maximum spot size of 55$\mu$m, which STv1 exceeds. To prevent data leakage across the training, validation, and holdout test sets, we split the data by patient, as some patients may have multiple samples, using a 80-10-10 split. To ensure proportional representation across disease and organ types, we use a stratified sampling strategy by organ and select only Visium samples for \data-Test due to the limited availability of Visium HD slides. For ST expression prediction tasks, we report two sets of results. 
First, \data-Test evaluates prediction performance on the top 10 highly variable genes (HVGs) by organ for Brain, Breast, Kidney, Lung, Prostate, and Skin samples in the holdout set, based on variance in \data-Train. Second, we evaluate the prediction performance of the top 50 HVGs on five tissue types of Visium (PRAD, READ, ccRCC, Lymph IDC) and Xenium (LUNG, IDC, PAAD, SKCM, COAD) from additional samples that were held out from \data-train and test set, in line with the Hest Benchmark~\cite{jaume_HEST1k_2024HEST-1k}. These evaluations represent in-distribution (\data) and out-of-distribution (HestBench) patch-level expression prediction tasks. 

\hheading{Image Data Processing.} All Whole Slide Images (WSIs) were first processed using a finetuned feature pyramidal network (FPN) for tissue segmentation. Only patches overlapping with the resulting tissue masks were used for subsequent steps. 
We utilized the HestCore library to extract 224 $\times$ 224px patches at multiple resolutions (0.25, 0.5, and 1.0 $\mu$m per pixel) where the centroid of the patch matches the centroid of the ST spot. Since the tissue samples may exceed the capture area of some ST assays, such as Visium, we select the intersection of patches that can be matched to a spot and barcode the location of all patches such that their relative position can be retrieved and matched to the transcripts. All training samples are normalized across each color channel $c$ $\text{pixel}_{c}(x, y) = \frac{\text{input}_{c}(x, y) - \mu_c}{\sigma_c}$ for channel-mean $\mu_c$ and standard deviation $\sigma_c$ at pixel position $(x, y)$. We used the same means and standard deviations as were used during the training of the respective patch foundation models (UNI-v2-h, CONCH, Virchow-v2, etc.). During training, we randomly sample a magnification for each barcode and apply multiple augmentations to the image, including horizontal (p=0.5) and vertical flips, RGB shifts, color jittering effects, greyscaling, and a Gaussian blur, all applied at varying probabilities (see \textbf{Extended Data Table~\ref{tbl:hyperparams_augmentation}}). 


\hheading{Transcriptomics Data Processing.} We process the transcriptomics samples by filtering and aligning their gene panels. This includes filtering blank, control, and ambiguous transcripts as denoted in the sample metadata. As the gene naming convention was not standardized across samples, we harmonized all samples annotated with Ensembl IDs to HGNC nomenclature using PyBiomart annotations\cite{DeRuiter2016Pybiomart}. We implemented a greedy selection algorithm to select the set of transcriptomics samples with a high degree of gene panel overlap, dropping any samples that do not match a minimum of 5,000 genes from a reference set of genes, which led to the exclusion of four samples. The resulting samples have an aligned set of $G=$ 14,581 genes for all spots. To reduce sparsity, we filter out genes that are only expressed in $<$10\% of spots as well as empty spots that contain no gene counts. We further normalize each spot $i$ by total counts over all genes $j$, such that each spot has the same total count after normalization, to reduce variability related to data capture artifacts like varying sequencing depths across samples, using $\mathbf{x}_{ij}^{(g, norm)} = \frac{\genex{ij}}{\sum_j^{\ngenes} \genex{ij} } \times 1e^{-4}$. We apply a log1p transform to reduce data heteroscedasticity commonly seen in count data using $\genemat = log(\genemat + 1)$ where $\genemat = \left( \mathbf{x}_{i}^{(g, norm)} \right)_{i=1}^{N} \in \mathbb{R}^{\nsamples \times \ngenes}$. For training, we set $G=$ 2,000 by selecting the most variable genes using the Seurat V3 method\cite{stuart_Comprehensive_2019SeuratV3} that bins genes by average expression and selects genes which standardized variance significantly exceeds the expected variance. 
We supplemented the HVG gene panel with 584 Tier 1 genes from the Catalogue of Somatic Mutations in Cancer (COSMIC)\cite{sondka2018cosmic} gene census to guarantee the inclusion of known marker genes.

\hheading{Local Smoothing.} Given that ST assays can produce sparse and noisy data due to variable sequencing coverage, we apply local smoothing to make the expression signal more distinguishable from measurement noise. Concretely, for each center spot $\genex{r,c}$ and its spatial coordinates $(r,c)$, we first identified its set of immediate neighbors on Visium hexagonal spot lattice, $N(r, c)$. We then computed the mean expression of this neighborhood, $\genex{context} = \frac{1}{N_{adj}} \sum_{(r', c') \in N(r, c)} \genex{r', c'}$ where $N_{adj} = |N(r,c)|$ is the number of valid, tissue-containing neighbors, which may be less than 6 spots on the tissue edges. The final smoothed expression value, $\genex{i}$, is the average of the center spot and its context mean $\genex{i} = \frac{\genex{r, c} + \genex{context} }{2}$, where $i=m(r,c)$ is the unique spot index derived from the spot barcode. For computational efficiency, this six-neighbor adjacency graph was pre-calculated for all spots in each sample, allowing computation of $\genex{context}$ in a single vectorized operation.

\heading{\ours Architecture} 

The primary objective of \ours is to finetune patch-level pathology foundation models with spatially resolved molecular information. To do this, \ours has two main components: 1) a self-supervised \textbf{gene model} $g(\cdot)$ that is learning denoised lower-dimensional embeddings $\genez{i} \in \mathbb{R}^{\embdim}$ from the high-dimensional gene inputs $\genex{i} \in \mathbb{R}^{\ngenes}$; and 2) the patch-level \textbf{image model} $f(\cdot)$ that we finetune using low-rank adaptation matrices (LoRA)\cite{hu2022lora}, which is encoding each histology patch $\imgx{i} \in \mathbb{R}^{224\times224\times3}$ into a 1D embedding $\imgz{i} \in \mathbb{R}^{\embdim}$. \ours uses a two-stage training procedure. \textbf{Stage I} is a warmup period for the gene encoder $g(\cdot)$ and is exclusively trained on a gene reconstruction task. \textbf{Stage II} concurrently finetunes the patch encoder $f(\cdot)$ by aligning the gene and image embeddings $\imgz{i}$ and $\genez{i}$ for each spot-image pair, while retaining the Stage~I objective as an auxiliary task to continue improving the quality of the gene embeddings during contrastive alignment. 

\hheading{\ours-omics.} We implement a variational autoencoder (VAE) as our gene model, which consists of an encoder $\geneenc{\cdot}$ and a decoder $\genedec{\cdot}$, both parameterized by $\phi$. The encoder $\geneenc{\cdot}$ is a feedforward network that maps our high-dimensional input gene panel $\genex{i} \in \mathbb{R}^{\ngenes}$ to the parameters of a latent Gaussian distribution, whilst ensuring that the latent embedding $\genez{i}$ matches the image embedding dimension $\embdim$. 
During the forward pass, $\genex{i}$ is transformed through several linear and batch normalization layers with ReLU activations to yield an intermediate representation $\mathbf{h_i} \in \mathbb{R}^{\embdim}$. 
Instead of using $\mathbf{h}$ as is, we linearly project it to the parameters of a Gaussian distribution $\mu_i=W_\mu\mathbf{h}_i$, the standard deviation $\sigma=exp(\frac{1}{2} \log (W_\sigma \mathbf{h}_i))$ and sample $\epsilon \sim \mathcal{N}(0, 1)$, known as the local reparameterization trick\cite{kingma_Variational_2015vae_reparametrization_trick}. Given these projections and the stochastic error, we compute our final latent representation $\genez{i} = \mu_i + \sigma_i \circ \epsilon$. The linear projections in this reparameterization trick are used to enable backpropagation in the stochastic sampling process, and we posit that modeling the latent variables with a distribution may help the model to distinguish the true signal from measurement noise, such as the sparsity mentioned in the previous section. 
The decoder $\genedec{\cdot}$, which mirrors the encoder's feedforward architecture, reconstructs the original gene panel $\geney{i}=\genedec{\genez{i}}$ from the latent embedding. The VAE is trained by optimizing the Evidence Lower Bound, which consists of a Kullback-Leibler (KL) divergence loss to regularize the latent space and the reconstruction loss $\mathcal{L}_{rec, i}(\genex{i}, \geney{i})$, further specified in the section on training \textbf{Stage I}.

\hheading{\ours-vision.} We design \ours to be a general-purpose finetuning recipe that should work well across image encoders. Therefore, we implement several common benchmarks used in computational pathology, namely ResNet50 with the original torchvision weights\cite{he_Deep_2016ResNet50}, the vision encoder of CONCH\cite{lu_Visuallanguage_2024CONCH}, a patch-level contrastively trained vision-language model (ViT-Base with 16$\times$ 16px image tokens, 12 core layers, 90M parameters), 
\mbox{H-optimus-mini}\cite{filiot_Distilling_2025hoptimus-mini}, a distilled version of Hoptimus-0\cite{hoptimus0} (ViT-Base with 14$\times$14px image tokens, 12 core layers, 86M parameters),
Phikon-v2\cite{filiot2024phikon}, a Dinov2-trained foundation model (ViT-Large with 16$\times$16px image tokens), UNI-v2-h\cite{chen_Generalpurpose_2024UNI}, a patch-level foundation model trained at fixed magnification (ViT-Huge with 14$\times$14 px image tokens, 32 core layers, 681M parameters), and Virchow-v2\cite{zimmermann_Virchow2_2024Virchow2a}, a large patch-level model trained on mixed magnification (ViT-Huge with 16$\times$16 px image tokens, 32 core layers, 632M parameters). Further information about each encoder tested can be found in \textbf{Extended Data Table~\ref{tbl:vision_encoder_overview}}. For each encoder $\imgenc{x}$, we fit LoRA adapters (see next section) on a variable number of core layers (specified in \textbf{Extended Data Table~\ref{tbl:seal_hyperparameters}}) and attention heads for finetuning in \textbf{Stage II}. 
Given the input patch $\imgx{i}$, the image encoder produces a latent representation $\imgz{i} = \imgenc{\imgx{i}} \in \mathbb{R}^{\embdim}$. During training, this is used to train the alignment objective with the gene embedding $\imgz{i} \leftrightarrow \genez{i}$, which is further specified the section on \textbf{Stage II}. For inference and evaluation, we train a lightweight task model $\downstream{\cdot}$ with a limited amount of parameters $\gamma$, such as a linear probe or attention-based multiple instance learning for different patch- and slide-level tasks, further detailed in section on \textbf{Stage III}.

\hheading{Low-Rank Adapters.} Directly finetuning all parameters of ViT backbones is both computationally expensive and risks catastrophic forgetting, a phenomenon where the model overwrites its biologically salient, pre-trained morphological knowledge. To circumvent this, we employ Low-Rank Adaptation (LoRA) as a parameter-efficient finetuning technique. The LoRA hypothesis posits that the change in weights during model adaptation ($\Delta W$) has a low intrinsic rank, meaning that the essential information required to adapt the model can be effectively compressed using a much smaller number of parameters than the full weight matrix. Therefore, we approximate the weight update using a low-rank decomposition. For a given pre-trained weight matrix $W_0 \in \mathbb{R}^{d \times k}$, the update $\Delta W_0$ can be represented by two smaller, trainable matrices, $B \in \mathbb{R}^{d \times r}$ and $A \in \mathbb{R}^{r \times k}$, where the rank $r \ll \min(d, k)$. During training, the original weights are frozen and only $A$ and $B$ are updated. The modified forward pass for the layer then becomes $h = W_0x + \Delta W x = W_0x + (\frac{\alpha}{r})BAx$. For \ours, we use a rank of $r=8$ and alpha $\alpha=8$ (\textbf{Extended Data Table~\ref{tbl:seal_hyperparameters}}).

\heading{\ours Pipeline}

\hheading{Stage I -- Self-supervised ST Representations.} One main challenge in finetuning vision encoders with spatial transcriptomics information is to learn meaningful and generalizable embeddings $\genez{i}$ for each transcriptomics spot $i$ that account for common challenges with ST data such as their high dimensionality, highly unequal variance, data sparsity, noise, as well as local and global expression context. High dimensionality, data sparsity, and local context are partially handled through gene selection and local smoothing in the preprocessing pipeline (detailed in the section on \textbf{Datasets \& Processing}). To handle noisy data, we use a VAE, which has shown effective denoising properties on ST data\cite{tian_Dependencyaware_2024SpaVAE}. 

We iterate on the standard \textbf{VAE architecture}~\cite{kingma_Variational_2015vae_reparametrization_trick}, which typically assumes an approximate diagonal Gaussian posterior over the latent variables. While this assumption makes optimization tractable, it restricts the posterior to simple shapes. For ST dat, this can represent a limitation, as gene expression distributions are often sparse, skewed, and governed by non-linear dependencies between genes~\cite{stuart_Comprehensive_2019SeuratV3,tian_Dependencyaware_2024SpaVAE}. To address this, we augment the VAE with planar normalizing flows. After sampling the initial latent variable 

\begin{equation}
    \label{eq:vae-standard-latent-sample}
    z_0 = \mu + \sigma \odot \epsilon, \quad \epsilon \sim \mathcal{N}(0, I),
\end{equation}

we iterative apply a set of learnable and invertible transformations of the form 
\begin{equation}
    \label{eq:vae-planar-flow-transform}
    z' = z + \mathbf{u}\,\tanh(\mathbf{w}^\top z + b),
\end{equation}

where $\mathbf{u}$, $\mathbf{w} \in \mathbb{R}^d$ and $b \in \mathbb{R}$ are trainable parameters.

Intuitively, each planar flow performs a smooth and non-linear transform of the latent space along a learned direction~\cite{rezende2015variational}. Starting from the simple initial Gaussian sample (Equation~\ref{eq:vae-standard-latent-sample}), successive flow steps (Equation~\ref{eq:vae-planar-flow-transform}) warp the latent distribution, allowing it to approximate more complex geometries. Since each flow step reshapes the latent space, we adjust the probability density using a density correcting term accordingly to ensure that the transformed posterior remains a properly normalized distribution.

Beyond the model architecture, we require a robust \textbf{training objective} to learn spot-level representations that: 1) is highly correlated with the ground truth gene panel; and 2) ensure a high-fidelity reconstruction on the initial panel $\geney{i} = \genedec{\genez{i}}$. Some prior works use regression-based reconstruction objectives, such as Mean Squared Error or Huber loss to evaluate the reconstruction\cite{xie_Spatially_2023BLEEP, lee_PathOmCLIP_2024PathOmCLIP}, given by Equation~\ref{eq:rec_loss_mse} as
\begin{equation}
    \label{eq:rec_loss_mse}
    \mathcal{L}_{mse} = \frac{1}{\ngenes} \sum_{i=1}^{G}(\genex{i} - \geney{i})^2 . 
\end{equation}

However, since the gene panel has a strong unequal variance, also after the count-normalization and log1p transforms, we found that objectives exclusively correcting for absolute error over-emphasize highly variable genes and neglect genes with lower variance, leading to a biased training objective. To account for lower variance genes, we introduce an additional loss term that is scale-invariant by repurposing the original Barlow Twins loss\cite{zbontar_Barlow_2021BarlowTwins}. While this loss was originally introduced as an alignment loss between an original and a distorted view of the same data, we found this to work well as a scale-invariant reconstruction loss, which consists of two main components. 
First, we calculate the gene-level cross-correlation along the batch dimension for each predicted gene in the batch $\left( \geney{b, i} \right)_{b=1}^{B}$ and its ground truth counterpart $\left( \genex{b, i} \right)_{b=1}^{B}$. 
\begin{equation}
    \label{eq:cross_corr}
    C_{ij} = \frac{\sum_b \genex{b,i}\geney{b,j}}{\sqrt{\sum_b\left(\genex{b, i}\right)^2} \sqrt{\sum_b\left(\geney{b, j}\right)^2}}. 
\end{equation}

Given the cross-correlation matrix for the training batch, we maximize the element-wise correlation between each predicted gene and the ground truth, which corresponds to the Barlow Twins invariance term (\Cref{eq:cross_corr,eq:rec_loss_inv}), which in our context means that the reconstructed gene panel is invariant to a uniform scaling of the gene expressions. As such, this objective ignores absolute magnitude and is therefore robust to the unequal variance across genes. 
Second, we minimize the correlation between unpaired genes, meaning that high values on the off-diagonal of $C$ that correspond to redundant information are penalized (Equation~\ref{eq:rec_loss_red}). In our context, this prevents the VAE from trivially grouping together highly correlated genes which may ignore subtle differences. However, since many genes' expressions may be correlated for legitimate reasons, we apply a lower weighting $\lambda_1$ to this penalty (Equation~\ref{eq:rec_loss_total}). 
\begin{align}
    \mathcal{L}_{inv} &= \sum_i^{\ngenes}(1-C_{ii})^2 \label{eq:rec_loss_inv} \\ 
    \mathcal{L}_{red} &= \sum_i^{\ngenes}\sum_{j\neq i} {C_{ij}}^2. \label{eq:rec_loss_red} 
\end{align}

The final loss is a weighted sum between our three main loss components, balancing the penalty between the correlation-based losses $\mathcal{L}_{inv}$ and $\mathcal{L}_{red}$ that are unbiased and invariant to the magnitude of the predictions, and $\mathcal{L}_{mse}$ that accounts for the magnitude of the predictions. 
\begin{equation}
    \label{eq:rec_loss_total}
    \mathcal{L}_{rec} = \lambda_0 \mathcal{L}_{inv} + \lambda_1 \mathcal{L}_{red} + \lambda_2 \mathcal{L}_{mse}
\end{equation}

We train the spot-level VAE with a batch size of 384 for a maximum of 3 epochs as a warm-up period with Adaptive Moment Estimation (Adam\cite{kingma_Adam_2017Adam}) as a stochastic optimizer using a learning rate of 0.0005, which is updated using a cosine annealing learning rate schedule. The full list of hyperparameters in the final models can be found in \textbf{Extended Data Table~\ref{tbl:seal_hyperparameters}}.


\hheading{Stage II - Contrastive Alignment.} Following warm-up period for the VAE $g(\cdot)_\phi$, we start the finetuning of the image encoder $\imgenc{\cdot}$. After loading in its pre-trained weights, we unfreeze the final \textbf{X} blocks of the encoder. We then align the image embeddings $\imgz{}$ and gene embeddings $\genez{}$ using an InfoNCE loss, which maximizes their joint likelihood. Given a set of $N \times N$ samples in each training batch, we maximise the joint likelihood of each pair while minimizing the likelihood of negative pairs (non-matching H\&E-ST pairs) in the batch: 
\begin{equation}
    \label{eq:InfoNCE}
    \mathcal{L}_{\text {InfoNCE }}=-\frac{1}{N} \sum_{i=1}^N\left[\log \frac{\exp \left(\left\langle \imgz{i}, \genez{i} \right\rangle / \tau\right)}{\sum_{k=1}^N \exp \left(\left\langle\imgz{i}, \genez{k}\right\rangle / \tau\right)}+\log \frac{\exp \left(\left\langle\imgz{i}, \genez{i}\right\rangle / \tau\right)}{\sum_{k=1}^N \exp \left(\left\langle\imgz{k}, \genez{i}\right\rangle / \tau\right)}\right], 
\end{equation}
where $\langle\cdot\rangle$ is the cosine similarity between embeddings, and $\tau$ is the temperature of the softmax. Intuitively, this loss forces matching embeddings to be close together with respect to their cosine similarity while pushing non-matching pairs apart such that we learn a joint embedding space, where each image patch is aligned with its corresponding gene profile, while mismatching pairs are easily distinguishable. Relevant hyperparameters for the contrastive loss are specified in \textbf{Extended Data Table~\ref{tbl:seal_hyperparameters}}. 

To preserve the gene panel \textbf{reconstruction} capabilities of \ours during the alignment, we maintain the gene reconstruction task as an auxiliary objective. The \ours-omics branch continues to be optimized using the Stage I objective $\mathcal{L}_{rec, gene}$ (Equation~\ref{eq:rec_loss_total}). To ensure the morphological features are concretely predictive of gene expression, the \ours-vision branch is also trained to reconstruct the transcriptomic profile from the image patch directly, optimized via $\mathcal{L}_{rec, vision}$.

Finally, the \ours dataset is a heterogeneous collection of studies, organs, and ST technologies, which introduces significant \textbf{batch effects} that do not correspond to biological signal. We implement a domain adaptation strategy as a further auxiliary task to ensure that the learned representations are discriminative for the primary task but invariant to the shift between data domains. We achieve this by training an auxiliary domain classifier to predict the study origin of each sample from both the vision and omics encoders. Crucially, a gradient reversal layer (GRL)\cite{ganin_Unsupervised_2015DomainAdaptation} is inserted between the feature encoders and the domain classifier. During the forward pass, the GRL acts as an identity transform. Meanwhile, during backpropagation, the GRL multiplies the gradient it receives from the domain classifier by a negative constant before passing it to the preceding encoders. This creates an adversarial training objective where the domain classifier's parameters are optimized to minimize domain classification error (i.e., spotting batch effects) while the feature encoders' parameters are optimized to maximize the same error (i.e., making features from different domains more indistinguishable). This process applies a higher domain adaptation penalty $\mathcal{L}_{da}$ if the model learns domain-specific artifacts and forces the finetuning process to produce more generalizable embeddings. 

The final loss function for Stage II training is a weighted function between the contrastive image reconstruction, gene reconstruction, and domain adaptation objectives: 
\begin{equation}
    \label{eq:stage2_loss}
    \mathcal{L} = \lambda_0 \mathcal{L}_{InfoNCE}(\imgz{}, \genez{}) + \lambda_1 \mathcal{L}_{rec, img}(\imgy{}) + \lambda_2 \mathcal{L}_{rec, gene}(\geney{}) + \lambda_3 \mathcal{L}_{da}, 
\end{equation}%
where the values of $\lambda$ are picked to ensure that no single loss dominates the gradient updates.


\hheading{Stage III - Evaluation.} 

\hheading{Evaluation Scope.}
To demonstrate the versatility of the learned morphomolecular embeddings, we evaluate \ours on a highly diverse set of 38 slide-level downstream tasks from 14 datasets across seven major organs (Breast, Brain, Colon, Kidney, Lung, Ovary, Pancreas) and 15 patch-level downstream tasks across nine major organs (Breast, Brain, Colon, Kidney, Lung, Lymph Node, Prostate, Rectum, Skin). Our evaluation covers five types of tasks: gene expression prediction (\textbf{Extended Data Tables~\ref{tab:prostate_results} to~\ref{tbl:hestbench}}), 
molecular status prediction (\textbf{Extended Data~\Cref{tab:bc_therapy_er_status_results,tab:bc_therapy_her2_status_results,tab:bcnb_er_results,tab:bcnb_pr_results,tab:bcnb_her2_results,tab:cptac_brca_TP53_mutation_results,tab:cptac_brca_PIK3CA_mutation_results,tab:cptac_ccrcc_VHL_mutation_results,tab:cptac_ccrcc_BAP1_mutation_results,tab:cptac_coad_MSI_H_results,tab:cptac_coad_KRAS_mutation_results,tab:cptac_coad_TP53_mutation_results,tab:cptac_gbm_TP53_mutation_results,tab:cptac_gbm_EGFR_mutation_results,tab:cptac_luad_EGFR_mutation_results,tab:cptac_luad_TP53_mutation_results,tab:cptac_luad_STK11_mutation_results,tab:mut-het-rcc_PBRM1_mutation_results,tab:mut-het-rcc_SETD2_mutation_results}}),
pathway expression prediction (\textbf{Extended Data~\Cref{tab:cptac_gbm_HALLMARK_G2M_CHECKPOINT_results,tab:cptac_brca_HALLMARK_G2M_CHECKPOINT_results,tab:cptac_coad_HALLMARK_G2M_CHECKPOINT_results,tab:cptac_luad_HALLMARK_G2M_CHECKPOINT_results,tab:cptac_gbm_HALLMARK_PI3K_AKT_MTOR_SIGNALING_results,tab:cptac_pda_HALLMARK_PI3K_AKT_MTOR_SIGNALING_results,tab:cptac_brca_HALLMARK_PI3K_AKT_MTOR_SIGNALING_results,tab:cptac_luad_HALLMARK_PI3K_AKT_MTOR_SIGNALING_results,tab:cptac_gbm_HALLMARK_EPITHELIAL_MESENCHYMAL_TRANSITION_results,tab:cptac_pda_HALLMARK_EPITHELIAL_MESENCHYMAL_TRANSITION_results,tab:cptac_brca_HALLMARK_EPITHELIAL_MESENCHYMAL_TRANSITION_results,tab:cptac_coad_HALLMARK_EPITHELIAL_MESENCHYMAL_TRANSITION_results,tab:cptac_luad_HALLMARK_EPITHELIAL_MESENCHYMAL_TRANSITION_results,tab:cptac_pda_HALLMARK_G2M_CHECKPOINT_results}}),
IHC status prediction (\textbf{Extended Data Figure~\ref{edf:morphological_tasks}}), treatment response prediction (\textbf{Extended Data Table~\ref{tab:ovarian_response_results}}), 
survival prediction (\textbf{Extended Data Table~\ref{tab:sr386__OS_results}}), 
gene-to-image retrieval, and image-to-gene retrieval (\textbf{Figure~\ref{fig:inhouse}}). We provide a brief description of each dataset below.
\begin{itemize}
    \item\textbf{BC therapy.} We used the public BC therapy\cite{sammut2022bctherapy} dataset comprising 166 patients with invasive breast cancer, each associated with an H\&E-stained image of the pre-neoadjuvant therapy biopsy. We design two molecular prediction tasks: prediction of ER status (n=166 cases) and prediction of HER2 status (n=166 cases), and cancer grade prediction (n=166 cases). All slides were scanned with an Aperio scanner at 20$\times$ (0.50$\mu$m/px). Results are reported in \textbf{Extended Data~\Cref{tab:bc_therapy_grade_results,tab:bc_therapy_er_status_results,tab:bc_therapy_her2_status_results}.}
    \item \textbf{BCNB.} We used the public BCNB dataset\cite{xu2021predicting} (Breast Cancer Core-Needle Biopsy) for IHC status prediction in breast cancer. BCNB comprises 1,058 WSIs (one WSI per patient), which we use for ER status prediction, PR status prediction, and HER2 status prediction. Additional information is provided in \textbf{Extended Data \Cref{tab:bcnb_er_results,tab:bcnb_pr_results,tab:bcnb_her2_results}}.
    \item\textbf{BReAst Carcinoma Subtyping (BRACS).} We used the public BRACS dataset\cite{brancati2021bracs} for coarse- and fine-grained breast morphological subtyping. BRACS consists of 547 breast carcinoma WSIs from 189 patients sourced from IRCCS Fondazione Pascale. Each WSI is used for coarse-grained (3 classes) and fine-grained (6 classes) morphological subtyping. Due to the limited size of the official test set, we redefined train-test splits with an 80:20 ratio. Because BRACS-Fine and BRACS-Coarse are slide-level prediction tasks with multiple slides per case, we kept all slides belonging to the same patient together, ensuring one patient does not end up in both train and test. Therefore, this dataset was not explicitly label-stratified (as each patient would have multiple labels). Results are reported in  \textbf{Extended Data~\Cref{edf:morphological_tasks}}.
    \item\textbf{CPTAC.} We used the public CPTAC data for pan-cancer mutation prediction\cite{edwards2015cptac,thangudu2020cptac}. Specifically, we included
    (1) CPTAC-BRCA (invasive breast cancer) for PIK3CA and TP53 mutation prediction,
    (2) CPTAC-CCRCC (clear-cell renal-cell carcinoma) for BAP1, VHL, and PBRM1 mutation prediction,
    (3) CPTAC-COAD (colon adenocarcinoma) for KRAS, SETD1B, TP53, APC, PIK3CA, ARID1A, and ACVR2A mutation prediction and MSI status,
    (4) CPTAC-GBM (glioblastoma) for EGFR and TP53 mutation prediction,
    (5) CPTAC-LUAD (lung adenocarcinoma) for EGFR, STK11, and TP53 mutation prediction, 
    (6) CPTAC-PDAC (pancreatic ductal adenocarcinoma) pathway expression prediction. 
    For all tasks, we predict the pathway expression for the PI3K/AKT/mTOR signalling (MTOR), Epithelial-Mesenchyman Transition (EMT), and G2/M checkpoint (G2M) pathways from the Hallmark gene sets\cite{liberzon2015molecular_hallmark}.
    Results are reported in \textbf{Extended Data~\Cref{tab:cptac_coad_MSI_H_results,tab:cptac_gbm_EGFR_mutation_results,tab:cptac_gbm_TP53_mutation_results,tab:cptac_brca_TP53_mutation_results,tab:cptac_ccrcc_VHL_mutation_results,tab:cptac_coad_KRAS_mutation_results,tab:cptac_coad_TP53_mutation_results,tab:cptac_luad_EGFR_mutation_results,tab:cptac_luad_TP53_mutation_results,tab:cptac_ccrcc_BAP1_mutation_results,tab:cptac_luad_STK11_mutation_results,tab:cptac_brca_PIK3CA_mutation_results,tab:cptac_gbm_HALLMARK_G2M_CHECKPOINT_results,tab:cptac_pda_HALLMARK_G2M_CHECKPOINT_results,tab:cptac_brca_HALLMARK_G2M_CHECKPOINT_results,tab:cptac_coad_HALLMARK_G2M_CHECKPOINT_results,tab:cptac_luad_HALLMARK_G2M_CHECKPOINT_results,tab:cptac_gbm_HALLMARK_PI3K_AKT_MTOR_SIGNALING_results,tab:cptac_pda_HALLMARK_PI3K_AKT_MTOR_SIGNALING_results,tab:cptac_brca_HALLMARK_PI3K_AKT_MTOR_SIGNALING_results,tab:cptac_luad_HALLMARK_PI3K_AKT_MTOR_SIGNALING_results,tab:cptac_gbm_HALLMARK_EPITHELIAL_MESENCHYMAL_TRANSITION_results,tab:cptac_pda_HALLMARK_EPITHELIAL_MESENCHYMAL_TRANSITION_results,tab:cptac_brca_HALLMARK_EPITHELIAL_MESENCHYMAL_TRANSITION_results,tab:cptac_coad_HALLMARK_EPITHELIAL_MESENCHYMAL_TRANSITION_results,tab:cptac_luad_HALLMARK_EPITHELIAL_MESENCHYMAL_TRANSITION_results}}.
    \item\textbf{DHMC-Kidney.} We used the public DHMC-Kidney dataset\cite{farahmand2022hertumorroi} comprising 563 cases of renal cell carcinoma, each associated with one H\&E-stained slide (\textbf{Extended Data Figure~\ref{edf:morphological_tasks}}).
    \item\textbf{EBRAINS.} We used the EBRAINS dataset\cite{roetzer2022ebrains} for coarse- and fine-grained brain tumor subtyping, as well as IDH1 mutation prediction. EBRAINS consists of 3,114 WSIs acquired by the EBRAINS Digital Tumor Atlas at the University of Vienna. We reused splits from UNI\cite{chen_Generalpurpose_2024UNI}, which kept categories with at least 30 samples, resulting in 2,319 slides. Each WSI is used for fine-grained morphological subtyping across 30 diagnostic classes. Splits are stratified by patients to ensure slides from a patient are not found in both train and test splits. Additional information is provided in \textbf{Extended Data~\Cref{edf:morphological_tasks}}.
    \item \textbf{ICAIRD.} We used the ICAIRD~\cite{InHwaUm2024icaird} data collection, consisting of H\&E-stained 100 whole slide images of renal cell carcinoma (Kidney) captured across five commonly used high-resolution slide scanners, each representing different data acquisition domains. As such, the images have subtle differences in color, contrast, and resolution depending on the scanner used. We test this dataset to investigate visual batch effects that pathology foundation models may pick up instead of the true biological signal in the slides. For this study, we included images from the Hamamatsu-2.0-HE (0.45 $\mu$m/px), Hamamatsu S60 (0.42 $\mu$m/px), Leica Aperio AT2 ($\mu$m/px), Zeiss Axio Z1-Grant (0.22 $\mu$m/px), and Zeiss Axio Z1-Louise (0.22 $\mu$m/px). 
    \item \textbf{MGB-Breast.} We used an internal cohort of invasive breast cancer (BRCA) for morphological and immunohistochemistry status prediction\cite{vaidya2024demographic,jaume2024multistain}. MGB-Breast comprises 1,264 WSIs (mix of biopsies and resections) scanned from Brigham and Women's Hospital (one WSI per patient). We curated two immunohistochemistry (IHC) status prediction tasks: estrogen receptor (ER) status prediction and progesterone receptor (PR) status prediction. ER and PR status were manually extracted from pathology reports. Results are provided in \textbf{Extended Data Figure~\ref{edf:morphological_tasks}}.
    \item\textbf{MUT-HET-RCC.} We used the MUT-HET-RCC dataset\cite{acosta2022intratumoral} for mutation prediction in renal cell carcinoma. MUT-HET-RCC comprises 1,291 WSIs (one WSI per patient) which we use for (1) PBRM1 mutation prediction, and (2) SETD2 mutation prediction. Additional information is provided in \textbf{Extended Data Table~\Cref{tab:mut-het-rcc_PBRM1_mutation_results,tab:mut-het-rcc_SETD2_mutation_results}}.
    \item\textbf{Ovarian Treatment Response.} We used the OV-Bevacizumab dataset\cite{wang2022weakly} for treatment response prediction in ovarian cancer. OV-Bevacizumab consists of 288 WSIs from 78 patients. Non-responders are defined as having a measurable regrowth of the tumor or as a serum CA-125 concentration of more than twice the value of the upper limit of normal during the treatment course for the bevacizumab therapy. We kept all patients who received bevacizumab as their first-line treatment and additionally removed four cases (case IDs: P00181938C, 2630938, 2224393, 2937351), which were labeled as both responders and non-responders, yielding 85 WSIs from 36 patients. Additional information is provided in \textbf{Extended Data~\Cref{tab:ovarian_response_results}}.
    \item\textbf{PANDA.} We used the public PANDA (Prostate cANcer graDe Assessment) data for prostate cancer grading (ISUP grading)\cite{bulten2022artificial}. PANDA comprises 10,616 core needle biopsies from Radboud University Medical Center and Karolinska Institute, each annotated with an ISUP grade (6-class classification task). We follow prior work\cite{chen_Generalpurpose_2024UNI,pati2022hierarchical} in excluding slides with equivocal labels~\cite{vaidya_Moleculardriven_2025THREADS}, which resulted in 9,555 slides. We used the train split (7,647 WSIs) and test split (954 WSIs) from GigaPath \cite{xu_Wholeslide_2024GigaPath}, and we did not use their validation split (954 WSIs). Additional information is provided in \textbf{Extended Data Table \ref{edf:morphological_tasks}}.
    \item\textbf{SURGEN.} We used public cases from the SR386 cohort of SurGen\cite{myles2024leveraging}, which includes 389 patients with colon and rectum adenocarcinoma, predicting overall survival for each patient. Additional information is provided in \textbf{Extended Data~\Cref{tab:sr386__OS_results}}.
\end{itemize}

\hheading{Linear Probing (Patch-level).} Patch-level tasks are evaluated using a linear probe of the frozen patch embeddings. We assess the quality of these representations on two patch-level gene expression prediction tasks (\ours-Test and HestBench). For the linear probe, we first apply Principal Component Analysis (PCA) to the frozen embeddings to reduce the dimensionality to 256 factors. This is done to prevent performance penalties of the downstream probe for larger embeddings\cite{jaume_HEST1k_2024HEST-1k}. A Ridge regression model is trained on the PCA components to predict the count-normalized HVG expression levels. All results are evaluated using \mbox{5-fold} cross-validation, with the performance quantified by the Pearson Correlation Coefficient (PCC) and Mean Squared Error (MSE) between the predicted and ground truth expression values across folds.

\hheading{Multiple Instance Learning (Slide-level). } To evaluate slide-level task performance, the patch embeddings from a given WSI are aggregated into a single slide-level representation using two standard Multiple Instance Learning (MIL) methods. We use mean pooling as a simple non-parametric baseline where all patch embeddings are averaged to produce the slide-level feature vector. Additionally, we train Attention-based MIL (ABMIL)\cite{ilse2018attention}, a learnable aggregation method where a gated attention mechanism assigns an importance score to each patch embedding to emphasize more relevant tissue regions. The slide-level representation is then calculated by averaging patch embeddings, weighted by the attention scores. 

\hheading{Retrieval.} \ours has the ability to run cross-modal image-to-gene (i2g) and gene-to-image (g2i) retrieval. Each of these approaches uses a query vector and a set of reference vectors from the aligned embedding space. For i2g retrieval, we use a patch embedding $\imgz{i}$ as the query and the set of all ST embeddings from a support set as the reference vectors. 
We calculate the cosine similarity vector (for single queries) or matrix (for batched queries) with each value as $s_{ij}=cos(\imgz{i}, \genez{j})$ for all $\genez{j} \in \mathcal{Z}_{train}^{(g)}$ and select the set of indices $\mathcal{K}$ for the Top-K retrieved ST embeddings (with K=50). The final i2g retrieved expression profile $\geney{i}$ is the similarity-weighted average of the corresponding ground-truth gene panels $\{ x_k^{(g)} \}_{k \in \mathcal{K}}$:
\begin{equation}
    \label{eq:i2g_retrieval_yhat}
    \geney{i} = \frac{\sum_{k \in \mathcal{K}} s_{ik} \cdot \genex{k}}{\sum_{k \in \mathcal{K}} s_{ik}}. 
\end{equation}

For g2i retrieval, we construct an in silico transcriptomic profile $x_{query}^{(g)} \in \mathbb{R}^G$ based on a specific biological pathway or gene signature. First, we define the ``active'' genes such as \textit{KLK3} and \textit{AMACR} for Prostate retrieval. This gene set is expanded by identifying correlated genes from the train gene panel, selecting all genes with a PCC$>$0.3 to the active genes. Next, we filter all ST spots within the \data-Train data for the query organ, retaining only spots with at least 50\% of genes that show expression above the 75th percentile for each gene. The \ours-omics embeddings $\genez{j}$ for all these spots are then averaged to create a single and robust molecular query $x_{query}^{(g)}$ and its embedding $z_{query}^{(g)}$. Using this query embedding and the patch reference set $Z_{train}^{(p)}$, we retrieve the patches with the highest cosine similarity.

\hheading{Evaluation Metrics.} 
We evaluate binary classification tasks using \textbf{macro AUC}, a threshold-free measure that plots the true positive rate against the false positive rate. The macro-AUC score ranges from 0.5, indicating random prediction, to 1, indicating perfect prediction. For multi-class subtyping, we report \textbf{balanced accuracy}, which accounts for class imbalance by computing the unweighted average of recall for each class. To assess multi-class grading tasks (rank-ordered), we utilize the \textbf{quadratic weighted Cohen's kappa}, which quantifies agreement between model and ground truth while penalizing disagreement on the distance between categories. Kappa scores range from -1 (complete disagreement) to 1 (perfect agreement), while 0 indicates agreement by random chance. 
For regression tasks, including gene expression and pathway expression prediction, we assessed performance using the Pearson Correlation Coefficient (PCC), Spearman's rank correlation coefficient, and Mean Squared Error (MSE). PCC measures the strength and direction of the linear relationship between predicted and ground truth values of expression as a measure between -1 and 1, with a higher correlation coefficient indicating improved predictive performance. Spearman's rank correlation captures the relative ordering of expression levels independent of their absolute magnitude or specific data distribution. Finally, to track the absolute magnitude, we also track the mean squared error between predicted and ground truth values.

\hheading{Statistical analysis.} For all tasks, we report the mean and standard error across all folds for the task-relevant evaluation metrics. If only a single fold is available, we estimate the confidence intervals through bootstrapping 100 iterations. We compare the performance after \ours training to the vision-only baselines using a one-sided paired Wilcoxon Signed-Rank to determine whether the performance difference between the models is statistically significant for all evaluation metrics.


\heading{Computing Hardware and Software} 
This project is implemented using Python 3.9 and PyTorch (version 2.1.2) with CUDA (12.1) (\url{https://pytorch.org/}), which can be replicated by leveraging several open-source libraries as outlined below. All training was done on three 24 GB NVIDIA GeForce RTX 3090 GPUs using PyTorch's distributed data parallel backend, which was called through Huggingface's \texttt{accelerate} library (\url{https://huggingface.co/docs/accelerate/en/index}). 
Implementations of the slide encoder backbones can be found at the following links: CONCH (\url{https://huggingface.co/MahmoodLab/CONCH}), H0-mini(\url{https://huggingface.co/bioptimus/H0-mini}), Phikon-v2 (\url{https://huggingface.co/owkin/phikon-v2}), UNI-v2-h (\url{https://huggingface.co/MahmoodLab/UNI2-h}), and Virchow-v2 (\url{https://huggingface.co/paige-ai/Virchow2}). Slide processing was done using the TRIDENT library (\url{https://github.com/mahmoodlab/trident}), which builds on OpenSlide (version 4.0.0.6). Processing and analysis of spatial transcriptomics data was performed using scanpy (version 1.10.3). The LoRA implementation uses the PEFT library (version 0.15.2). The OpenClip implementation was used (\url{https://github.com/mlfoundations/open_clip}). The linear probing and dimensionality reduction was implemented through the NVIDIA Rapids \texttt{cuml} library (version 24.8). Image augmentations were implemented using albumentations (\url{https://albumentations.ai/}). Statistical testing was done using scipy's \texttt{stats} package (version 1.15.2) and further metrics were calculated using scikit-learn (version 1.4.0). Further dependencies can be found in the \texttt{environment.yaml} file on \github.

\heading{Code availability}
The model weights and code are available at \github.

\heading{Data availability} The source data used to derive the \ours training, test, and HestBench can be accessed using the HEST package (\url{https://github.com/mahmoodlab/HEST}). Data splits for the public downstream tasks are available at (\url{https://github.com/mahmoodlab/Patho-Bench}). The slides for the majority of downstream tasks are publicly available CPTAC (\url{https://www.cancerimagingarchive.net/collection/cptac-brca/}), Ebrains (\url{https://doi.org/10.25493/WQ48-ZGX}), BRACS (\url{	https://www.bracs.icar.cnr.it/}), BCNB (\url{https://bupt-ai-cz.github.io/BCNB/}), PANDA (\url{	https://panda.grand-challenge.org/data/}), MUT-HET-RCC (\url{	https://doi.org/10.25452/figshare.plus.c.5983795}), Surgen (\url{https://www.ebi.ac.uk/biostudies/bioimages/studies/S-BIAD1285}), DHMC-Kidney (\url{https://bmirds.github.io/KidneyCancer/}), BC Therapy (\url{	https://zenodo.org/records/6337925#.Y30d1y-l1Ls}). Further evaluation data from BWH and MGH are proprietary patient data, and cannot be made publicly available. 

%% file: supp/1-additional.tex
\paragraph{Ethics Statement}
The retrospective analysis of internal pathology images and associated reports used in this study received approval from the Mass General Brigham institutional review board. Prior to the computational analysis and model development, all internal digital data, including whole slide images (WSIs), pathology reports, and electronic medical records, were anonymized. Since the study did not involve direct patient participation or recruitment, informed consent was waived for the analysis of archival pathology slides. 

\section*{Author Contributions}
K.H, A.H.S, and F.M conceived the study and designed the experiments. 
C.A.P conducted experimental analysis, especially for cross-modal retrieval tasks and the preparation and analysis of the in-house prostate samples. 
G.J helped the curation of the MAPLE dataset.
S.J.W and A.V helped set up the evaluation pipeline for slide-level tasks.
K.H and A.H.S prepared the manuscript. All authors contributed to the writing. 
N.S, M.J and F.M supervised the research. 

\section*{Acknowledgements}
This work was funded in part by the Brigham and Women’s Hospital (BWH) President’s Fund, Mass General Hospital (MGH) Pathology and by the National Institute of Health (NIH) National Institute of General Medical Sciences (NIGMS) through R35GM138216.
K.H was supported by a scholarship through the Gates Cambridge Trust and University of Cambridge Computer Laboratory Travel Grant. C.A.P was supported by the Rafael del Pino Foundation. N.S and M.J acknowledge the support of the U.S. Army Medical Research and Development Command of the Department of Defense through the FY22 Breast Cancer Research Program of the Congressionally Directed Medical Research Programs, Clinical Research Extension Award GRANT13769713.
The content is solely the responsibility of the authors and does not reflect the official views of the NIH, NIGMS, and DoD.

%% file: supp/2-figures.tex
\begin{figure*}
\centering
\includegraphics[width=\textwidth]{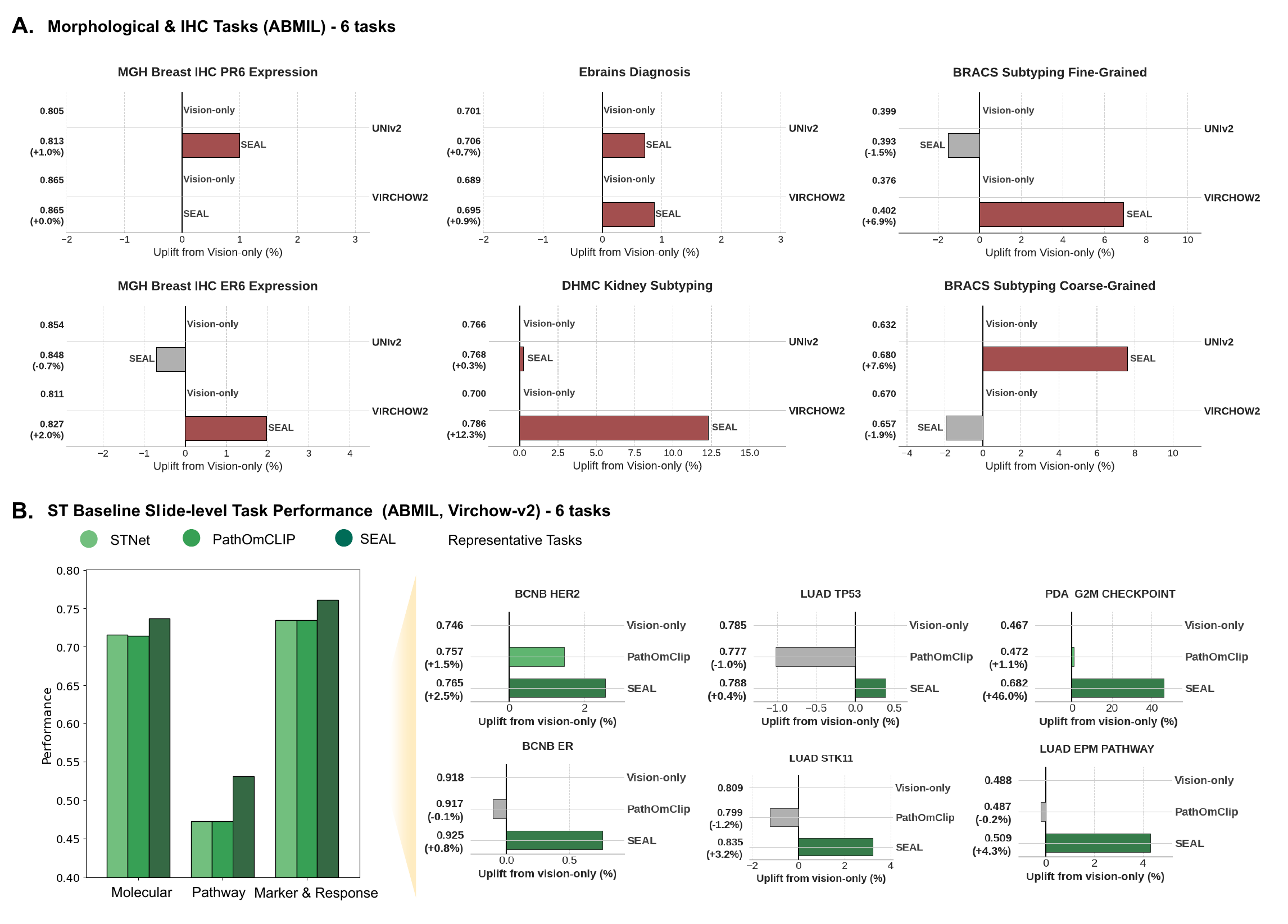}
\caption{\textbf{Slide-level morphological and IHC tasks \& slide-level comparison with ST baselines.}
\textbf{(A)}. Slide-level performance on two IHC slide-level prediction and four slide-level morphological classification tasks: IHC ER6 and PR6 expression prediction (6 classes, measured using Cohen's weighted kappa), Ebrains diagnosis classification (30 classes, measured using balanced accuracy), DHMC Kidney subtyping (5 classes, measured using balanced accuracy), BRACS subtyping as fine-grained (7 classes, measured using balanced accuracy) and coarse-grained (3 classes, measured using balanced accuracy) classification task. The chart shows the relative change of \ours over the vision-only baseline for Uni-v2 and Virchow-v2.
\textbf{(B)}. Comparison of \ours with the two best-performing patch-level fine-tuning methods (ST-Net, PathOmCLIP) on the full set of slide-level tasks: molecular status prediction, pathway expression, and marker \& response prediction tasks. Performance is measured relative to the ST-Net baseline. 
}
\label{edf:morphological_tasks}
\end{figure*}

\begin{figure*}
\centering
\includegraphics[width=\textwidth]{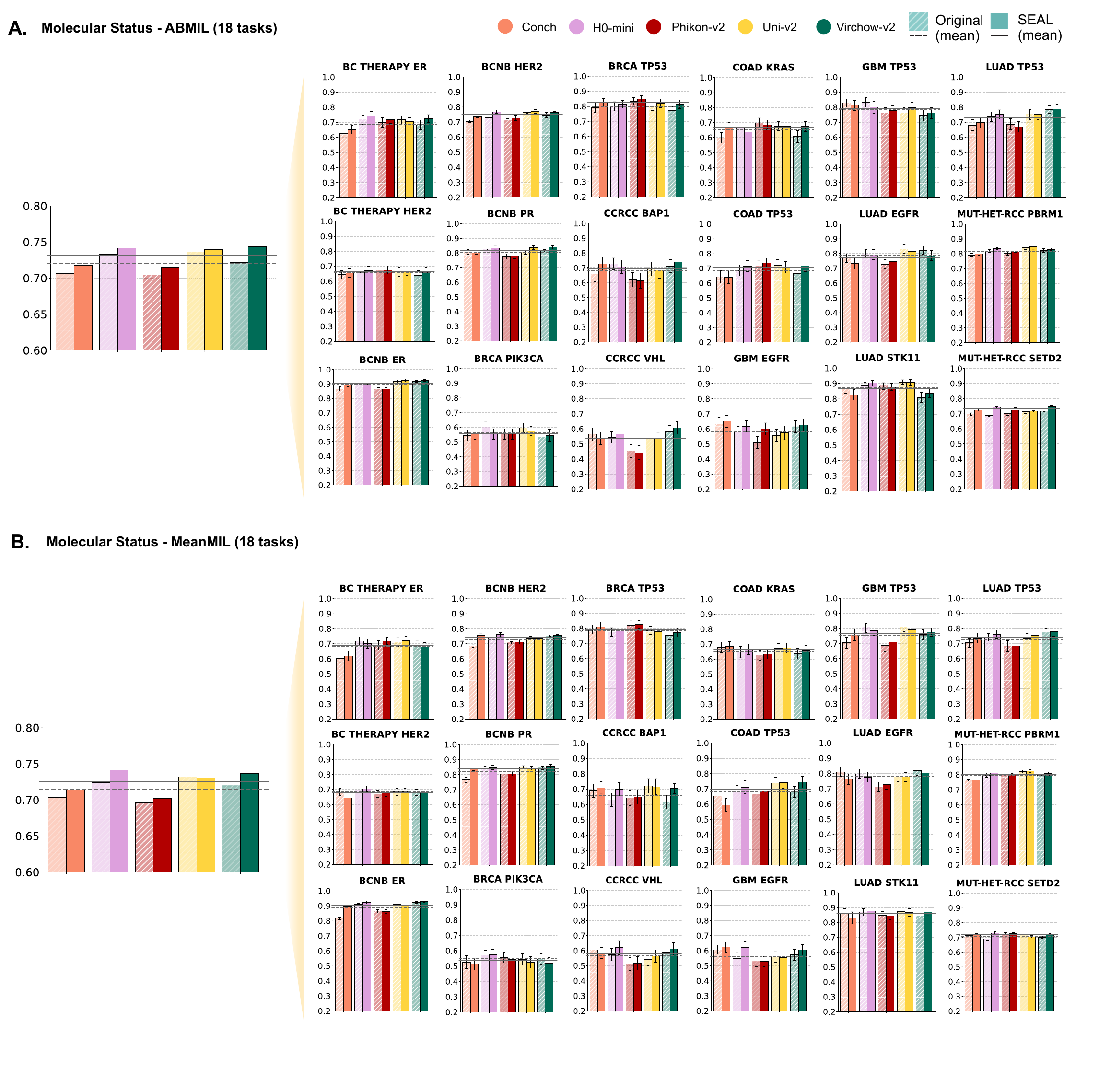}
\caption{\textbf{Comparison between original and $\ours$-finetuned molecular prediction tasks.}
\textbf{(A).} Attention-based MIL (ABMIL) performance for 18 slide-level molecular subtyping tasks. ABMIL was applied on patch embeddings extracted from five vision encoders (Conch, H0-mini, Phikon-v2, Uni-v2, and Virchow-v2) and their $\ours$-finetuned variants. Binary tasks are evaluated using AUC and multi-class classification tasks using balanced accuracy.
\textbf{(B).} Same tasks as (A), replacing ABMIL by mean MIL.
}
\label{edf:molecular_performance}
\end{figure*}

\begin{figure*}
\centering
\includegraphics[width=\textwidth]{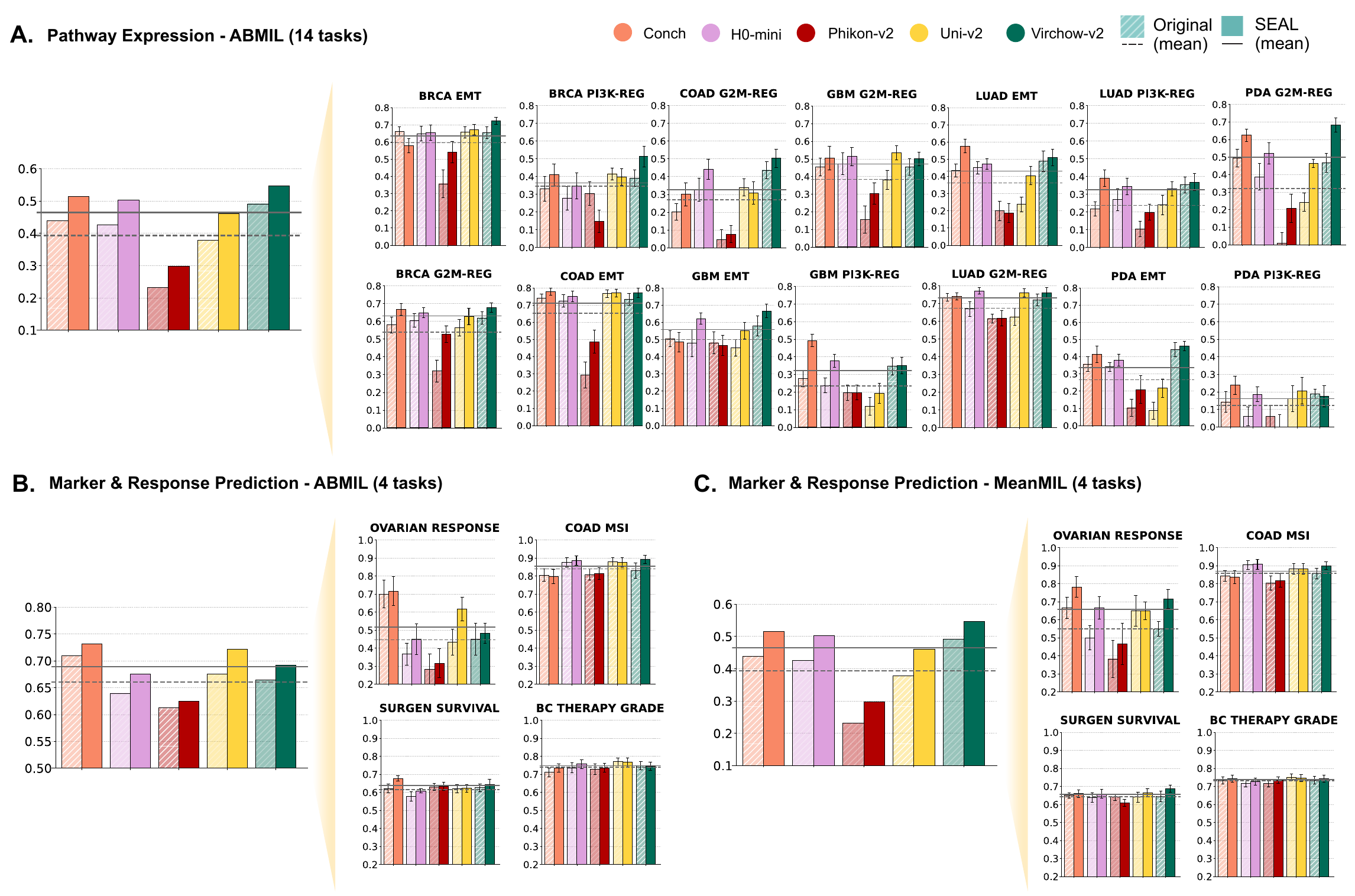}
\caption{\textbf{$\ours$ pathway expression \& treatment response prediction.}
\textbf{(A)}. Slide-level prediction performance of Bulk RNA expression of three MSigDB Hallmark pathways: Genes up-regulated by activation of the PI3K/AKT/mTOR pathway (PI3K-REG), genes involved in the G2/M checkpoint (G2M-REG), and genes defining epithelial-mesenchymal transition (EMT). Performance is measured as the Pearson Correlation Coefficient between predicted pathway expression and measured bulk pathway expression. 
\textbf{(B).} ABMIL results on four slide-level tasks: treatment response prediction of patients with ovarian cancer and treated with Bevacizumab, microsatellite instability prediction in colon adenocarcinoma (COAD), overall survival prediction of colon and rectum adenocarcinoma, and breast cancer grading. 
\textbf{(C).} MeanMIL results on the same set of four slide-level tasks as in \textbf{(B).}.
}
\label{edf:pathway_other_performance}
\end{figure*}

\begin{figure*}
    \centering
    \includegraphics[width=\textwidth]{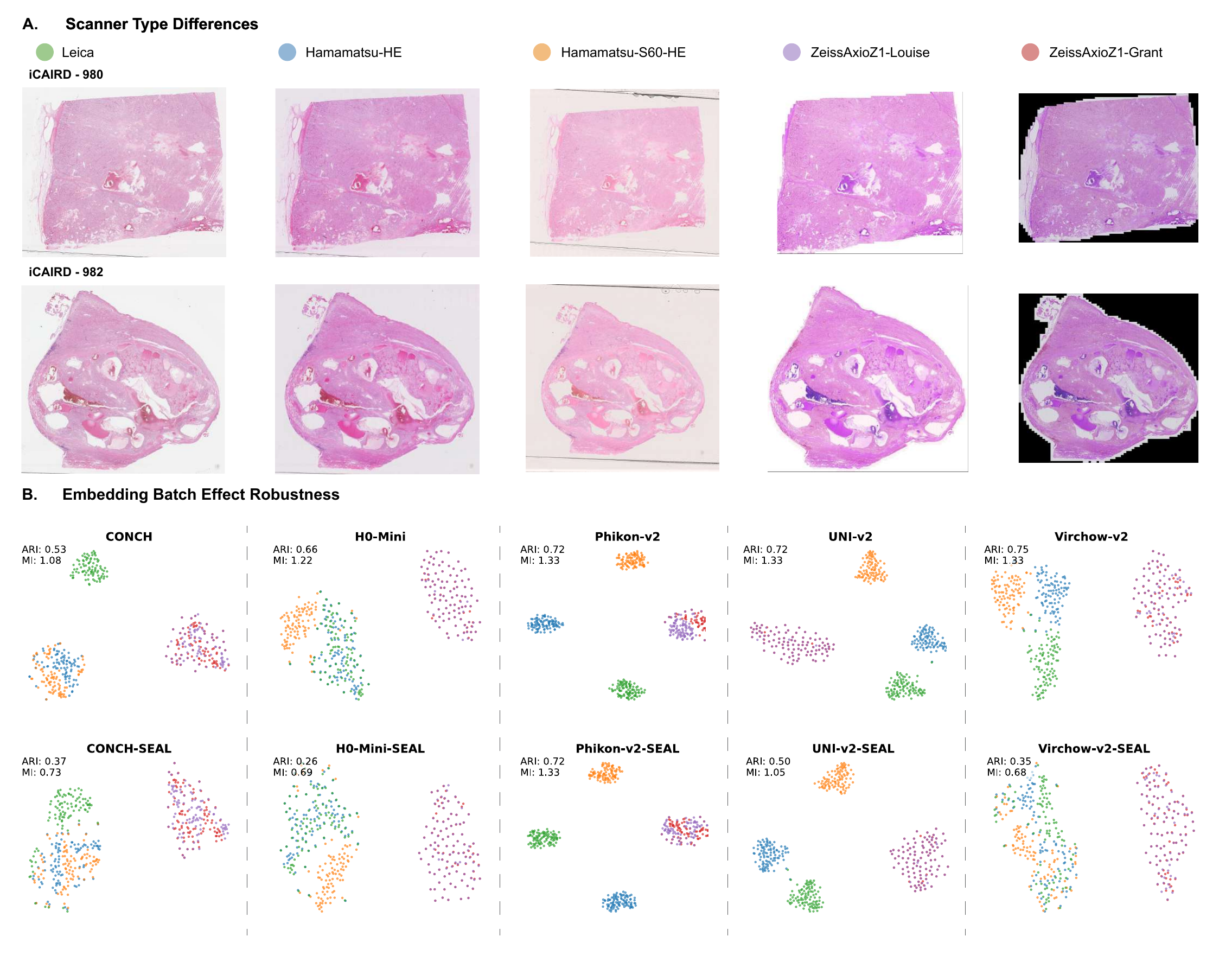}
    \caption{\textbf{Robustness Analysis of SEAL.}
    We analyze to what extent the baseline foundation models and \ours are susceptible to batch effects introduced by differing data acquisition domains of the same Hematoxilin \& Eosin slides. The ICAIRD dataset contains 100 renal cell carcinoma slides captured across five scanners: Hamamatsu-2.0-HE (0.45 $\mu$m/px), Hamamatsu S60 (0.42 $\mu$m/px), Leica Aperio AT2 ($\mu$m/px), Zeiss Axio Z1-Grant (0.22 $\mu$m/px), and Zeiss Axio Z1-Louise (0.22 $\mu$m/px).
    \textbf{(A).} Visual differences of two sample slides in the ICAIRD dataset, exhibiting subtle differences in color, contrast, resolution, and other imaging characteristics across scanners.
    \textbf{(B).} Benchmarking of five vision foundation models (CONCH, H0-Mini, Phikon-v2, UNI-v2, Virchow-v2) and \ours for each of the encoders. The scatter plots show the t-SNE of the mean patch embedding for each slide in the dataset, colour-coded by the slide scanner. Weaker clusters (lower ARI and MI) correspond to lower susceptibility to visual batch effects and consequently higher robustness. ARI: Adjusted Rand Index, MI: Mutual Information.
    }
    \label{edf:robustness}
\end{figure*}

\clearpage

%% file: tables/01-data_hyperparameters.tex
\begin{table}
\centering
\caption{Training, validation, and test dataset overview of the \datasetshort{} dataset by tissue type. The dataset was curated by filtering human Visium Slides from the HEST-1k dataset. The data was split using a stratified 80-10-10 split by patient and organ to ensure that no spots from the same patient were present in the validation or test set. The in-domain test set was further subset based on the support in the training data. }
\label{tbl:data_overview_organ}
\begin{tabular}{l|rr|rr|rr|rr}
\toprule
 & \multicolumn{2}{c}{\textbf{Train}} & \multicolumn{2}{c}{\textbf{Validation}} & \multicolumn{2}{c}{\textbf{Test}} & \multicolumn{2}{c}{\textbf{Total}} \\
 & WSIs & Spots & WSIs & Spots & WSIs & Spots & WSIs & Spots \\
\midrule
Bladder & 5 & 5,200 & 1 & 1,017 & 0 & 0 & 6 & 6,217 \\
Bowel & 55 & 172,901 & 6 & 14,312 & 0 & 0 & 61 & 187,213 \\
Brain & 30 & 100,727 & 4 & 9,291 & 6 & 22,526 & 40 & 132,544 \\
Breast & 5 & 19,685 & 1 & 2,518 & 1 & 4,015 & 7 & 26,218 \\
Cervix & 5 & 7,764 & 0 & 0 & 0 & 0 & 5 & 7,764 \\
Heart & 38 & 95,166 & 4 & 9,674 & 0 & 0 & 42 & 104,840 \\
Kidney & 29 & 36,742 & 3 & 2,105 & 6 & 8,136 & 38 & 46,983 \\
Liver & 11 & 54,905 & 1 & 4,991 & 0 & 0 & 12 & 59,896 \\
Lung & 28 & 56,776 & 3 & 3,777 & 6 & 9,063 & 37 & 69,616 \\
Ovary & 2 & 8,129 & 0 & 0 & 0 & 0 & 2 & 8,129 \\
Pancreas & 4 & 13,001 & 0 & 0 & 0 & 0 & 4 & 13,001 \\
Prostate & 9 & 33,080 & 1 & 4,248 & 2 & 8,179 & 12 & 45,507 \\
Skin & 46 & 41,203 & 5 & 4,606 & 12 & 17,298 & 63 & 63,107 \\
Uterus & 15 & 18,312 & 2 & 3,120 & 0 & 0 & 17 & 21,432 \\
\midrule
Sum & 282 & 663,591 & 31 & 59,659 & 33 & 69,217 & 346 & 792,467 \\
\bottomrule
\end{tabular}
\end{table}

\begin{table}
\centering
\caption{Training, validation, and test dataset overview of the \datasetshort{} dataset by disease state. The dataset was curated by filtering human Visium Slides from the HEST-1k dataset. The data was split using a stratified 80-10-10 split by patient and organ to ensure that no spots from the same patient were present in the validation or test set.}
\label{tbl:data_overview_disease_state}
\begin{tabular}{l|rr|rr|rr|rr}
\toprule
 & \multicolumn{2}{c}{\textbf{\textbf{Train}}} & \multicolumn{2}{c}{\textbf{\textbf{Validation}}} & \multicolumn{2}{c}{\textbf{\textbf{Test}}} & \multicolumn{2}{c}{\textbf{\textbf{Total}}} \\
 & WSIs & Spots & WSIs & Spots & WSIs & Spots & WSIs & Spots \\
\midrule
Cancer & 88 & 277,015 & 13 & 31,386 & 10 & 31,115 & 111 & 339,516 \\
Diseased & 90 & 128,466 & 9 & 11,805 & 12 & 13,892 & 111 & 154,163 \\
Healthy & 88 & 228,191 & 5 & 12,279 & 5 & 15,147 & 98 & 255,617 \\
Treated & 16 & 29,919 & 4 & 4,189 & 6 & 9,063 & 26 & 43,171 \\
\midrule
Sum & 282 & 663,591 & 31 & 59,659 & 33 & 69,217 & 346 & 792,467 \\
\bottomrule
\end{tabular}
\end{table}

{
\renewcommand{\arraystretch}{1.1}
\rowcolors{2}{gray!10}{white}
\begin{table}
\centering
\caption{Source overview of the \datasetshort{} Visium slides used for training of SEAL. All raw were obtained through the HEST-1k library.}
\label{tbl:train_data_sources}
\begin{tabular}{p{12.5cm}|r|r}
\toprule
Dataset & Study Link & Samples used \\
\midrule
Spatial transcriptomics landscape of non-communicable inflammatory skin diseases & \href{https://www.nature.com/articles/s41467-022-35319-w}{Link} & 59 \\
Spatially resolved multiomics of human cardiac niches & \href{https://www.nature.com/articles/s41586-023-06311-1#data-availability}{Link} & 41 \\
COLON MAP: Colon Molecular Atlas Project & \href{https://www.cell.com/cell/fulltext/S0092-8674(23)01222-9?_returnURL=https%3A%2F%2Flinkinghub.elsevier.com%2Fretrieve%2Fpii%2FS0092867423012229%3Fshowall%3Dtrue#secsectitle0020}{Link} & 39 \\
Spatial localization with Spatial Transcriptomics for an atlas of healthy and injured cell states and niches in the human kidney & \href{https://www.ncbi.nlm.nih.gov/pmc/articles/PMC10356613/}{Link} & 23 \\
Spatially resolved transcriptomic profiling of degraded & \href{https://www.nature.com/articles/s41467-023-36071-5}{Link} & 21 \\
A spatially resolved atlas of the human lung characterizes a gland-associated immune niche & \href{https://www.nature.com/articles/s41588-022-01243-4}{Link} & 20 \\
SARS-CoV-2 Niches in Human Placenta Revealed by Spatial Transcriptomics & \href{https://pubmed.ncbi.nlm.nih.gov/37423216/}{Link} & 15 \\
Spatial transcriptomics reveal unresolved wound repair as potential driver of PFA Ependymoma progression & \href{https://academic.oup.com/neuro-oncology/article/24/Supplement_1/i45/6600572}{Link} & 14 \\
spatialLIBD & \href{https://www.nature.com/articles/s41593-020-00787-0#data-availability}{Link} & 12 \\
Multimodal decoding of human liver regeneration & \href{https://www.nature.com/articles/s41586-024-07376-2}{Link} & 8 \\
Genome-wide Spatial Expression Profiling in Formalin-fixed Tissues: Supplementary Data & \href{https://pubmed.ncbi.nlm.nih.gov/36776149/}{Link} & 8 \\
Genome-wide Spatial Expression Profiling in Formalin-fixed Tissues & \href{https://pubmed.ncbi.nlm.nih.gov/36776149/}{Link} & 7 \\
A Spatial Transcriptomic atlas of the human kidney papilla identifies significant immune injury in patients with stone disease & \href{https://www.nature.com/articles/s41467-023-38975-8}{Link} & 7 \\
Prostate ST Internal & N/A (internal) & 5 \\
Spatial Multimodal Analysis: MALDI-MSI and Spatial Transcriptomics within the same tissue section & \href{https://www.nature.com/articles/s41587-023-01937-y}{Link} & 5 \\
Human ileum, Visium & \href{https://pubmed.ncbi.nlm.nih.gov/34845373/}{Link} & 4 \\
Human squamous cell carcinoma & \href{https://www.ncbi.nlm.nih.gov/pmc/articles/PMC7391009/}{Link} & 4 \\
Schwann Cells Shape Tumor Cells and Cancer-Associated Fibroblasts in the Pancreatic Ductal Adenocarcinoma Microenvironment & \href{https://www.nature.com/articles/s41467-023-40314-w}{Link} & 4 \\
Single-cell and spatial transcriptomics characterisation of the immunological landscape in the healthy and PSC human liver & \href{https://pubmed.ncbi.nlm.nih.gov/38199298/}{Link} & 4 \\
Charting the Heterogeneity of Colorectal Cancer Consensus Molecular Subtypes using Spatial Transcriptomics: datasets & \href{https://www.biorxiv.org/content/10.1101/2023.01.23.525135v2}{Link} & 4 \\
Spatiotemporally Deciphering the Unknown Role of Persistent HPV Infection in Precancer to Cervical Cancer Progression: Integrating Single-Cell RNA-Sequencing Landscape and Spatial Transcriptomics Atlas & \href{https://www.ncbi.nlm.nih.gov/pmc/articles/PMC10040725/}{Link} & 4 \\
A single-cell transcriptomic analysis of endometriosis & \href{https://www.nature.com/articles/s41588-022-01254-1}{Link} & 2 \\
COLON MAP: Colon Molecular Atlas Project & \href{https://www.cell.com/cell/fulltext/S0092-8674(21)01381-7?_returnURL=https%3A%2F%2Flinkinghub.elsevier.com%2Fretrieve%2Fpii%2FS0092867421013817%3Fshowall%3Dtrue}{Link} & 2 \\
Batf3-dendritic cells and 4-1BB-4-1BB ligand axis are required at the effector phase within the tumor microenvironment for PD-1-PD-L1 blockade efficacy & \href{https://www.sciencedirect.com/science/article/pii/S2211124724004698?via%3Dihub#app2}{Link} & 2 \\
High resolution mapping of the tumor microenvironment using integrated single-cell, spatial and in situ analysis & \href{https://www.nature.com/articles/s41467-023-43458-x}{Link} & 1 \\
\bottomrule
\end{tabular}
\end{table}
}

\begin{table}
\small
\renewcommand{\arraystretch}{1.4}
\caption{\textbf{Hyperparameter overview by encoder backbone used for training and evaluation}. }
\label{tbl:seal_hyperparameters}
\begin{tabular}{l|rrrrr}
\hline
                               & \multicolumn{1}{l}{\textbf{Conch-SEAL}} & \multicolumn{1}{l}{\textbf{H0-mini-SEAL}} & \multicolumn{1}{l}{\textbf{Phikon-v2-SEAL}} & \multicolumn{1}{l}{\textbf{Uni-v2-SEAL}} & \multicolumn{1}{l}{\textbf{Virchow-v2-SEAL}} \\ \hline
\textbf{Fine-tuning blocks}    & 3                                       & 3                                         & 7                                           & 5                                        & 7                                            \\
\textbf{Optimizer}             & AdamW                                   & AdamW                                     & AdamW                                       & AdamW                                    & AdamW                                        \\
\textbf{LoRA Rank}             & 8                                       & 8                                         & 8                                           & 8                                        & 8                                            \\
\textbf{LoRA  Alpha}           & 8                                       & 8                                         & 8                                           & 8                                        & 8                                            \\
\textbf{LoRA Dropout}          & 0.25                                    & 0.25                                      & 0.25                                        & 0.5                                      & 0.25                                         \\
\textbf{SEAL-Image LR}         & 0.0001                                  & 0.0001                                    & 0.0001                                      & 0.0001                                   & 0.0001                                       \\
\textbf{SEAL-Omics LR}         & 0.0001                                  & 0.0001                                    & 0.0001                                      & 0.0001                                   & 0.0001                                       \\
\textbf{Stage I Warmup LR}     & 0.0005                                  & 0.0005                                    & 0.0005                                      & 0.0005                                   & 0.0005                                       \\
\textbf{$\lambda_{contrast}$}  & 1                                       & 1                                         & 3                                           & 1                                        & 1                                            \\
\textbf{$\lambda_{rec, img}$}  & 1                                       & 1                                         & 3                                           & 3                                        & 1                                            \\
\textbf{$\lambda_{rec, gene}$} & 1                                       & 1                                         & 1                                           & 1                                        & 1                                            \\
\textbf{$\lambda_{da}$} & 0.001                                       & 0.001                                         & 0.001                                           & 0.001                                        & 0.001                                            \\
\textbf{Auxiliary projection}  & linear                                  & linear                                    & none                                        & linear                                   & none                                         \\
\textbf{Embedding Dims}        & 512                                     & 1536                                      & 1024                                        & 1536                                     & 2560                                         \\
\textbf{Scheduler}             & CosAnneal                               & CosAnneal                                 & CosAnneal                                   & CosAnneal                                & CosAnneal                                    \\
\textbf{Batch Size}            & 384                                     & 384                                       & 300                                         & 150                                      & 384                                          \\
\textbf{VAE Norm Flow}         & True                                    & True                                      & True                                        & True                                     & True                                         \\
\textbf{VAE Dropout}           & 0                                       & 0                                         & 0                                           & 0                                        & 0                                            \\
\textbf{Decoder Dropout}       & 0                                       & 0                                         & 0                                           & 0.3                                      & 0                                            \\
\textbf{Layer Decay}           & 0.7                                     & 0.7                                       & 0.7                                         & 0.7                                      & 0.7                                          \\
\textbf{Temperature}           & 0.05                                    & 0.05                                      & 0.05                                        & 0.05                                     & 0.05                                         \\
\textbf{Local smoothing}       & True                                    & True                                      & True                                        & True                                     & True                                         \\
\textbf{Weight decay}          & 0.2                                     & 0.2                                       & 0.2                                         & 0.2                                      & 0.2                                          \\
\textbf{Gene Panel Size}       & 2000                                    & 2000                                      & 2000                                        & 2000                                     & 2000                                         \\
\textbf{Warmup Epochs}         & 3                                       & 3                                         & 3                                           & 3                                        & 3                                            \\
\textbf{Linear Probe}          & ridge                                   & ridge                                     & ridge                                       & ridge                                    & ridge                                       
\end{tabular}
\end{table}

\begin{table}
\small
\renewcommand{\arraystretch}{1.3}
\caption{\textbf{Vision encoders fine-tuned using \ours}. Encoders were selected to reflect a diverse range of model backbones (size and patch size), pre-training data distributions, and self-supervised learning (SSL) methods.}
\label{tbl:vision_encoder_overview}
\begin{tabular}{l|llrllll}
\hline
\multicolumn{1}{l|}{\textbf{Encoder}}                                               & \textbf{Backbone} & \textbf{Pre-training} & \multicolumn{1}{l}{\textbf{\# Params}} & \textbf{Dim} & \textbf{\# Tiles} & \textbf{\# Slides} & \textbf{Top 3 Organs}                        \\ \hline
\textbf{Conch~\cite{lu_Visuallanguage_2024CONCH}}            & ViT-B/16       & Contrastive   & 90 M                                       & 512                    & 1.7 M  & n/a             & Gastro, Soft tissue, Lung       \\
\textbf{H0-Mini~\cite{filiot_Distilling_2025hoptimus-mini}}  & ViT-B/14       & Distillation & 86 M                                       & 1536                   & 43 M                           & 6093               & n/a                                          \\
\textbf{Phikon-v2~\cite{filiot2024phikon}}                     & ViT-L/16      & Dino-v2               & 307 M                                      & 1024                   & 460 M                          & 58359              & Breast, Lung, Colorectal      \\
\textbf{Uni-v2-h~\cite{chen_Generalpurpose_2024UNI}}         & ViT-H/14       & Dino-v2               & 681 M                                      & 1536                   & 100 M                          & 100000             & Heart, Lung, Kidney      \\
\textbf{Virchow-v2~\cite{zimmermann_Virchow2_2024Virchow2a}} & ViT-H/16       & Dino-v2               & 632 M                                      & 2560                   & 2000 M                         & 3100000            & Breast, Lymph Node, Skin
\end{tabular}
\end{table}

\begin{table}
\renewcommand{\arraystretch}{1.3}
\caption{\textbf{Image augmentation transforms applied during Stage II training}. The augmentations refer to the Albumentations implementations of standard geometric, color jitter, and blurring augmentations, applied at different likelihoods for each sample (p). The "one of group" implies if only one of these implementations is applied for each sample,each at varying probabilities.}
\label{tbl:hyperparams_augmentation}
\begin{tabular}{l|rlr}
\hline
\textbf{Method}                      & \textbf{Likelihood (p)} & \textbf{Arguments}                         & \textbf{One of Group} \\ \hline
\textbf{Resized Crop}                & 1.0                                 & Crop scale=(0.8, 1.0)                      &                       \\
\textbf{Horizontal Flip}             & 0.5                                 &                                            &                       \\
\textbf{Vertical Flip}               & 0.25                                &                                            &                       \\
\textbf{Image Compression}           & 1.0                                 & Quality range=(50,100)                     &                       \\
\textbf{Hue/Saturation/Value Jitter} & 1.0                                 & hue=(-40, 40), sat=(-20, 20), val=(-30,30) & A                     \\
\textbf{RGB Shift}                   & 1.0                                 & r=(-30, 30), g=(-40, 20), b=(-30, 30)      & A                     \\
\textbf{Brightness/Contrast Shift}   & 1.0                                 & b=0.1, c=0.1                               & A                     \\
\textbf{Channel Shuffle}             & 0.5                                 &                                            & A                     \\
\textbf{Gaussian Blur}               & 0.1                                 & blur\_lim=(9,9), sigma\_lim=(0.1,2)        &                      
\end{tabular}
\end{table}

%% file: tables/02-maple_test_results.tex
\begin{table}[htbp]
    \centering
    \footnotesize
    \caption{\textbf{Comparison of 5 Vision-Only vs. SEAL fine-tuned encoders} on the \textbf{Prostate} Maple-test set. Results show a linear probe using PCA (256 factors) + Ridge regression for the top 10 highly variable genes (HVGs) across Maple-train prostate samples for each encoder. Reported results show the mean and standard deviation across 5 probing folds and p-values calculated across the folds. HVGs: IGHM, IGHG3, CD226, CTSE, GNAL, RHOJ, IGLC1, MPP7, SCD5, GLTP}
    \label{tab:prostate_results}
    \adjustbox{width=\textwidth,center}{
    \begin{tabular}{lrrrrrr}
        \toprule
        \textbf{Method} & \textbf{PCC ($\uparrow$) (std.)} & \textbf{PCC \%} & \textbf{MSE ($\downarrow$) (std.)} & \textbf{MSE \%} & \textbf{PCC folds} & \textbf{p-value} \\
        \midrule
        CONCH & \makecell{0.386 \\ {\scriptsize $\pm$ 0.002}} & - & \makecell{0.854 \\ {\scriptsize $\pm$ 0.004}} & - & [0.3857, 0.385, 0.388, 0.3861, 0.3866] & - \\
        CONCH-SEAL & \makecell{0.420 \\ {\scriptsize $\pm$ 0.003}} & +8.81\% & \makecell{0.826 \\ {\scriptsize $\pm$ 0.005}} & -3.28\% & [0.421, 0.4208, 0.4206, 0.4199, 0.4184] & $<$0.001 \\
        \midrule
        H0MINI & \makecell{0.443 \\ {\scriptsize $\pm$ 0.002}} & - & \makecell{0.805 \\ {\scriptsize $\pm$ 0.003}} & - & [0.4436, 0.4428, 0.4433, 0.4421, 0.4419] & - \\
        H0MINI-SEAL & \makecell{0.480 \\ {\scriptsize $\pm$ 0.003}} & +8.35\% & \makecell{0.766 \\ {\scriptsize $\pm$ 0.005}} & -4.84\% & [0.4795, 0.4791, 0.4791, 0.4823, 0.4805] & $<$0.001 \\
        \midrule
        PHIKON2 & \makecell{0.444 \\ {\scriptsize $\pm$ 0.002}} & - & \makecell{0.808 \\ {\scriptsize $\pm$ 0.004}} & - & [0.4436, 0.4446, 0.4435, 0.443, 0.444] & - \\
        PHIKON2-SEAL & \makecell{0.506 \\ {\scriptsize $\pm$ 0.002}} & +13.96\% & \makecell{0.743 \\ {\scriptsize $\pm$ 0.004}} & -8.04\% & [0.5056, 0.5088, 0.5058, 0.5039, 0.5056] & $<$0.001 \\
        \midrule
        UNIv2 & \makecell{0.485 \\ {\scriptsize $\pm$ 0.002}} & - & \makecell{0.764 \\ {\scriptsize $\pm$ 0.004}} & - & [0.483, 0.4846, 0.484, 0.486, 0.4859] & - \\
        UNIv2-SEAL & \makecell{0.548 \\ {\scriptsize $\pm$ 0.002}} & +12.99\% & \makecell{0.689 \\ {\scriptsize $\pm$ 0.004}} & -9.82\% & [0.5457, 0.5503, 0.5472, 0.5478, 0.5485] & $<$0.001 \\
        \midrule
        VIRCHOW2 & \makecell{0.483 \\ {\scriptsize $\pm$ 0.002}} & - & \makecell{0.764 \\ {\scriptsize $\pm$ 0.003}} & - & [0.4832, 0.4821, 0.4834, 0.4838, 0.4833] & - \\
        VIRCHOW2-SEAL & \makecell{0.576 \\ {\scriptsize $\pm$ 0.002}} & +19.25\% & \makecell{0.658 \\ {\scriptsize $\pm$ 0.003}} & -13.87\% & [0.5752, 0.5752, 0.5744, 0.5765, 0.5772] & $<$0.001 \\
        \bottomrule
    \end{tabular}
    }
\end{table}

\begin{table}[htbp]
    \centering
    \footnotesize
    \caption{\textbf{Comparison of 5 Vision-Only vs. SEAL fine-tuned encoders} on the \textbf{Brain} Maple-test set. Results show a linear probe using PCA (256 factors) + Ridge regression for the top 10 highly variable genes (HVGs) across Maple-train brain samples for each encoder. Reported results show the mean and standard deviation across 5 probing folds and p-values calculated across the folds. HVGs: ARHGEF28, TMEM117, RAD51B, SOX7, SCEL, MROH7, TRIM62, PRDM5, FGD6, LRRC10B}
    \label{tab:brain_results}
    \adjustbox{width=\textwidth,center}{
    \begin{tabular}{lrrrrrr}
        \toprule
        \textbf{Method} & \textbf{PCC ($\uparrow$) (std.)} & \textbf{PCC \%} & \textbf{MSE ($\downarrow$) (std.)} & \textbf{MSE \%} & \textbf{PCC folds} & \textbf{p-value} \\
        \midrule
        CONCH & \makecell{0.316 \\ {\scriptsize $\pm$ 0.003}} & - & \makecell{0.176 \\ {\scriptsize $\pm$ 0.002}} & - & [0.3135, 0.3164, 0.3154, 0.3156, 0.3167] & - \\
        CONCH-SEAL & \makecell{0.351 \\ {\scriptsize $\pm$ 0.003}} & +11.08\% & \makecell{0.172 \\ {\scriptsize $\pm$ 0.002}} & -2.27\% & [0.3499, 0.3502, 0.3516, 0.3507, 0.3524] & $<$0.001 \\
        \midrule
        H0MINI & \makecell{0.363 \\ {\scriptsize $\pm$ 0.003}} & - & \makecell{0.170 \\ {\scriptsize $\pm$ 0.002}} & - & [0.3628, 0.3618, 0.3635, 0.3643, 0.3608] & - \\
        H0MINI-SEAL & \makecell{0.393 \\ {\scriptsize $\pm$ 0.003}} & +8.26\% & \makecell{0.166 \\ {\scriptsize $\pm$ 0.002}} & -2.35\% & [0.3922, 0.3932, 0.3937, 0.3932, 0.3928] & $<$0.001 \\
        \midrule
        PHIKON2 & \makecell{0.361 \\ {\scriptsize $\pm$ 0.003}} & - & \makecell{0.170 \\ {\scriptsize $\pm$ 0.002}} & - & [0.3614, 0.362, 0.3605, 0.3596, 0.3609] & - \\
        PHIKON2-SEAL & \makecell{0.408 \\ {\scriptsize $\pm$ 0.003}} & +13.02\% & \makecell{0.164 \\ {\scriptsize $\pm$ 0.002}} & -3.53\% & [0.4061, 0.4101, 0.4087, 0.4081, 0.407] & $<$0.001 \\
        \midrule
        UNIv2 & \makecell{0.396 \\ {\scriptsize $\pm$ 0.003}} & - & \makecell{0.166 \\ {\scriptsize $\pm$ 0.001}} & - & [0.3945, 0.3967, 0.3949, 0.3983, 0.3959] & - \\
        UNIv2-SEAL & \makecell{0.439 \\ {\scriptsize $\pm$ 0.004}} & +10.86\% & \makecell{0.159 \\ {\scriptsize $\pm$ 0.002}} & -4.22\% & [0.4374, 0.44, 0.4373, 0.4396, 0.4398] & $<$0.001 \\
        \midrule
        VIRCHOW2 & \makecell{0.388 \\ {\scriptsize $\pm$ 0.003}} & - & \makecell{0.167 \\ {\scriptsize $\pm$ 0.002}} & - & [0.3889, 0.3867, 0.3878, 0.3882, 0.3889] & - \\
        VIRCHOW2-SEAL & \makecell{0.460 \\ {\scriptsize $\pm$ 0.003}} & +18.56\% & \makecell{0.157 \\ {\scriptsize $\pm$ 0.001}} & -5.99\% & [0.4579, 0.4593, 0.4596, 0.4609, 0.4608] & $<$0.001 \\
        \bottomrule
    \end{tabular}
    }
\end{table}

\begin{table}[htbp]
    \centering
    \footnotesize
    \caption{\textbf{Comparison of 5 Vision-Only vs. SEAL fine-tuned encoders} on the \textbf{Breast} Maple-test set. Results show a linear probe using PCA (256 factors) + Ridge regression for the top 10 highly variable genes (HVGs) across Maple-train breast samples for each encoder. Reported results show the mean and standard deviation across 5 probing folds and p-values calculated across the folds. HVGs: PLP1, PTPRZ1, SLC22A4, PDCD1LG2, IGHA2, GPM6A, ANO8, TGDS, PCDH9, PSTPIP1}
    \label{tab:breast_results}
    \adjustbox{width=\textwidth,center}{
    \begin{tabular}{lrrrrrr}
        \toprule
        \textbf{Method} & \textbf{PCC ($\uparrow$) (std.)} & \textbf{PCC \%} & \textbf{MSE ($\downarrow$) (std.)} & \textbf{MSE \%} & \textbf{PCC folds} & \textbf{p-value} \\
        \midrule
        CONCH & \makecell{0.436 \\ {\scriptsize $\pm$ 0.002}} & - & \makecell{0.373 \\ {\scriptsize $\pm$ 0.002}} & - & [0.4358, 0.4356, 0.4355, 0.4358, 0.4358] & - \\
        CONCH-SEAL & \makecell{0.466 \\ {\scriptsize $\pm$ 0.002}} & +6.88\% & \makecell{0.361 \\ {\scriptsize $\pm$ 0.002}} & -3.22\% & [0.466, 0.467, 0.4672, 0.4668, 0.4654] & $<$0.001 \\
        \midrule
        H0MINI & \makecell{0.474 \\ {\scriptsize $\pm$ 0.003}} & - & \makecell{0.353 \\ {\scriptsize $\pm$ 0.002}} & - & [0.476, 0.474, 0.4733, 0.4734, 0.473] & - \\
        H0MINI-SEAL & \makecell{0.509 \\ {\scriptsize $\pm$ 0.002}} & +7.38\% & \makecell{0.332 \\ {\scriptsize $\pm$ 0.002}} & -5.95\% & [0.5088, 0.5088, 0.5084, 0.5112, 0.5087] & $<$0.001 \\
        \midrule
        PHIKON2 & \makecell{0.477 \\ {\scriptsize $\pm$ 0.002}} & - & \makecell{0.351 \\ {\scriptsize $\pm$ 0.002}} & - & [0.4756, 0.476, 0.4775, 0.4771, 0.4785] & - \\
        PHIKON2-SEAL & \makecell{0.528 \\ {\scriptsize $\pm$ 0.002}} & +10.69\% & \makecell{0.317 \\ {\scriptsize $\pm$ 0.002}} & -9.69\% & [0.5276, 0.5282, 0.5279, 0.5278, 0.5275] & $<$0.001 \\
        \midrule
        UNIv2 & \makecell{0.506 \\ {\scriptsize $\pm$ 0.003}} & - & \makecell{0.328 \\ {\scriptsize $\pm$ 0.002}} & - & [0.5053, 0.5085, 0.5058, 0.5065, 0.5057] & - \\
        UNIv2-SEAL & \makecell{0.557 \\ {\scriptsize $\pm$ 0.002}} & +10.08\% & \makecell{0.300 \\ {\scriptsize $\pm$ 0.002}} & -8.54\% & [0.5562, 0.5585, 0.5568, 0.5565, 0.5586] & $<$0.001 \\
        \midrule
        VIRCHOW2 & \makecell{0.501 \\ {\scriptsize $\pm$ 0.002}} & - & \makecell{0.333 \\ {\scriptsize $\pm$ 0.002}} & - & [0.5003, 0.5005, 0.5022, 0.5019, 0.5008] & - \\
        VIRCHOW2-SEAL & \makecell{0.581 \\ {\scriptsize $\pm$ 0.002}} & +15.97\% & \makecell{0.281 \\ {\scriptsize $\pm$ 0.002}} & -15.62\% & [0.5815, 0.5806, 0.5808, 0.581, 0.581] & $<$0.001 \\
        \bottomrule
    \end{tabular}
    }
\end{table}

\begin{table}[htbp]
    \centering
    \footnotesize
    \caption{\textbf{Comparison of 5 Vision-Only vs. SEAL fine-tuned encoders} on the \textbf{Kidney} Maple-test set. Results show a linear probe using PCA (256 factors) + Ridge regression for the top 10 highly variable genes (HVGs) across Maple-train kidney samples for each encoder. Reported results show the mean and standard deviation across 5 probing folds and p-values calculated across the folds. HVGs: CD84, ABI3BP, NLRP1, MNDA, ZNF549, SLC44A4, IGHM, TRAF7, SCNN1B, PAICS}
    \label{tab:kidney_results}
    \adjustbox{width=\textwidth,center}{
    \begin{tabular}{lrrrrrr}
        \toprule
        \textbf{Method} & \textbf{PCC ($\uparrow$) (std.)} & \textbf{PCC \%} & \textbf{MSE ($\downarrow$) (std.)} & \textbf{MSE \%} & \textbf{PCC folds} & \textbf{p-value} \\
        \midrule
        CONCH & \makecell{0.399 \\ {\scriptsize $\pm$ 0.002}} & - & \makecell{0.652 \\ {\scriptsize $\pm$ 0.003}} & - & [0.3992, 0.3989, 0.3997, 0.3995, 0.3996] & - \\
        CONCH-SEAL & \makecell{0.432 \\ {\scriptsize $\pm$ 0.002}} & +8.27\% & \makecell{0.633 \\ {\scriptsize $\pm$ 0.004}} & -2.91\% & [0.4319, 0.4318, 0.4322, 0.4326, 0.432] & $<$0.001 \\
        \midrule
        H0MINI & \makecell{0.449 \\ {\scriptsize $\pm$ 0.002}} & - & \makecell{0.619 \\ {\scriptsize $\pm$ 0.003}} & - & [0.4494, 0.4487, 0.4487, 0.4486, 0.4497] & - \\
        H0MINI-SEAL & \makecell{0.480 \\ {\scriptsize $\pm$ 0.002}} & +6.90\% & \makecell{0.598 \\ {\scriptsize $\pm$ 0.003}} & -3.39\% & [0.4792, 0.4815, 0.4777, 0.4805, 0.4794] & $<$0.001 \\
        \midrule
        PHIKON2 & \makecell{0.445 \\ {\scriptsize $\pm$ 0.002}} & - & \makecell{0.624 \\ {\scriptsize $\pm$ 0.004}} & - & [0.444, 0.4454, 0.4446, 0.4446, 0.4445] & - \\
        PHIKON2-SEAL & \makecell{0.498 \\ {\scriptsize $\pm$ 0.002}} & +11.91\% & \makecell{0.585 \\ {\scriptsize $\pm$ 0.003}} & -6.25\% & [0.4982, 0.4986, 0.497, 0.499, 0.4973] & $<$0.001 \\
        \midrule
        UNIv2 & \makecell{0.483 \\ {\scriptsize $\pm$ 0.002}} & - & \makecell{0.599 \\ {\scriptsize $\pm$ 0.004}} & - & [0.4816, 0.4829, 0.4839, 0.4828, 0.4823] & - \\
        UNIv2-SEAL & \makecell{0.536 \\ {\scriptsize $\pm$ 0.002}} & +10.97\% & \makecell{0.557 \\ {\scriptsize $\pm$ 0.002}} & -7.01\% & [0.5349, 0.536, 0.5351, 0.5383, 0.5366] & $<$0.001 \\
        \midrule
        VIRCHOW2 & \makecell{0.476 \\ {\scriptsize $\pm$ 0.002}} & - & \makecell{0.602 \\ {\scriptsize $\pm$ 0.003}} & - & [0.476, 0.474, 0.4769, 0.476, 0.4774] & - \\
        VIRCHOW2-SEAL & \makecell{0.556 \\ {\scriptsize $\pm$ 0.002}} & +16.81\% & \makecell{0.541 \\ {\scriptsize $\pm$ 0.003}} & -10.13\% & [0.5563, 0.557, 0.5561, 0.5549, 0.5562] & $<$0.001 \\
        \bottomrule
    \end{tabular}
    }
\end{table}

\begin{table}[htbp]
    \centering
    \footnotesize
    \caption{\textbf{Comparison of 5 Vision-Only vs. SEAL fine-tuned encoders} on the \textbf{Skin} Maple-test set. Results show a linear probe using PCA (256 factors) + Ridge regression for the top 10 highly variable genes (HVGs) across Maple-train skin samples for each encoder. Reported results show the mean and standard deviation across 5 probing folds and p-values calculated across the folds. HVGs: TMCC3, LRRC75A, SLAIN1, RTN4RL2, RHOH, COL4A5, KIFC1, CYP3A5, ENPEP, SPRYD7}
    \label{tab:skin_results}
    \adjustbox{width=\textwidth,center}{
    \begin{tabular}{lrrrrrr}
        \toprule
        \textbf{Method} & \textbf{PCC ($\uparrow$) (std.)} & \textbf{PCC \%} & \textbf{MSE ($\downarrow$) (std.)} & \textbf{MSE \%} & \textbf{PCC folds} & \textbf{p-value} \\
        \midrule
        CONCH & \makecell{0.427 \\ {\scriptsize $\pm$ 0.002}} & - & \makecell{0.343 \\ {\scriptsize $\pm$ 0.002}} & - & [0.4262, 0.4271, 0.4277, 0.4281, 0.427] & - \\
        CONCH-SEAL & \makecell{0.455 \\ {\scriptsize $\pm$ 0.002}} & +6.56\% & \makecell{0.335 \\ {\scriptsize $\pm$ 0.002}} & -2.33\% & [0.4542, 0.4553, 0.4553, 0.4558, 0.4532] & $<$0.001 \\
        \midrule
        H0MINI & \makecell{0.475 \\ {\scriptsize $\pm$ 0.002}} & - & \makecell{0.328 \\ {\scriptsize $\pm$ 0.002}} & - & [0.4745, 0.4747, 0.474, 0.4766, 0.4746] & - \\
        H0MINI-SEAL & \makecell{0.506 \\ {\scriptsize $\pm$ 0.002}} & +6.53\% & \makecell{0.317 \\ {\scriptsize $\pm$ 0.002}} & -3.35\% & [0.5066, 0.5067, 0.505, 0.5059, 0.5039] & $<$0.001 \\
        \midrule
        PHIKON2 & \makecell{0.470 \\ {\scriptsize $\pm$ 0.003}} & - & \makecell{0.329 \\ {\scriptsize $\pm$ 0.002}} & - & [0.471, 0.4697, 0.4692, 0.4685, 0.4699] & - \\
        PHIKON2-SEAL & \makecell{0.521 \\ {\scriptsize $\pm$ 0.003}} & +10.85\% & \makecell{0.311 \\ {\scriptsize $\pm$ 0.002}} & -5.47\% & [0.5207, 0.5212, 0.5206, 0.5197, 0.5211] & $<$0.001 \\
        \midrule
        UNIv2 & \makecell{0.515 \\ {\scriptsize $\pm$ 0.002}} & - & \makecell{0.312 \\ {\scriptsize $\pm$ 0.002}} & - & [0.5128, 0.5148, 0.5146, 0.515, 0.5156] & - \\
        UNIv2-SEAL & \makecell{0.549 \\ {\scriptsize $\pm$ 0.003}} & +6.60\% & \makecell{0.299 \\ {\scriptsize $\pm$ 0.002}} & -4.17\% & [0.5497, 0.55, 0.5487, 0.549, 0.5489] & $<$0.001 \\
        \midrule
        VIRCHOW2 & \makecell{0.499 \\ {\scriptsize $\pm$ 0.002}} & - & \makecell{0.318 \\ {\scriptsize $\pm$ 0.002}} & - & [0.4978, 0.4987, 0.4986, 0.498, 0.5001] & - \\
        VIRCHOW2-SEAL & \makecell{0.575 \\ {\scriptsize $\pm$ 0.003}} & +15.23\% & \makecell{0.291 \\ {\scriptsize $\pm$ 0.002}} & -8.49\% & [0.5754, 0.5759, 0.5742, 0.5749, 0.5748] & $<$0.001 \\
        \bottomrule
    \end{tabular}
    }
\end{table}

\begin{table}[htbp]
    \centering
    \footnotesize
    \caption{\textbf{Comparison of 5 Vision-Only vs. SEAL fine-tuned encoders} on the \textbf{Lung} Maple-test set. Results show a linear probe using PCA (256 factors) + Ridge regression for the top 10 highly variable genes (HVGs) across Maple-train lung samples for each encoder. Reported results show the mean and standard deviation across 5 probing folds and p-values calculated across the folds. HVGs: ANKRD42, ADRA2C, URB2, TMEM200B, ARL13B, ISLR, RSF1, RBP4, PDGFRA, LCP1}
    \label{tab:lung_results}
    \adjustbox{width=\textwidth,center}{
    \begin{tabular}{lrrrrrr}
        \toprule
        \textbf{Method} & \textbf{PCC ($\uparrow$) (std.)} & \textbf{PCC \%} & \textbf{MSE ($\downarrow$) (std.)} & \textbf{MSE \%} & \textbf{PCC folds} & \textbf{p-value} \\
        \midrule
        CONCH & \makecell{0.366 \\ {\scriptsize $\pm$ 0.002}} & - & \makecell{0.575 \\ {\scriptsize $\pm$ 0.002}} & - & [0.3646, 0.3657, 0.3649, 0.3676, 0.3649] & - \\
        CONCH-SEAL & \makecell{0.400 \\ {\scriptsize $\pm$ 0.002}} & +9.29\% & \makecell{0.555 \\ {\scriptsize $\pm$ 0.002}} & -3.48\% & [0.3992, 0.399, 0.4003, 0.4007, 0.4021] & $<$0.001 \\
        \midrule
        H0MINI & \makecell{0.416 \\ {\scriptsize $\pm$ 0.002}} & - & \makecell{0.539 \\ {\scriptsize $\pm$ 0.003}} & - & [0.4164, 0.4157, 0.4157, 0.4159, 0.4161] & - \\
        H0MINI-SEAL & \makecell{0.443 \\ {\scriptsize $\pm$ 0.002}} & +6.49\% & \makecell{0.517 \\ {\scriptsize $\pm$ 0.003}} & -4.08\% & [0.4439, 0.4442, 0.4427, 0.4437, 0.4426] & $<$0.001 \\
        \midrule
        PHIKON2 & \makecell{0.414 \\ {\scriptsize $\pm$ 0.002}} & - & \makecell{0.543 \\ {\scriptsize $\pm$ 0.003}} & - & [0.4151, 0.4145, 0.4133, 0.4144, 0.4135] & - \\
        PHIKON2-SEAL & \makecell{0.463 \\ {\scriptsize $\pm$ 0.003}} & +11.84\% & \makecell{0.507 \\ {\scriptsize $\pm$ 0.003}} & -6.63\% & [0.4627, 0.465, 0.4641, 0.4622, 0.4628] & $<$0.001 \\
        \midrule
        UNIv2 & \makecell{0.449 \\ {\scriptsize $\pm$ 0.002}} & - & \makecell{0.515 \\ {\scriptsize $\pm$ 0.003}} & - & [0.4491, 0.4474, 0.45, 0.448, 0.4479] & - \\
        UNIv2-SEAL & \makecell{0.496 \\ {\scriptsize $\pm$ 0.003}} & +10.47\% & \makecell{0.479 \\ {\scriptsize $\pm$ 0.002}} & -6.99\% & [0.4944, 0.4981, 0.4944, 0.4983, 0.4963] & $<$0.001 \\
        \midrule
        VIRCHOW2 & \makecell{0.439 \\ {\scriptsize $\pm$ 0.002}} & - & \makecell{0.525 \\ {\scriptsize $\pm$ 0.002}} & - & [0.4399, 0.4381, 0.4388, 0.4394, 0.4391] & - \\
        VIRCHOW2-SEAL & \makecell{0.518 \\ {\scriptsize $\pm$ 0.002}} & +18.00\% & \makecell{0.462 \\ {\scriptsize $\pm$ 0.002}} & -12.00\% & [0.5185, 0.5181, 0.5182, 0.5194, 0.5174] & $<$0.001 \\
        \bottomrule
    \end{tabular}
    }
\end{table}

%% file: tables/03-hestbench_results.tex
\begin{table}[ht]
\centering
\footnotesize
\caption{Full HestBench results using a Ridge regression linear probe of the baseline pathology patch encoders and the finetuned version using SEAL.}
\label{tbl:hestbench}
\begin{tabular}{l|ccccccccc|c}
\toprule
\textbf{Encoder} & \textbf{IDC} & \textbf{PRAD} & \textbf{PAAD} & \textbf{SKCM} & \textbf{COAD} & \textbf{READ} & \textbf{CCRCC} & \textbf{LUAD} & \textbf{LYMPH IDC} & \textbf{Average} \\
\midrule
Conch & \makecell{0.5440 \\ {{\scriptsize $\pm$ 0.0750}}} & \makecell{0.3540 \\ {{\scriptsize $\pm$ 0.0040}}} & \makecell{0.4450 \\ {{\scriptsize $\pm$ 0.0590}}} & \makecell{0.6030 \\ {{\scriptsize $\pm$ 0.0500}}} & \makecell{0.2660 \\ {{\scriptsize $\pm$ 0.0040}}} & \makecell{0.1660 \\ {{\scriptsize $\pm$ 0.0480}}} & \makecell{0.2050 \\ {{\scriptsize $\pm$ 0.0300}}} & \makecell{0.5240 \\ {{\scriptsize $\pm$ 0.0230}}} & \makecell{0.2530 \\ {{\scriptsize $\pm$ 0.0460}}} & \makecell{0.3733 \\ {{\scriptsize $\pm$ 0.0377}}} \\
+SEAL & \makecell{0.5480 \\ {{\scriptsize $\pm$ 0.0720}}} & \makecell{0.3670 \\ {{\scriptsize $\pm$ 0.0090}}} & \makecell{0.4390 \\ {{\scriptsize $\pm$ 0.0490}}} & \makecell{0.5930 \\ {{\scriptsize $\pm$ 0.0420}}} & \makecell{0.2850 \\ {{\scriptsize $\pm$ 0.0070}}} & \makecell{0.1610 \\ {{\scriptsize $\pm$ 0.0560}}} & \makecell{0.2040 \\ {{\scriptsize $\pm$ 0.0360}}} & \makecell{0.5340 \\ {{\scriptsize $\pm$ 0.0210}}} & \makecell{0.2540 \\ {{\scriptsize $\pm$ 0.0450}}} & \makecell{0.3761 \\ {{\scriptsize $\pm$ 0.0374}}} \\ \midrule
H0mini & \makecell{0.5900 \\ {{\scriptsize $\pm$ 0.0730}}} & \makecell{0.3590 \\ {{\scriptsize $\pm$ 0.0370}}} & \makecell{0.5050 \\ {{\scriptsize $\pm$ 0.0470}}} & \makecell{0.6190 \\ {{\scriptsize $\pm$ 0.0600}}} & \makecell{0.2630 \\ {{\scriptsize $\pm$ 0.0270}}} & \makecell{0.2030 \\ {{\scriptsize $\pm$ 0.0550}}} & \makecell{0.2640 \\ {{\scriptsize $\pm$ 0.0440}}} & \makecell{0.5650 \\ {{\scriptsize $\pm$ 0.0320}}} & \makecell{0.2650 \\ {{\scriptsize $\pm$ 0.0350}}} & \makecell{0.4037 \\ {{\scriptsize $\pm$ 0.0456}}} \\
+SEAL & \makecell{0.5920 \\ {{\scriptsize $\pm$ 0.0790}}} & \makecell{0.3840 \\ {{\scriptsize $\pm$ 0.0170}}} & \makecell{0.5090 \\ {{\scriptsize $\pm$ 0.0500}}} & \makecell{0.6080 \\ {{\scriptsize $\pm$ 0.0710}}} & \makecell{0.3010 \\ {{\scriptsize $\pm$ 0.0220}}} & \makecell{0.1980 \\ {{\scriptsize $\pm$ 0.0580}}} & \makecell{0.2650 \\ {{\scriptsize $\pm$ 0.0430}}} & \makecell{0.5590 \\ {{\scriptsize $\pm$ 0.0270}}} & \makecell{0.2660 \\ {{\scriptsize $\pm$ 0.0340}}} & \makecell{0.4091 \\ {{\scriptsize $\pm$ 0.0446}}} \\ \midrule
Virchow2 & \makecell{0.5950 \\ {{\scriptsize $\pm$ 0.0850}}} & \makecell{0.3530 \\ {{\scriptsize $\pm$ 0.0380}}} & \makecell{0.4760 \\ {{\scriptsize $\pm$ 0.0680}}} & \makecell{0.6430 \\ {{\scriptsize $\pm$ 0.0330}}} & \makecell{0.2580 \\ {{\scriptsize $\pm$ 0.0330}}} & \makecell{0.1990 \\ {{\scriptsize $\pm$ 0.0610}}} & \makecell{0.2700 \\ {{\scriptsize $\pm$ 0.0510}}} & \makecell{0.5730 \\ {{\scriptsize $\pm$ 0.0150}}} & \makecell{0.2580 \\ {{\scriptsize $\pm$ 0.0340}}} & \makecell{0.4028 \\ {{\scriptsize $\pm$ 0.0464}}} \\
+SEAL & \makecell{0.5960 \\ {{\scriptsize $\pm$ 0.0860}}} & \makecell{0.3610 \\ {{\scriptsize $\pm$ 0.0430}}} & \makecell{0.4870 \\ {{\scriptsize $\pm$ 0.0610}}} & \makecell{0.6320 \\ {{\scriptsize $\pm$ 0.0450}}} & \makecell{0.2690 \\ {{\scriptsize $\pm$ 0.0300}}} & \makecell{0.2020 \\ {{\scriptsize $\pm$ 0.0610}}} & \makecell{0.2770 \\ {{\scriptsize $\pm$ 0.0720}}} & \makecell{0.5690 \\ {{\scriptsize $\pm$ 0.0230}}} & \makecell{0.2610 \\ {{\scriptsize $\pm$ 0.0280}}} & \makecell{0.4060 \\ {{\scriptsize $\pm$ 0.0499}}} \\ \midrule
Univ2 & \makecell{0.5900 \\ {{\scriptsize $\pm$ 0.0810}}} & \makecell{0.3570 \\ {{\scriptsize $\pm$ 0.0490}}} & \makecell{0.5000 \\ {{\scriptsize $\pm$ 0.0400}}} & \makecell{0.6610 \\ {{\scriptsize $\pm$ 0.0150}}} & \makecell{0.3020 \\ {{\scriptsize $\pm$ 0.0040}}} & \makecell{0.2220 \\ {{\scriptsize $\pm$ 0.0380}}} & \makecell{0.2640 \\ {{\scriptsize $\pm$ 0.0470}}} & \makecell{0.5590 \\ {{\scriptsize $\pm$ 0.0130}}} & \makecell{0.2730 \\ {{\scriptsize $\pm$ 0.0400}}} & \makecell{0.4142 \\ {{\scriptsize $\pm$ 0.0363}}} \\
+SEAL & \makecell{0.5850 \\ {{\scriptsize $\pm$ 0.0710}}} & \makecell{0.3650 \\ {{\scriptsize $\pm$ 0.0080}}} & \makecell{0.5210 \\ {{\scriptsize $\pm$ 0.0650}}} & \makecell{0.6580 \\ {{\scriptsize $\pm$ 0.0470}}} & \makecell{0.3190 \\ {{\scriptsize $\pm$ 0.0110}}} & \makecell{0.2160 \\ {{\scriptsize $\pm$ 0.0420}}} & \makecell{0.2640 \\ {{\scriptsize $\pm$ 0.0330}}} & \makecell{0.5730 \\ {{\scriptsize $\pm$ 0.0020}}} & \makecell{0.2660 \\ {{\scriptsize $\pm$ 0.0360}}} & \makecell{0.4186 \\ {{\scriptsize $\pm$ 0.0350}}} \\ \midrule
Gigapath & \makecell{0.5520 \\ {{\scriptsize $\pm$ 0.0730}}} & \makecell{0.3760 \\ {{\scriptsize $\pm$ 0.0240}}} & \makecell{0.4770 \\ {{\scriptsize $\pm$ 0.0500}}} & \makecell{0.5570 \\ {{\scriptsize $\pm$ 0.0670}}} & \makecell{0.3000 \\ {{\scriptsize $\pm$ 0.0240}}} & \makecell{0.1910 \\ {{\scriptsize $\pm$ 0.0660}}} & \makecell{0.2410 \\ {{\scriptsize $\pm$ 0.0330}}} & \makecell{0.5430 \\ {{\scriptsize $\pm$ 0.0340}}} & \makecell{0.2500 \\ {{\scriptsize $\pm$ 0.0550}}} & \makecell{0.3874 \\ {{\scriptsize $\pm$ 0.0473}}} \\
+SEAL & \makecell{0.5640 \\ {{\scriptsize $\pm$ 0.0740}}} & \makecell{0.3480 \\ {{\scriptsize $\pm$ 0.0300}}} & \makecell{0.4880 \\ {{\scriptsize $\pm$ 0.0620}}} & \makecell{0.5790 \\ {{\scriptsize $\pm$ 0.1200}}} & \makecell{0.2940 \\ {{\scriptsize $\pm$ 0.0430}}} & \makecell{0.2000 \\ {{\scriptsize $\pm$ 0.0510}}} & \makecell{0.2610 \\ {{\scriptsize $\pm$ 0.0310}}} & \makecell{0.5430 \\ {{\scriptsize $\pm$ 0.0390}}} & \makecell{0.2590 \\ {{\scriptsize $\pm$ 0.0340}}} & \makecell{0.3929 \\ {{\scriptsize $\pm$ 0.0538}}} \\
\bottomrule
\end{tabular}
\end{table}

%% file: tables/04-data_efficiency.tex
\begin{table}\centering
\caption{\textbf{Training data efficiency of patch-level tasks} measured in \textbf{Pearson Correlation Coefficient (PCC)} of the \textbf{expression prediction} task on different subsets of the training data for \textbf{Virchow-v2-SEAL}. All training was done for the same number of SEAL training iterations, such that models trained on 1\% of training data experienced the same number of backpropagation steps as those trained on 100\%. 0\% data represents the vision-only baseline, 100\% the final SEAL model.}
\label{tbl:patch_Virchow-v2_data_efficiency}
\begin{tabular}{ll|rrrr}
\toprule
 & \textbf{Training Portion} & \textbf{0\%} & \textbf{1\%} & \textbf{10\%} & \textbf{100\%} \\
\midrule
\multirow[t]{2}{*}{\textbf{Brain}} & {Absolute} & $0.388 \pm 0.003$ & $0.414 \pm 0.010$ & $0.417 \pm 0.010$ & $0.460 \pm 0.003$ \\
 & {Uplift (\%)} & 0.000 & 6.701 & 7.474 & 18.557 \\
\cline{1-6}
\multirow[t]{2}{*}{\textbf{Breast}} & {Absolute} & $0.501 \pm 0.002$ & $0.533 \pm 0.007$ & $0.537 \pm 0.007$ & $0.581 \pm 0.002$ \\
 & {Uplift (\%)} & 0.000 & 6.387 & 7.186 & 15.968 \\
\cline{1-6}
\multirow[t]{2}{*}{\textbf{Kidney}} & {Absolute} & $0.476 \pm 0.002$ & $0.563 \pm 0.006$ & $0.567 \pm 0.006$ & $0.556 \pm 0.002$ \\
 & {Uplift (\%)} & 0.000 & 18.277 & 19.118 & 16.807 \\
\cline{1-6}
\multirow[t]{2}{*}{\textbf{Lung}} & {Absolute} & $0.439 \pm 0.002$ & $0.448 \pm 0.007$ & $0.450 \pm 0.007$ & $0.518 \pm 0.002$ \\
 & {Uplift (\%)} & 0.000 & 2.050 & 2.506 & 17.995 \\
\cline{1-6}
\multirow[t]{2}{*}{\textbf{Prostate}} & {Absolute} & $0.483 \pm 0.002$ & $0.570 \pm 0.007$ & $0.577 \pm 0.008$ & $0.576 \pm 0.002$ \\
 & {Uplift (\%)} & 0.000 & 18.012 & 19.462 & 19.255 \\
\cline{1-6}
\multirow[t]{2}{*}{\textbf{Skin}} & {Absolute} & $0.499 \pm 0.002$ & $0.545 \pm 0.007$ & $0.548 \pm 0.007$ & $0.575 \pm 0.003$ \\
 & {Uplift (\%)} & 0.000 & 9.218 & 9.820 & 15.230 \\
\cline{1-6}
\bottomrule
\end{tabular}
\end{table}

\begin{table}\centering
\caption{\textbf{Training data efficiency of slide-level tasks} measured in \textbf{ROC Area Under Curve (AUC)} of the \textbf{mutation status prediction} task on different subsets of the training data for \textbf{Virchow-v2-SEAL}. All training was done for the same number of SEAL training iterations, such that models trained on 1\% of training data experienced the same number of backpropagation steps as those trained on 100\%. 0\% data represents the vision-only baseline, 100\% the final SEAL model.}
\label{tbl:slide_Virchow-v2_data_efficiency}
\begin{tabular}{ll|rrrr}
\toprule
 & \textbf{Training Portion} & \textbf{0\%} & \textbf{1\%} & \textbf{10\%} & \textbf{100\%} \\
\midrule
\multirow[t]{2}{*}{\textbf{BCNB ER}} & {Absolute} & $0.918 \pm 0.017$ & $0.922 \pm 0.026$ & $0.925 \pm 0.022$ & $0.925 \pm 0.023$ \\
 & {Uplift (\%)} & 0.000 & 0.436 & 0.763 & 0.763 \\
\cline{1-6}
\multirow[t]{2}{*}{\textbf{BCNB HER2}} & {Absolute} & $0.746 \pm 0.054$ & $0.751 \pm 0.025$ & $0.760 \pm 0.024$ & $0.765 \pm 0.015$ \\
 & {Uplift (\%)} & 0.000 & 0.670 & 1.877 & 2.547 \\
\cline{1-6}
\multirow[t]{2}{*}{\textbf{BCNB PR}} & {Absolute} & $0.816 \pm 0.028$ & $0.828 \pm 0.026$ & $0.839 \pm 0.046$ & $0.839 \pm 0.034$ \\
 & {Uplift (\%)} & 0.000 & 1.471 & 2.819 & 2.819 \\
\cline{1-6}
\multirow[t]{2}{*}{\textbf{CPTAC COAD KRAS }} & {Absolute} & $0.606 \pm 0.122$ & $0.621 \pm 0.108$ & $0.655 \pm 0.120$ & $0.673 \pm 0.109$ \\
 & {Uplift (\%)} & 0.000 & 2.475 & 8.086 & 11.056 \\
\cline{1-6}
\multirow[t]{2}{*}{\textbf{CPTAC COAD TP53 }} & {Absolute} & $0.664 \pm 0.136$ & $0.610 \pm 0.131$ & $0.636 \pm 0.120$ & $0.717 \pm 0.127$ \\
 & {Uplift (\%)} & 0.000 & -8.133 & -4.217 & 7.982 \\
\cline{1-6}
\bottomrule
\end{tabular}
\end{table}

\begin{table}\centering
\caption{\textbf{Training data efficiency of patch-level tasks} measured in \textbf{Pearson Correlation Coefficient (PCC)} of the \textbf{expression prediction} task on different subsets of the training data for \textbf{Uni-v2-SEAL}. All training was done for the same number of SEAL training iterations, such that models trained on 1\% of training data experienced the same number of backpropagation steps as those trained on 100\%. 0\% data represents the vision-only baseline, 100\% the final SEAL model.}
\label{tbl:patch_Uni-v2_data_efficiency}
\begin{tabular}{ll|rrrr}
\toprule
 & \textbf{Training Portion} & \textbf{0\%} & \textbf{1\%} & \textbf{10\%} & \textbf{100\%} \\
\midrule
\multirow[t]{2}{*}{\textbf{Brain}} & {Absolute} & $0.396 \pm 0.003$ & $0.432 \pm 0.003$ & $0.415 \pm 0.009$ & $0.439 \pm 0.004$ \\
 & {Uplift (\%)} & 0.000 & 9.091 & 4.798 & 10.859 \\
\cline{1-6}
\multirow[t]{2}{*}{\textbf{Breast}} & {Absolute} & $0.506 \pm 0.003$ & $0.555 \pm 0.003$ & $0.535 \pm 0.007$ & $0.557 \pm 0.002$ \\
 & {Uplift (\%)} & 0.000 & 9.684 & 5.731 & 10.079 \\
\cline{1-6}
\multirow[t]{2}{*}{\textbf{Kidney}} & {Absolute} & $0.483 \pm 0.002$ & $0.531 \pm 0.002$ & $0.567 \pm 0.007$ & $0.536 \pm 0.002$ \\
 & {Uplift (\%)} & 0.000 & 9.938 & 17.391 & 10.973 \\
\cline{1-6}
\multirow[t]{2}{*}{\textbf{Lung}} & {Absolute} & $0.449 \pm 0.002$ & $0.491 \pm 0.002$ & $0.450 \pm 0.008$ & $0.496 \pm 0.003$ \\
 & {Uplift (\%)} & 0.000 & 9.354 & 0.223 & 10.468 \\
\cline{1-6}
\multirow[t]{2}{*}{\textbf{Prostate}} & {Absolute} & $0.485 \pm 0.002$ & $0.546 \pm 0.002$ & $0.576 \pm 0.007$ & $0.548 \pm 0.002$ \\
 & {Uplift (\%)} & 0.000 & 12.577 & 18.763 & 12.990 \\
\cline{1-6}
\multirow[t]{2}{*}{\textbf{Skin}} & {Absolute} & $0.515 \pm 0.002$ & $0.549 \pm 0.003$ & $0.547 \pm 0.008$ & $0.549 \pm 0.003$ \\
 & {Uplift (\%)} & 0.000 & 6.602 & 6.214 & 6.602 \\
\cline{1-6}
\bottomrule
\end{tabular}
\end{table}

\begin{table}\centering
\caption{\textbf{Training data efficiency of slide-level tasks} measured in \textbf{ROC Area Under Curve (AUC)} of the \textbf{mutation status prediction} task on different subsets of the training data for \textbf{Uni-v2-SEAL}. All training was done for the same number of SEAL training iterations, such that models trained on 1\% of training data experienced the same number of backpropagation steps as those trained on 100\%. 0\% data represents the vision-only baseline, 100\% the final SEAL model.}
\label{tbl:slide_Uni-v2_data_efficiency}
\begin{tabular}{ll|rrrr}
\toprule
 & \textbf{Training Portion} & \textbf{0\%} & \textbf{1\%} & \textbf{10\%} & \textbf{100\%} \\
\midrule
\multirow[t]{2}{*}{\textbf{BCNB ER}} & {Absolute} & $0.917 \pm 0.036$ & $0.907 \pm 0.023$ & $0.912 \pm 0.031$ & $0.926 \pm 0.030$ \\
 & {Uplift (\%)} & 0.000 & -1.091 & -0.545 & 0.981 \\
\cline{1-6}
\multirow[t]{2}{*}{\textbf{BCNB HER2}} & {Absolute} & $0.765 \pm 0.026$ & $0.758 \pm 0.040$ & $0.765 \pm 0.060$ & $0.771 \pm 0.038$ \\
 & {Uplift (\%)} & 0.000 & -0.915 & 0.000 & 0.784 \\
\cline{1-6}
\multirow[t]{2}{*}{\textbf{BCNB PR}} & {Absolute} & $0.806 \pm 0.044$ & $0.813 \pm 0.029$ & $0.820 \pm 0.030$ & $0.836 \pm 0.040$ \\
 & {Uplift (\%)} & 0.000 & 0.868 & 1.737 & 3.722 \\
\cline{1-6}
\multirow[t]{2}{*}{\textbf{CPTAC COAD KRAS }} & {Absolute} & $0.676 \pm 0.097$ & $0.674 \pm 0.105$ & $0.661 \pm 0.097$ & $0.675 \pm 0.126$ \\
 & {Uplift (\%)} & 0.000 & -0.296 & -2.219 & -0.148 \\
\cline{1-6}
\multirow[t]{2}{*}{\textbf{CPTAC COAD TP53 }} & {Absolute} & $0.720 \pm 0.122$ & $0.683 \pm 0.109$ & $0.725 \pm 0.109$ & $0.708 \pm 0.124$ \\
 & {Uplift (\%)} & 0.000 & -5.139 & 0.694 & -1.667 \\
\cline{1-6}
\bottomrule
\end{tabular}
\end{table}

%% file: tables/05-slide_level_results.tex
\begin{table}[htbp]
    \centering
    \scriptsize
    \setlength{\abovecaptionskip}{3pt}
    \setlength{\belowcaptionskip}{-9pt}
    \renewcommand{\arraystretch}{0.9}
    \caption{Performance comparison between SEAL fine-tuned encoders and vision-only baselines on the \textbf{BRCA neoadjuvant therapy response} dataset, showing the \textbf{ER status classification} task (n=166). This is a slide-level task evaluated using \textbf{AUC} across 50 cross-validation folds. The best-performing model for each encoder backbone in \textbf{bold}.}
    \label{tab:bc_therapy_er_status_results}
    \adjustbox{width=\textwidth,center}{
    \begin{tabular}{lrrrrrrr}
        \toprule
 & \multicolumn{3}{c}{\textbf{ABMIL}} & \multicolumn{3}{c}{\textbf{MeanMIL}} \\
\cmidrule(lr){2-4} \cmidrule(lr){5-7}
\textbf{Model} & \textbf{AUC (↑)} & \textbf{\% change} & \textbf{Sig.} & \textbf{AUC (↑)} & \textbf{\% change} & \textbf{Sig.} \\
        \midrule
CONCH & 0.626 {\scriptsize $\pm$ 0.091} & - & - & 0.602 {\scriptsize $\pm$ 0.094} & - & - \\
+SEAL & \textbf{0.651 {\scriptsize $\pm$ 0.092}} & +4.0\% & * & \textbf{0.619 {\scriptsize $\pm$ 0.097}} & +2.8\% & - \\
        \midrule
H0MINI & 0.716 {\scriptsize $\pm$ 0.097} & - & - & \textbf{0.717 {\scriptsize $\pm$ 0.092}} & - & - \\
+SEAL & \textbf{0.742 {\scriptsize $\pm$ 0.090}} & +3.6\% & * & 0.701 {\scriptsize $\pm$ 0.098} & -2.2\% & - \\
        \midrule
PHIKON2 & 0.698 {\scriptsize $\pm$ 0.100} & - & - & 0.690 {\scriptsize $\pm$ 0.096} & - & - \\
+SEAL & \textbf{0.718 {\scriptsize $\pm$ 0.081}} & +2.9\% & * & \textbf{0.718 {\scriptsize $\pm$ 0.076}} & +4.1\% & - \\
        \midrule
UNIv2 & \textbf{0.717 {\scriptsize $\pm$ 0.086}} & - & - & 0.712 {\scriptsize $\pm$ 0.091} & - & - \\
+SEAL & 0.704 {\scriptsize $\pm$ 0.087} & -1.8\% & - & \textbf{0.720 {\scriptsize $\pm$ 0.091}} & +1.1\% & * \\
        \midrule
VIRCHOW2 & 0.682 {\scriptsize $\pm$ 0.099} & - & - & \textbf{0.690 {\scriptsize $\pm$ 0.091}} & - & - \\
+SEAL & \textbf{0.724 {\scriptsize $\pm$ 0.085}} & +6.2\% & ** & 0.680 {\scriptsize $\pm$ 0.091} & -1.4\% & - \\
        \midrule
        \midrule[\heavyrulewidth]
Average & 0.688 {\scriptsize $\pm$ 0.095} & - & - & 0.682 {\scriptsize $\pm$ 0.093} & - & - \\
+SEAL & \textbf{0.708 {\scriptsize $\pm$ 0.087}} & +2.9\% & - & \textbf{0.688 {\scriptsize $\pm$ 0.091}} & +0.8\% & - \\
        \bottomrule
    \end{tabular}}
\end{table}

\begin{table}[htbp]
    \centering
    \scriptsize
    \setlength{\abovecaptionskip}{3pt}
    \setlength{\belowcaptionskip}{-9pt}
    \renewcommand{\arraystretch}{0.9}
    \caption{Performance comparison between SEAL fine-tuned encoders and vision-only baselines on the \textbf{BRCA neoadjuvant therapy response} dataset, showing the \textbf{HER2 status classification} task (n=166). This is a slide-level task evaluated using \textbf{AUC} across 50 cross-validation folds. The best-performing model for each encoder backbone in \textbf{bold}.}
    \label{tab:bc_therapy_her2_status_results}
    \adjustbox{width=\textwidth,center}{
    \begin{tabular}{lrrrrrrr}
        \toprule
 & \multicolumn{3}{c}{\textbf{ABMIL}} & \multicolumn{3}{c}{\textbf{MeanMIL}} \\
\cmidrule(lr){2-4} \cmidrule(lr){5-7}
\textbf{Model} & \textbf{AUC (↑)} & \textbf{\% change} & \textbf{Sig.} & \textbf{AUC (↑)} & \textbf{\% change} & \textbf{Sig.} \\
        \midrule
CONCH & 0.645 {\scriptsize $\pm$ 0.087} & - & - & \textbf{0.683 {\scriptsize $\pm$ 0.074}} & - & - \\
+SEAL & \textbf{0.656 {\scriptsize $\pm$ 0.091}} & +1.7\% & - & 0.643 {\scriptsize $\pm$ 0.086} & -5.9\% & - \\
        \midrule
H0MINI & 0.658 {\scriptsize $\pm$ 0.091} & - & - & 0.697 {\scriptsize $\pm$ 0.063} & - & - \\
+SEAL & \textbf{0.671 {\scriptsize $\pm$ 0.088}} & +2.0\% & - & \textbf{0.704 {\scriptsize $\pm$ 0.062}} & +1.0\% & * \\
        \midrule
PHIKON2 & 0.675 {\scriptsize $\pm$ 0.086} & - & - & 0.669 {\scriptsize $\pm$ 0.072} & - & - \\
+SEAL & \textbf{0.676 {\scriptsize $\pm$ 0.087}} & +0.1\% & - & \textbf{0.676 {\scriptsize $\pm$ 0.072}} & +1.0\% & * \\
        \midrule
UNIv2 & 0.663 {\scriptsize $\pm$ 0.092} & - & - & 0.684 {\scriptsize $\pm$ 0.073} & - & - \\
+SEAL & \textbf{0.667 {\scriptsize $\pm$ 0.083}} & +0.6\% & - & 0.684 {\scriptsize $\pm$ 0.072} & +0.0\% & - \\
        \midrule
VIRCHOW2 & 0.639 {\scriptsize $\pm$ 0.103} & - & - & \textbf{0.679 {\scriptsize $\pm$ 0.065}} & - & - \\
+SEAL & \textbf{0.660 {\scriptsize $\pm$ 0.094}} & +3.3\% & - & 0.677 {\scriptsize $\pm$ 0.067} & -0.3\% & - \\
        \midrule
        \midrule[\heavyrulewidth]
Average & 0.656 {\scriptsize $\pm$ 0.092} & - & - & \textbf{0.682 {\scriptsize $\pm$ 0.069}} & - & - \\
+SEAL & \textbf{0.666 {\scriptsize $\pm$ 0.089}} & +1.5\% & - & 0.677 {\scriptsize $\pm$ 0.072} & -0.8\% & - \\
        \bottomrule
    \end{tabular}}
\end{table}

\begin{table}[htbp]
    \centering
    \scriptsize
    \setlength{\abovecaptionskip}{3pt}
    \setlength{\belowcaptionskip}{-9pt}
    \renewcommand{\arraystretch}{0.9}
    \caption{Performance comparison between SEAL fine-tuned encoders and vision-only baselines on the \textbf{BCNB} dataset, showing the \textbf{ER status classification} task (n=1058). This is a slide-level task evaluated using \textbf{AUC} across 5 cross-validation folds. The best-performing model for each encoder backbone in \textbf{bold}.}
    \label{tab:bcnb_er_results}
    \adjustbox{width=\textwidth,center}{
    \begin{tabular}{lrrrrrrr}
        \toprule
 & \multicolumn{3}{c}{\textbf{ABMIL}} & \multicolumn{3}{c}{\textbf{MeanMIL}} \\
\cmidrule(lr){2-4} \cmidrule(lr){5-7}
\textbf{Model} & \textbf{AUC (↑)} & \textbf{\% change} & \textbf{Sig.} & \textbf{AUC (↑)} & \textbf{\% change} & \textbf{Sig.} \\
        \midrule
CONCH & 0.866 {\scriptsize $\pm$ 0.045} & - & - & 0.816 {\scriptsize $\pm$ 0.029} & - & - \\
+SEAL & \textbf{0.891 {\scriptsize $\pm$ 0.032}} & +2.9\% & - & \textbf{0.893 {\scriptsize $\pm$ 0.030}} & +9.4\% & - \\
        \midrule
H0MINI & \textbf{0.909 {\scriptsize $\pm$ 0.036}} & - & - & 0.910 {\scriptsize $\pm$ 0.022} & - & - \\
+SEAL & 0.896 {\scriptsize $\pm$ 0.038} & -1.4\% & - & \textbf{0.923 {\scriptsize $\pm$ 0.030}} & +1.4\% & - \\
        \midrule
PHIKON2 & 0.864 {\scriptsize $\pm$ 0.037} & - & - & \textbf{0.864 {\scriptsize $\pm$ 0.035}} & - & - \\
+SEAL & \textbf{0.866 {\scriptsize $\pm$ 0.024}} & +0.2\% & - & 0.862 {\scriptsize $\pm$ 0.038} & -0.2\% & - \\
        \midrule
UNIv2 & 0.917 {\scriptsize $\pm$ 0.036} & - & - & \textbf{0.910 {\scriptsize $\pm$ 0.027}} & - & - \\
+SEAL & \textbf{0.926 {\scriptsize $\pm$ 0.030}} & +1.0\% & - & 0.905 {\scriptsize $\pm$ 0.029} & -0.5\% & - \\
        \midrule
VIRCHOW2 & 0.918 {\scriptsize $\pm$ 0.017} & - & - & 0.924 {\scriptsize $\pm$ 0.021} & - & - \\
+SEAL & \textbf{0.925 {\scriptsize $\pm$ 0.023}} & +0.8\% & - & \textbf{0.930 {\scriptsize $\pm$ 0.023}} & +0.6\% & - \\
        \midrule
        \midrule[\heavyrulewidth]
Average & 0.895 {\scriptsize $\pm$ 0.034} & - & - & 0.885 {\scriptsize $\pm$ 0.027} & - & - \\
+SEAL & \textbf{0.901 {\scriptsize $\pm$ 0.029}} & +0.7\% & - & \textbf{0.903 {\scriptsize $\pm$ 0.030}} & +2.0\% & - \\
        \bottomrule
    \end{tabular}}
\end{table}

\begin{table}[htbp]
    \centering
    \scriptsize
    \setlength{\abovecaptionskip}{3pt}
    \setlength{\belowcaptionskip}{-9pt}
    \renewcommand{\arraystretch}{0.9}
    \caption{Performance comparison between SEAL fine-tuned encoders and vision-only baselines on the \textbf{BCNB} dataset, showing the \textbf{HER2 status classification} task (n=1058). This is a slide-level task evaluated using \textbf{AUC} across 5 cross-validation folds. The best-performing model for each encoder backbone in \textbf{bold}.}
    \label{tab:bcnb_her2_results}
    \adjustbox{width=\textwidth,center}{
    \begin{tabular}{lrrrrrrr}
        \toprule
 & \multicolumn{3}{c}{\textbf{ABMIL}} & \multicolumn{3}{c}{\textbf{MeanMIL}} \\
\cmidrule(lr){2-4} \cmidrule(lr){5-7}
\textbf{Model} & \textbf{AUC (↑)} & \textbf{\% change} & \textbf{Sig.} & \textbf{AUC (↑)} & \textbf{\% change} & \textbf{Sig.} \\
        \midrule
CONCH & 0.704 {\scriptsize $\pm$ 0.024} & - & - & 0.685 {\scriptsize $\pm$ 0.026} & - & - \\
+SEAL & \textbf{0.734 {\scriptsize $\pm$ 0.021}} & +4.3\% & * & \textbf{0.756 {\scriptsize $\pm$ 0.033}} & +10.4\% & ** \\
        \midrule
H0MINI & 0.730 {\scriptsize $\pm$ 0.052} & - & - & 0.740 {\scriptsize $\pm$ 0.035} & - & - \\
+SEAL & \textbf{0.768 {\scriptsize $\pm$ 0.034}} & +5.2\% & * & \textbf{0.761 {\scriptsize $\pm$ 0.043}} & +2.8\% & - \\
        \midrule
PHIKON2 & 0.712 {\scriptsize $\pm$ 0.037} & - & - & 0.706 {\scriptsize $\pm$ 0.028} & - & - \\
+SEAL & \textbf{0.727 {\scriptsize $\pm$ 0.054}} & +2.1\% & - & \textbf{0.711 {\scriptsize $\pm$ 0.038}} & +0.7\% & - \\
        \midrule
UNIv2 & 0.765 {\scriptsize $\pm$ 0.026} & - & - & \textbf{0.736 {\scriptsize $\pm$ 0.037}} & - & - \\
+SEAL & \textbf{0.771 {\scriptsize $\pm$ 0.038}} & +0.8\% & - & 0.735 {\scriptsize $\pm$ 0.030} & -0.1\% & - \\
        \midrule
VIRCHOW2 & 0.746 {\scriptsize $\pm$ 0.054} & - & - & 0.753 {\scriptsize $\pm$ 0.016} & - & - \\
+SEAL & \textbf{0.765 {\scriptsize $\pm$ 0.015}} & +2.5\% & - & \textbf{0.754 {\scriptsize $\pm$ 0.028}} & +0.1\% & - \\
        \midrule
        \midrule[\heavyrulewidth]
Average & 0.731 {\scriptsize $\pm$ 0.039} & - & - & 0.724 {\scriptsize $\pm$ 0.028} & - & - \\
+SEAL & \textbf{0.753 {\scriptsize $\pm$ 0.032}} & +3.0\% & - & \textbf{0.743 {\scriptsize $\pm$ 0.034}} & +2.7\% & - \\
        \bottomrule
    \end{tabular}}
\end{table}

\begin{table}[htbp]
    \centering
    \scriptsize
    \setlength{\abovecaptionskip}{3pt}
    \setlength{\belowcaptionskip}{-9pt}
    \renewcommand{\arraystretch}{0.9}
    \caption{Performance comparison between SEAL fine-tuned encoders and vision-only baselines on the \textbf{BCNB} dataset, showing the \textbf{PR status classification} task (n=1058). This is a slide-level task evaluated using \textbf{AUC} across 5 cross-validation folds. The best-performing model for each encoder backbone in \textbf{bold}.}
    \label{tab:bcnb_pr_results}
    \adjustbox{width=\textwidth,center}{
    \begin{tabular}{lrrrrrrr}
        \toprule
 & \multicolumn{3}{c}{\textbf{ABMIL}} & \multicolumn{3}{c}{\textbf{MeanMIL}} \\
\cmidrule(lr){2-4} \cmidrule(lr){5-7}
\textbf{Model} & \textbf{AUC (↑)} & \textbf{\% change} & \textbf{Sig.} & \textbf{AUC (↑)} & \textbf{\% change} & \textbf{Sig.} \\
        \midrule
CONCH & \textbf{0.807 {\scriptsize $\pm$ 0.054}} & - & - & 0.765 {\scriptsize $\pm$ 0.056} & - & - \\
+SEAL & 0.801 {\scriptsize $\pm$ 0.036} & -0.7\% & - & \textbf{0.843 {\scriptsize $\pm$ 0.048}} & +10.2\% & - \\
        \midrule
H0MINI & 0.822 {\scriptsize $\pm$ 0.016} & - & - & 0.842 {\scriptsize $\pm$ 0.037} & - & - \\
+SEAL & \textbf{0.835 {\scriptsize $\pm$ 0.041}} & +1.6\% & - & \textbf{0.846 {\scriptsize $\pm$ 0.045}} & +0.5\% & - \\
        \midrule
PHIKON2 & 0.776 {\scriptsize $\pm$ 0.054} & - & - & 0.805 {\scriptsize $\pm$ 0.046} & - & - \\
+SEAL & \textbf{0.779 {\scriptsize $\pm$ 0.051}} & +0.4\% & - & 0.805 {\scriptsize $\pm$ 0.042} & +0.0\% & - \\
        \midrule
UNIv2 & 0.806 {\scriptsize $\pm$ 0.044} & - & - & \textbf{0.848 {\scriptsize $\pm$ 0.037}} & - & - \\
+SEAL & \textbf{0.836 {\scriptsize $\pm$ 0.040}} & +3.7\% & ** & 0.841 {\scriptsize $\pm$ 0.040} & -0.8\% & - \\
        \midrule
VIRCHOW2 & 0.816 {\scriptsize $\pm$ 0.028} & - & - & 0.844 {\scriptsize $\pm$ 0.037} & - & - \\
+SEAL & \textbf{0.839 {\scriptsize $\pm$ 0.034}} & +2.8\% & - & \textbf{0.856 {\scriptsize $\pm$ 0.042}} & +1.4\% & * \\
        \midrule
        \midrule[\heavyrulewidth]
Average & 0.805 {\scriptsize $\pm$ 0.039} & - & - & 0.821 {\scriptsize $\pm$ 0.043} & - & - \\
+SEAL & \textbf{0.818 {\scriptsize $\pm$ 0.040}} & +1.6\% & - & \textbf{0.838 {\scriptsize $\pm$ 0.043}} & +2.1\% & - \\
        \bottomrule
    \end{tabular}}
\end{table}

\begin{table}[htbp]
    \centering
    \scriptsize
    \setlength{\abovecaptionskip}{3pt}
    \setlength{\belowcaptionskip}{-9pt}
    \renewcommand{\arraystretch}{0.9}
    \caption{Performance comparison between SEAL fine-tuned encoders and vision-only baselines on the \textbf{CPTAC BRCA} dataset, showing the \textbf{PIK3CA mutation classification} task (n=112). This is a slide-level task evaluated using \textbf{AUC} across 50 cross-validation folds. The best-performing model for each encoder backbone in \textbf{bold}.}
    \label{tab:cptac_brca_PIK3CA_mutation_results}
    \adjustbox{width=\textwidth,center}{
    \begin{tabular}{lrrrrrrr}
        \toprule
 & \multicolumn{3}{c}{\textbf{ABMIL}} & \multicolumn{3}{c}{\textbf{MeanMIL}} \\
\cmidrule(lr){2-4} \cmidrule(lr){5-7}
\textbf{Model} & \textbf{AUC (↑)} & \textbf{\% change} & \textbf{Sig.} & \textbf{AUC (↑)} & \textbf{\% change} & \textbf{Sig.} \\
        \midrule
CONCH & 0.545 {\scriptsize $\pm$ 0.116} & - & - & \textbf{0.526 {\scriptsize $\pm$ 0.131}} & - & - \\
+SEAL & \textbf{0.553 {\scriptsize $\pm$ 0.113}} & +1.5\% & - & 0.511 {\scriptsize $\pm$ 0.116} & -2.9\% & - \\
        \midrule
H0MINI & \textbf{0.598 {\scriptsize $\pm$ 0.123}} & - & - & 0.570 {\scriptsize $\pm$ 0.107} & - & - \\
+SEAL & 0.554 {\scriptsize $\pm$ 0.115} & -7.4\% & - & \textbf{0.574 {\scriptsize $\pm$ 0.108}} & +0.7\% & - \\
        \midrule
PHIKON2 & \textbf{0.555 {\scriptsize $\pm$ 0.114}} & - & - & \textbf{0.556 {\scriptsize $\pm$ 0.111}} & - & - \\
+SEAL & 0.554 {\scriptsize $\pm$ 0.116} & -0.2\% & - & 0.544 {\scriptsize $\pm$ 0.103} & -2.2\% & - \\
        \midrule
UNIv2 & \textbf{0.595 {\scriptsize $\pm$ 0.103}} & - & - & \textbf{0.544 {\scriptsize $\pm$ 0.120}} & - & - \\
+SEAL & 0.575 {\scriptsize $\pm$ 0.099} & -3.4\% & - & 0.524 {\scriptsize $\pm$ 0.115} & -3.7\% & - \\
        \midrule
VIRCHOW2 & 0.533 {\scriptsize $\pm$ 0.129} & - & - & \textbf{0.545 {\scriptsize $\pm$ 0.116}} & - & - \\
+SEAL & \textbf{0.544 {\scriptsize $\pm$ 0.133}} & +2.1\% & - & 0.518 {\scriptsize $\pm$ 0.117} & -5.0\% & - \\
        \midrule
        \midrule[\heavyrulewidth]
Average & \textbf{0.565 {\scriptsize $\pm$ 0.117}} & - & - & \textbf{0.548 {\scriptsize $\pm$ 0.117}} & - & - \\
+SEAL & 0.556 {\scriptsize $\pm$ 0.115} & -1.6\% & - & 0.534 {\scriptsize $\pm$ 0.112} & -2.6\% & - \\
        \bottomrule
    \end{tabular}}
\end{table}

\begin{table}[htbp]
    \centering
    \scriptsize
    \setlength{\abovecaptionskip}{3pt}
    \setlength{\belowcaptionskip}{-9pt}
    \renewcommand{\arraystretch}{0.9}
    \caption{Performance comparison between SEAL fine-tuned encoders and vision-only baselines on the \textbf{CPTAC BRCA} dataset, showing the \textbf{TP53 Mutation classification} task (n=112). This is a slide-level task evaluated using \textbf{AUC} across 50 cross-validation folds. The best-performing model for each encoder backbone in \textbf{bold}.}
    \label{tab:cptac_brca_TP53_mutation_results}
    \adjustbox{width=\textwidth,center}{
    \begin{tabular}{lrrrrrrr}
        \toprule
 & \multicolumn{3}{c}{\textbf{ABMIL}} & \multicolumn{3}{c}{\textbf{MeanMIL}} \\
\cmidrule(lr){2-4} \cmidrule(lr){5-7}
\textbf{Model} & \textbf{AUC (↑)} & \textbf{\% change} & \textbf{Sig.} & \textbf{AUC (↑)} & \textbf{\% change} & \textbf{Sig.} \\
        \midrule
CONCH & 0.792 {\scriptsize $\pm$ 0.103} & - & - & 0.794 {\scriptsize $\pm$ 0.096} & - & - \\
+SEAL & \textbf{0.824 {\scriptsize $\pm$ 0.093}} & +4.0\% & ** & \textbf{0.812 {\scriptsize $\pm$ 0.093}} & +2.3\% & ** \\
        \midrule
H0MINI & 0.801 {\scriptsize $\pm$ 0.096} & - & - & 0.775 {\scriptsize $\pm$ 0.088} & - & - \\
+SEAL & \textbf{0.814 {\scriptsize $\pm$ 0.085}} & +1.6\% & - & \textbf{0.781 {\scriptsize $\pm$ 0.098}} & +0.8\% & - \\
        \midrule
PHIKON2 & 0.834 {\scriptsize $\pm$ 0.079} & - & - & 0.822 {\scriptsize $\pm$ 0.091} & - & - \\
+SEAL & \textbf{0.849 {\scriptsize $\pm$ 0.074}} & +1.8\% & ** & \textbf{0.826 {\scriptsize $\pm$ 0.086}} & +0.5\% & * \\
        \midrule
UNIv2 & 0.801 {\scriptsize $\pm$ 0.093} & - & - & \textbf{0.787 {\scriptsize $\pm$ 0.088}} & - & - \\
+SEAL & \textbf{0.820 {\scriptsize $\pm$ 0.090}} & +2.4\% & ** & 0.779 {\scriptsize $\pm$ 0.092} & -1.0\% & - \\
        \midrule
VIRCHOW2 & 0.772 {\scriptsize $\pm$ 0.084} & - & - & 0.754 {\scriptsize $\pm$ 0.095} & - & - \\
+SEAL & \textbf{0.813 {\scriptsize $\pm$ 0.090}} & +5.3\% & - & \textbf{0.773 {\scriptsize $\pm$ 0.092}} & +2.5\% & * \\
        \midrule
        \midrule[\heavyrulewidth]
Average & 0.800 {\scriptsize $\pm$ 0.091} & - & - & 0.786 {\scriptsize $\pm$ 0.092} & - & - \\
+SEAL & \textbf{0.824 {\scriptsize $\pm$ 0.086}} & +3.0\% & - & \textbf{0.794 {\scriptsize $\pm$ 0.092}} & +1.0\% & - \\
        \bottomrule
    \end{tabular}}
\end{table}

\begin{table}[htbp]
    \centering
    \scriptsize
    \setlength{\abovecaptionskip}{3pt}
    \setlength{\belowcaptionskip}{-9pt}
    \renewcommand{\arraystretch}{0.9}
    \caption{Performance comparison between SEAL fine-tuned encoders and vision-only baselines on the \textbf{CPTAC CCRCC} dataset, showing the \textbf{BAP1 mutation classification} task (n=218). This is a slide-level task evaluated using \textbf{AUC} across 50 cross-validation folds. The best-performing model for each encoder backbone in \textbf{bold}.}
    \label{tab:cptac_ccrcc_BAP1_mutation_results}
    \adjustbox{width=\textwidth,center}{
    \begin{tabular}{lrrrrrrr}
        \toprule
 & \multicolumn{3}{c}{\textbf{ABMIL}} & \multicolumn{3}{c}{\textbf{MeanMIL}} \\
\cmidrule(lr){2-4} \cmidrule(lr){5-7}
\textbf{Model} & \textbf{AUC (↑)} & \textbf{\% change} & \textbf{Sig.} & \textbf{AUC (↑)} & \textbf{\% change} & \textbf{Sig.} \\
        \midrule
CONCH & 0.658 {\scriptsize $\pm$ 0.157} & - & - & 0.689 {\scriptsize $\pm$ 0.141} & - & - \\
+SEAL & \textbf{0.728 {\scriptsize $\pm$ 0.139}} & +10.6\% & - & \textbf{0.709 {\scriptsize $\pm$ 0.126}} & +2.9\% & - \\
        \midrule
H0MINI & \textbf{0.727 {\scriptsize $\pm$ 0.126}} & - & - & 0.631 {\scriptsize $\pm$ 0.139} & - & - \\
+SEAL & 0.707 {\scriptsize $\pm$ 0.145} & -2.8\% & - & \textbf{0.700 {\scriptsize $\pm$ 0.136}} & +10.9\% & - \\
        \midrule
PHIKON2 & \textbf{0.620 {\scriptsize $\pm$ 0.155}} & - & - & 0.642 {\scriptsize $\pm$ 0.152} & - & - \\
+SEAL & 0.613 {\scriptsize $\pm$ 0.163} & -1.1\% & - & \textbf{0.649 {\scriptsize $\pm$ 0.148}} & +1.1\% & - \\
        \midrule
UNIv2 & \textbf{0.693 {\scriptsize $\pm$ 0.150}} & - & - & \textbf{0.720 {\scriptsize $\pm$ 0.145}} & - & - \\
+SEAL & 0.685 {\scriptsize $\pm$ 0.172} & -1.2\% & - & 0.715 {\scriptsize $\pm$ 0.161} & -0.7\% & - \\
        \midrule
VIRCHOW2 & 0.712 {\scriptsize $\pm$ 0.135} & - & - & 0.614 {\scriptsize $\pm$ 0.142} & - & - \\
+SEAL & \textbf{0.738 {\scriptsize $\pm$ 0.132}} & +3.7\% & - & \textbf{0.705 {\scriptsize $\pm$ 0.104}} & +14.8\% & - \\
        \midrule
        \midrule[\heavyrulewidth]
Average & 0.682 {\scriptsize $\pm$ 0.145} & - & - & 0.659 {\scriptsize $\pm$ 0.144} & - & - \\
+SEAL & \textbf{0.694 {\scriptsize $\pm$ 0.150}} & +1.8\% & - & \textbf{0.696 {\scriptsize $\pm$ 0.135}} & +5.5\% & - \\
        \bottomrule
    \end{tabular}}
\end{table}

\begin{table}[htbp]
    \centering
    \scriptsize
    \setlength{\abovecaptionskip}{3pt}
    \setlength{\belowcaptionskip}{-9pt}
    \renewcommand{\arraystretch}{0.9}
    \caption{Performance comparison between SEAL fine-tuned encoders and vision-only baselines on the \textbf{CPTAC CCRCC} dataset, showing the \textbf{VHL mutation classification} task (n=218). This is a slide-level task evaluated using \textbf{AUC} across 50 cross-validation folds. The best-performing model for each encoder backbone in \textbf{bold}.}
    \label{tab:cptac_ccrcc_VHL_mutation_results}
    \adjustbox{width=\textwidth,center}{
    \begin{tabular}{lrrrrrrr}
        \toprule
 & \multicolumn{3}{c}{\textbf{ABMIL}} & \multicolumn{3}{c}{\textbf{MeanMIL}} \\
\cmidrule(lr){2-4} \cmidrule(lr){5-7}
\textbf{Model} & \textbf{AUC (↑)} & \textbf{\% change} & \textbf{Sig.} & \textbf{AUC (↑)} & \textbf{\% change} & \textbf{Sig.} \\
        \midrule
CONCH & \textbf{0.564 {\scriptsize $\pm$ 0.136}} & - & - & \textbf{0.606 {\scriptsize $\pm$ 0.115}} & - & - \\
+SEAL & 0.537 {\scriptsize $\pm$ 0.129} & -4.8\% & - & 0.585 {\scriptsize $\pm$ 0.119} & -3.5\% & - \\
        \midrule
H0MINI & 0.543 {\scriptsize $\pm$ 0.121} & - & - & 0.574 {\scriptsize $\pm$ 0.129} & - & - \\
+SEAL & \textbf{0.567 {\scriptsize $\pm$ 0.124}} & +4.4\% & - & \textbf{0.621 {\scriptsize $\pm$ 0.146}} & +8.2\% & - \\
        \midrule
PHIKON2 & \textbf{0.454 {\scriptsize $\pm$ 0.136}} & - & - & 0.511 {\scriptsize $\pm$ 0.135} & - & - \\
+SEAL & 0.443 {\scriptsize $\pm$ 0.143} & -2.4\% & - & \textbf{0.515 {\scriptsize $\pm$ 0.141}} & +0.8\% & - \\
        \midrule
UNIv2 & \textbf{0.538 {\scriptsize $\pm$ 0.128}} & - & - & 0.542 {\scriptsize $\pm$ 0.133} & - & - \\
+SEAL & 0.533 {\scriptsize $\pm$ 0.127} & -0.9\% & - & \textbf{0.564 {\scriptsize $\pm$ 0.131}} & +4.1\% & - \\
        \midrule
VIRCHOW2 & 0.580 {\scriptsize $\pm$ 0.135} & - & - & 0.588 {\scriptsize $\pm$ 0.136} & - & - \\
+SEAL & \textbf{0.607 {\scriptsize $\pm$ 0.129}} & +4.7\% & * & \textbf{0.612 {\scriptsize $\pm$ 0.133}} & +4.1\% & * \\
        \midrule
        \midrule[\heavyrulewidth]
Average & 0.536 {\scriptsize $\pm$ 0.131} & - & - & 0.564 {\scriptsize $\pm$ 0.130} & - & - \\
+SEAL & \textbf{0.537 {\scriptsize $\pm$ 0.130}} & +0.3\% & - & \textbf{0.579 {\scriptsize $\pm$ 0.134}} & +2.7\% & - \\
        \bottomrule
    \end{tabular}}
\end{table}

\begin{table}[htbp]
    \centering
    \scriptsize
    \setlength{\abovecaptionskip}{3pt}
    \setlength{\belowcaptionskip}{-9pt}
    \renewcommand{\arraystretch}{0.9}
    \caption{Performance comparison between SEAL fine-tuned encoders and vision-only baselines on the \textbf{CPTAC COAD} dataset, showing the \textbf{KRAS mutation classification} task (n=97). This is a slide-level task evaluated using \textbf{AUC} across 50 cross-validation folds. The best-performing model for each encoder backbone in \textbf{bold}.}
    \label{tab:cptac_coad_KRAS_mutation_results}
    \adjustbox{width=\textwidth,center}{
    \begin{tabular}{lrrrrrrr}
        \toprule
 & \multicolumn{3}{c}{\textbf{ABMIL}} & \multicolumn{3}{c}{\textbf{MeanMIL}} \\
\cmidrule(lr){2-4} \cmidrule(lr){5-7}
\textbf{Model} & \textbf{AUC (↑)} & \textbf{\% change} & \textbf{Sig.} & \textbf{AUC (↑)} & \textbf{\% change} & \textbf{Sig.} \\
        \midrule
CONCH & 0.598 {\scriptsize $\pm$ 0.110} & - & - & 0.678 {\scriptsize $\pm$ 0.108} & - & - \\
+SEAL & \textbf{0.666 {\scriptsize $\pm$ 0.111}} & +11.4\% & - & \textbf{0.685 {\scriptsize $\pm$ 0.105}} & +1.0\% & - \\
        \midrule
H0MINI & \textbf{0.669 {\scriptsize $\pm$ 0.108}} & - & - & 0.647 {\scriptsize $\pm$ 0.118} & - & - \\
+SEAL & 0.635 {\scriptsize $\pm$ 0.095} & -5.1\% & - & \textbf{0.666 {\scriptsize $\pm$ 0.113}} & +2.9\% & * \\
        \midrule
PHIKON2 & \textbf{0.696 {\scriptsize $\pm$ 0.109}} & - & - & 0.626 {\scriptsize $\pm$ 0.105} & - & - \\
+SEAL & 0.684 {\scriptsize $\pm$ 0.099} & -1.7\% & - & \textbf{0.634 {\scriptsize $\pm$ 0.106}} & +1.3\% & * \\
        \midrule
UNIv2 & \textbf{0.676 {\scriptsize $\pm$ 0.097}} & - & - & 0.671 {\scriptsize $\pm$ 0.101} & - & - \\
+SEAL & 0.675 {\scriptsize $\pm$ 0.126} & -0.1\% & - & \textbf{0.674 {\scriptsize $\pm$ 0.104}} & +0.4\% & - \\
        \midrule
VIRCHOW2 & 0.606 {\scriptsize $\pm$ 0.122} & - & - & 0.636 {\scriptsize $\pm$ 0.110} & - & - \\
+SEAL & \textbf{0.673 {\scriptsize $\pm$ 0.109}} & +11.1\% & - & \textbf{0.660 {\scriptsize $\pm$ 0.101}} & +3.8\% & - \\
        \midrule
        \midrule[\heavyrulewidth]
Average & 0.649 {\scriptsize $\pm$ 0.109} & - & - & 0.652 {\scriptsize $\pm$ 0.108} & - & - \\
+SEAL & \textbf{0.667 {\scriptsize $\pm$ 0.108}} & +2.7\% & - & \textbf{0.664 {\scriptsize $\pm$ 0.106}} & +1.9\% & - \\
        \bottomrule
    \end{tabular}}
\end{table}

\begin{table}[htbp]
    \centering
    \scriptsize
    \setlength{\abovecaptionskip}{3pt}
    \setlength{\belowcaptionskip}{-9pt}
    \renewcommand{\arraystretch}{0.9}
    \caption{Performance comparison between SEAL fine-tuned encoders and vision-only baselines on the \textbf{CPTAC COAD} dataset, showing the \textbf{TP53 Mutation classification} task (n=97). This is a slide-level task evaluated using \textbf{AUC} across 50 cross-validation folds. The best-performing model for each encoder backbone in \textbf{bold}.}
    \label{tab:cptac_coad_TP53_mutation_results}
    \adjustbox{width=\textwidth,center}{
    \begin{tabular}{lrrrrrrr}
        \toprule
 & \multicolumn{3}{c}{\textbf{ABMIL}} & \multicolumn{3}{c}{\textbf{MeanMIL}} \\
\cmidrule(lr){2-4} \cmidrule(lr){5-7}
\textbf{Model} & \textbf{AUC (↑)} & \textbf{\% change} & \textbf{Sig.} & \textbf{AUC (↑)} & \textbf{\% change} & \textbf{Sig.} \\
        \midrule
CONCH & \textbf{0.643 {\scriptsize $\pm$ 0.132}} & - & - & \textbf{0.654 {\scriptsize $\pm$ 0.134}} & - & - \\
+SEAL & 0.638 {\scriptsize $\pm$ 0.134} & -0.8\% & - & 0.595 {\scriptsize $\pm$ 0.135} & -9.0\% & - \\
        \midrule
H0MINI & 0.686 {\scriptsize $\pm$ 0.118} & - & - & 0.678 {\scriptsize $\pm$ 0.143} & - & - \\
+SEAL & \textbf{0.715 {\scriptsize $\pm$ 0.115}} & +4.2\% & * & \textbf{0.711 {\scriptsize $\pm$ 0.133}} & +4.9\% & ** \\
        \midrule
PHIKON2 & 0.718 {\scriptsize $\pm$ 0.105} & - & - & 0.665 {\scriptsize $\pm$ 0.135} & - & - \\
+SEAL & \textbf{0.738 {\scriptsize $\pm$ 0.100}} & +2.8\% & ** & \textbf{0.686 {\scriptsize $\pm$ 0.127}} & +3.2\% & - \\
        \midrule
UNIv2 & \textbf{0.720 {\scriptsize $\pm$ 0.122}} & - & - & 0.738 {\scriptsize $\pm$ 0.117} & - & - \\
+SEAL & 0.708 {\scriptsize $\pm$ 0.124} & -1.7\% & - & \textbf{0.741 {\scriptsize $\pm$ 0.122}} & +0.4\% & - \\
        \midrule
VIRCHOW2 & 0.664 {\scriptsize $\pm$ 0.136} & - & - & 0.679 {\scriptsize $\pm$ 0.115} & - & - \\
+SEAL & \textbf{0.717 {\scriptsize $\pm$ 0.127}} & +8.0\% & ** & \textbf{0.744 {\scriptsize $\pm$ 0.117}} & +9.6\% & - \\
        \midrule
        \midrule[\heavyrulewidth]
Average & 0.686 {\scriptsize $\pm$ 0.123} & - & - & 0.683 {\scriptsize $\pm$ 0.129} & - & - \\
+SEAL & \textbf{0.703 {\scriptsize $\pm$ 0.120}} & +2.5\% & - & \textbf{0.695 {\scriptsize $\pm$ 0.127}} & +1.8\% & - \\
        \bottomrule
    \end{tabular}}
\end{table}

\begin{table}[htbp]
    \centering
    \scriptsize
    \setlength{\abovecaptionskip}{3pt}
    \setlength{\belowcaptionskip}{-9pt}
    \renewcommand{\arraystretch}{0.9}
    \caption{Performance comparison between SEAL fine-tuned encoders and vision-only baselines on the \textbf{CPTAC GBM} dataset, showing the \textbf{EGFR mutation classification} task (n=243). This is a slide-level task evaluated using \textbf{AUC} across 50 cross-validation folds. The best-performing model for each encoder backbone in \textbf{bold}.}
    \label{tab:cptac_gbm_EGFR_mutation_results}
    \adjustbox{width=\textwidth,center}{
    \begin{tabular}{lrrrrrrr}
        \toprule
 & \multicolumn{3}{c}{\textbf{ABMIL}} & \multicolumn{3}{c}{\textbf{MeanMIL}} \\
\cmidrule(lr){2-4} \cmidrule(lr){5-7}
\textbf{Model} & \textbf{AUC (↑)} & \textbf{\% change} & \textbf{Sig.} & \textbf{AUC (↑)} & \textbf{\% change} & \textbf{Sig.} \\
        \midrule
CONCH & 0.633 {\scriptsize $\pm$ 0.136} & - & - & 0.606 {\scriptsize $\pm$ 0.093} & - & - \\
+SEAL & \textbf{0.651 {\scriptsize $\pm$ 0.127}} & +2.8\% & - & \textbf{0.623 {\scriptsize $\pm$ 0.103}} & +2.8\% & * \\
        \midrule
H0MINI & 0.579 {\scriptsize $\pm$ 0.122} & - & - & 0.549 {\scriptsize $\pm$ 0.121} & - & - \\
+SEAL & \textbf{0.617 {\scriptsize $\pm$ 0.119}} & +6.6\% & ** & \textbf{0.622 {\scriptsize $\pm$ 0.113}} & +13.3\% & - \\
        \midrule
PHIKON2 & 0.510 {\scriptsize $\pm$ 0.127} & - & - & 0.527 {\scriptsize $\pm$ 0.104} & - & - \\
+SEAL & \textbf{0.600 {\scriptsize $\pm$ 0.127}} & +17.6\% & - & \textbf{0.528 {\scriptsize $\pm$ 0.106}} & +0.2\% & - \\
        \midrule
UNIv2 & 0.557 {\scriptsize $\pm$ 0.132} & - & - & \textbf{0.558 {\scriptsize $\pm$ 0.113}} & - & - \\
+SEAL & \textbf{0.574 {\scriptsize $\pm$ 0.142}} & +3.1\% & - & 0.553 {\scriptsize $\pm$ 0.108} & -0.9\% & - \\
        \midrule
VIRCHOW2 & 0.615 {\scriptsize $\pm$ 0.124} & - & - & 0.572 {\scriptsize $\pm$ 0.114} & - & - \\
+SEAL & \textbf{0.625 {\scriptsize $\pm$ 0.128}} & +1.6\% & - & \textbf{0.604 {\scriptsize $\pm$ 0.110}} & +5.6\% & - \\
        \midrule
        \midrule[\heavyrulewidth]
Average & 0.579 {\scriptsize $\pm$ 0.128} & - & - & 0.562 {\scriptsize $\pm$ 0.109} & - & - \\
+SEAL & \textbf{0.613 {\scriptsize $\pm$ 0.129}} & +6.0\% & - & \textbf{0.586 {\scriptsize $\pm$ 0.108}} & +4.2\% & - \\
        \bottomrule
    \end{tabular}}
\end{table}

\begin{table}[htbp]
    \centering
    \scriptsize
    \setlength{\abovecaptionskip}{3pt}
    \setlength{\belowcaptionskip}{-9pt}
    \renewcommand{\arraystretch}{0.9}
    \caption{Performance comparison between SEAL fine-tuned encoders and vision-only baselines on the \textbf{CPTAC GBM} dataset, showing the \textbf{TP53 Mutation classification} task (n=243). This is a slide-level task evaluated using \textbf{AUC} across 50 cross-validation folds. The best-performing model for each encoder backbone in \textbf{bold}.}
    \label{tab:cptac_gbm_TP53_mutation_results}
    \adjustbox{width=\textwidth,center}{
    \begin{tabular}{lrrrrrrr}
        \toprule
 & \multicolumn{3}{c}{\textbf{ABMIL}} & \multicolumn{3}{c}{\textbf{MeanMIL}} \\
\cmidrule(lr){2-4} \cmidrule(lr){5-7}
\textbf{Model} & \textbf{AUC (↑)} & \textbf{\% change} & \textbf{Sig.} & \textbf{AUC (↑)} & \textbf{\% change} & \textbf{Sig.} \\
        \midrule
CONCH & \textbf{0.828 {\scriptsize $\pm$ 0.084}} & - & - & 0.705 {\scriptsize $\pm$ 0.119} & - & - \\
+SEAL & 0.812 {\scriptsize $\pm$ 0.107} & -1.9\% & - & \textbf{0.757 {\scriptsize $\pm$ 0.121}} & +7.4\% & - \\
        \midrule
H0MINI & \textbf{0.832 {\scriptsize $\pm$ 0.101}} & - & - & \textbf{0.803 {\scriptsize $\pm$ 0.100}} & - & - \\
+SEAL & 0.799 {\scriptsize $\pm$ 0.125} & -4.0\% & - & 0.785 {\scriptsize $\pm$ 0.103} & -2.2\% & - \\
        \midrule
PHIKON2 & 0.763 {\scriptsize $\pm$ 0.108} & - & - & 0.686 {\scriptsize $\pm$ 0.130} & - & - \\
+SEAL & \textbf{0.777 {\scriptsize $\pm$ 0.105}} & +1.8\% & * & \textbf{0.708 {\scriptsize $\pm$ 0.116}} & +3.2\% & - \\
        \midrule
UNIv2 & 0.763 {\scriptsize $\pm$ 0.112} & - & - & \textbf{0.808 {\scriptsize $\pm$ 0.093}} & - & - \\
+SEAL & \textbf{0.797 {\scriptsize $\pm$ 0.110}} & +4.5\% & ** & 0.793 {\scriptsize $\pm$ 0.094} & -1.9\% & - \\
        \midrule
VIRCHOW2 & 0.745 {\scriptsize $\pm$ 0.119} & - & - & 0.759 {\scriptsize $\pm$ 0.109} & - & - \\
+SEAL & \textbf{0.761 {\scriptsize $\pm$ 0.114}} & +2.1\% & - & \textbf{0.775 {\scriptsize $\pm$ 0.087}} & +2.1\% & - \\
        \midrule
        \midrule[\heavyrulewidth]
Average & 0.786 {\scriptsize $\pm$ 0.105} & - & - & 0.752 {\scriptsize $\pm$ 0.110} & - & - \\
+SEAL & \textbf{0.789 {\scriptsize $\pm$ 0.112}} & +0.4\% & - & \textbf{0.764 {\scriptsize $\pm$ 0.104}} & +1.5\% & - \\
        \bottomrule
    \end{tabular}}
\end{table}

\begin{table}[htbp]
    \centering
    \scriptsize
    \setlength{\abovecaptionskip}{3pt}
    \setlength{\belowcaptionskip}{-9pt}
    \renewcommand{\arraystretch}{0.9}
    \caption{Performance comparison between SEAL fine-tuned encoders and vision-only baselines on the \textbf{CPTAC LUAD} dataset, showing the \textbf{EGFR mutation classification} task (n=324). This is a slide-level task evaluated using \textbf{AUC} across 50 cross-validation folds. The best-performing model for each encoder backbone in \textbf{bold}.}
    \label{tab:cptac_luad_EGFR_mutation_results}
    \adjustbox{width=\textwidth,center}{
    \begin{tabular}{lrrrrrrr}
        \toprule
 & \multicolumn{3}{c}{\textbf{ABMIL}} & \multicolumn{3}{c}{\textbf{MeanMIL}} \\
\cmidrule(lr){2-4} \cmidrule(lr){5-7}
\textbf{Model} & \textbf{AUC (↑)} & \textbf{\% change} & \textbf{Sig.} & \textbf{AUC (↑)} & \textbf{\% change} & \textbf{Sig.} \\
        \midrule
CONCH & \textbf{0.769 {\scriptsize $\pm$ 0.091}} & - & - & \textbf{0.810 {\scriptsize $\pm$ 0.094}} & - & - \\
+SEAL & 0.732 {\scriptsize $\pm$ 0.121} & -4.8\% & - & 0.762 {\scriptsize $\pm$ 0.111} & -5.9\% & - \\
        \midrule
H0MINI & \textbf{0.797 {\scriptsize $\pm$ 0.094}} & - & - & \textbf{0.796 {\scriptsize $\pm$ 0.101}} & - & - \\
+SEAL & 0.793 {\scriptsize $\pm$ 0.106} & -0.5\% & - & 0.776 {\scriptsize $\pm$ 0.107} & -2.5\% & - \\
        \midrule
PHIKON2 & 0.728 {\scriptsize $\pm$ 0.100} & - & - & 0.712 {\scriptsize $\pm$ 0.096} & - & - \\
+SEAL & \textbf{0.745 {\scriptsize $\pm$ 0.092}} & +2.3\% & * & \textbf{0.727 {\scriptsize $\pm$ 0.093}} & +2.1\% & ** \\
        \midrule
UNIv2 & \textbf{0.830 {\scriptsize $\pm$ 0.089}} & - & - & 0.777 {\scriptsize $\pm$ 0.099} & - & - \\
+SEAL & 0.815 {\scriptsize $\pm$ 0.110} & -1.8\% & - & 0.777 {\scriptsize $\pm$ 0.096} & +0.0\% & - \\
        \midrule
VIRCHOW2 & \textbf{0.822 {\scriptsize $\pm$ 0.089}} & - & - & \textbf{0.819 {\scriptsize $\pm$ 0.097}} & - & - \\
+SEAL & 0.786 {\scriptsize $\pm$ 0.105} & -4.4\% & - & 0.802 {\scriptsize $\pm$ 0.099} & -2.1\% & - \\
        \midrule
        \midrule[\heavyrulewidth]
Average & \textbf{0.789 {\scriptsize $\pm$ 0.093}} & - & - & \textbf{0.783 {\scriptsize $\pm$ 0.097}} & - & - \\
+SEAL & 0.774 {\scriptsize $\pm$ 0.107} & -1.9\% & - & 0.769 {\scriptsize $\pm$ 0.101} & -1.8\% & - \\
        \bottomrule
    \end{tabular}}
\end{table}

\begin{table}[htbp]
    \centering
    \scriptsize
    \setlength{\abovecaptionskip}{3pt}
    \setlength{\belowcaptionskip}{-9pt}
    \renewcommand{\arraystretch}{0.9}
    \caption{Performance comparison between SEAL fine-tuned encoders and vision-only baselines on the \textbf{CPTAC LUAD} dataset, showing the \textbf{STK11 Mutation classification} task (n=324). This is a slide-level task evaluated using \textbf{AUC} across 50 cross-validation folds. The best-performing model for each encoder backbone in \textbf{bold}.}
    \label{tab:cptac_luad_STK11_mutation_results}
    \adjustbox{width=\textwidth,center}{
    \begin{tabular}{lrrrrrrr}
        \toprule
 & \multicolumn{3}{c}{\textbf{ABMIL}} & \multicolumn{3}{c}{\textbf{MeanMIL}} \\
\cmidrule(lr){2-4} \cmidrule(lr){5-7}
\textbf{Model} & \textbf{AUC (↑)} & \textbf{\% change} & \textbf{Sig.} & \textbf{AUC (↑)} & \textbf{\% change} & \textbf{Sig.} \\
        \midrule
CONCH & \textbf{0.866 {\scriptsize $\pm$ 0.093}} & - & - & \textbf{0.861 {\scriptsize $\pm$ 0.103}} & - & - \\
+SEAL & 0.827 {\scriptsize $\pm$ 0.110} & -4.5\% & - & 0.831 {\scriptsize $\pm$ 0.127} & -3.5\% & - \\
        \midrule
H0MINI & 0.886 {\scriptsize $\pm$ 0.072} & - & - & 0.869 {\scriptsize $\pm$ 0.076} & - & - \\
+SEAL & \textbf{0.901 {\scriptsize $\pm$ 0.060}} & +1.7\% & * & \textbf{0.877 {\scriptsize $\pm$ 0.080}} & +0.9\% & * \\
        \midrule
PHIKON2 & \textbf{0.884 {\scriptsize $\pm$ 0.070}} & - & - & \textbf{0.847 {\scriptsize $\pm$ 0.084}} & - & - \\
+SEAL & 0.876 {\scriptsize $\pm$ 0.071} & -0.9\% & - & 0.844 {\scriptsize $\pm$ 0.082} & -0.4\% & - \\
        \midrule
UNIv2 & \textbf{0.908 {\scriptsize $\pm$ 0.052}} & - & - & \textbf{0.873 {\scriptsize $\pm$ 0.072}} & - & - \\
+SEAL & 0.907 {\scriptsize $\pm$ 0.063} & -0.1\% & - & 0.866 {\scriptsize $\pm$ 0.079} & -0.8\% & - \\
        \midrule
VIRCHOW2 & 0.809 {\scriptsize $\pm$ 0.104} & - & - & 0.845 {\scriptsize $\pm$ 0.099} & - & - \\
+SEAL & \textbf{0.835 {\scriptsize $\pm$ 0.090}} & +3.2\% & * & \textbf{0.871 {\scriptsize $\pm$ 0.081}} & +3.1\% & - \\
        \midrule
        \midrule[\heavyrulewidth]
Average & \textbf{0.871 {\scriptsize $\pm$ 0.078}} & - & - & \textbf{0.859 {\scriptsize $\pm$ 0.087}} & - & - \\
+SEAL & 0.869 {\scriptsize $\pm$ 0.079} & -0.2\% & - & 0.858 {\scriptsize $\pm$ 0.090} & -0.1\% & - \\
        \bottomrule
    \end{tabular}}
\end{table}

\begin{table}[htbp]
    \centering
    \scriptsize
    \setlength{\abovecaptionskip}{3pt}
    \setlength{\belowcaptionskip}{-9pt}
    \renewcommand{\arraystretch}{0.9}
    \caption{Performance comparison between SEAL fine-tuned encoders and vision-only baselines on the \textbf{CPTAC LUAD} dataset, showing the \textbf{TP53 Mutation classification} task (n=324). This is a slide-level task evaluated using \textbf{AUC} across 50 cross-validation folds. The best-performing model for each encoder backbone in \textbf{bold}.}
    \label{tab:cptac_luad_TP53_mutation_results}
    \adjustbox{width=\textwidth,center}{
    \begin{tabular}{lrrrrrrr}
        \toprule
 & \multicolumn{3}{c}{\textbf{ABMIL}} & \multicolumn{3}{c}{\textbf{MeanMIL}} \\
\cmidrule(lr){2-4} \cmidrule(lr){5-7}
\textbf{Model} & \textbf{AUC (↑)} & \textbf{\% change} & \textbf{Sig.} & \textbf{AUC (↑)} & \textbf{\% change} & \textbf{Sig.} \\
        \midrule
CONCH & 0.681 {\scriptsize $\pm$ 0.115} & - & - & 0.705 {\scriptsize $\pm$ 0.095} & - & - \\
+SEAL & \textbf{0.698 {\scriptsize $\pm$ 0.114}} & +2.5\% & - & \textbf{0.736 {\scriptsize $\pm$ 0.102}} & +4.4\% & - \\
        \midrule
H0MINI & 0.736 {\scriptsize $\pm$ 0.099} & - & - & 0.741 {\scriptsize $\pm$ 0.078} & - & - \\
+SEAL & \textbf{0.752 {\scriptsize $\pm$ 0.097}} & +2.2\% & - & \textbf{0.761 {\scriptsize $\pm$ 0.086}} & +2.7\% & ** \\
        \midrule
PHIKON2 & \textbf{0.687 {\scriptsize $\pm$ 0.113}} & - & - & 0.683 {\scriptsize $\pm$ 0.117} & - & - \\
+SEAL & 0.670 {\scriptsize $\pm$ 0.113} & -2.5\% & - & \textbf{0.685 {\scriptsize $\pm$ 0.113}} & +0.3\% & - \\
        \midrule
UNIv2 & 0.751 {\scriptsize $\pm$ 0.102} & - & - & 0.735 {\scriptsize $\pm$ 0.102} & - & - \\
+SEAL & \textbf{0.753 {\scriptsize $\pm$ 0.113}} & +0.3\% & - & \textbf{0.753 {\scriptsize $\pm$ 0.093}} & +2.4\% & ** \\
        \midrule
VIRCHOW2 & 0.785 {\scriptsize $\pm$ 0.082} & - & - & 0.771 {\scriptsize $\pm$ 0.086} & - & - \\
+SEAL & \textbf{0.788 {\scriptsize $\pm$ 0.099}} & +0.4\% & - & \textbf{0.778 {\scriptsize $\pm$ 0.095}} & +0.9\% & - \\
        \midrule
        \midrule[\heavyrulewidth]
Average & 0.728 {\scriptsize $\pm$ 0.102} & - & - & 0.727 {\scriptsize $\pm$ 0.096} & - & - \\
+SEAL & \textbf{0.732 {\scriptsize $\pm$ 0.107}} & +0.6\% & - & \textbf{0.743 {\scriptsize $\pm$ 0.098}} & +2.1\% & - \\
        \bottomrule
    \end{tabular}}
\end{table}

\begin{table}[htbp]
    \centering
    \scriptsize
    \setlength{\abovecaptionskip}{3pt}
    \setlength{\belowcaptionskip}{-9pt}
    \renewcommand{\arraystretch}{0.9}
    \caption{Performance comparison between SEAL fine-tuned encoders and vision-only baselines on the \textbf{MUT-HET-RCC} dataset, showing the \textbf{PBRM1 mutation classification} task (n=1291). This is a slide-level task evaluated using \textbf{AUC} across 5 cross-validation folds. The best-performing model for each encoder backbone in \textbf{bold}.}
    \label{tab:mut-het-rcc_PBRM1_mutation_results}
    \adjustbox{width=\textwidth,center}{
    \begin{tabular}{lrrrrrrr}
        \toprule
 & \multicolumn{3}{c}{\textbf{ABMIL}} & \multicolumn{3}{c}{\textbf{MeanMIL}} \\
\cmidrule(lr){2-4} \cmidrule(lr){5-7}
\textbf{Model} & \textbf{AUC (↑)} & \textbf{\% change} & \textbf{Sig.} & \textbf{AUC (↑)} & \textbf{\% change} & \textbf{Sig.} \\
        \midrule
CONCH & 0.791 {\scriptsize $\pm$ 0.036} & - & - & 0.762 {\scriptsize $\pm$ 0.017} & - & - \\
+SEAL & \textbf{0.798 {\scriptsize $\pm$ 0.027}} & +0.9\% & - & \textbf{0.763 {\scriptsize $\pm$ 0.023}} & +0.1\% & - \\
        \midrule
H0MINI & 0.819 {\scriptsize $\pm$ 0.027} & - & - & 0.796 {\scriptsize $\pm$ 0.046} & - & - \\
+SEAL & \textbf{0.836 {\scriptsize $\pm$ 0.021}} & +2.1\% & - & \textbf{0.809 {\scriptsize $\pm$ 0.024}} & +1.6\% & - \\
        \midrule
PHIKON2 & 0.804 {\scriptsize $\pm$ 0.047} & - & - & 0.798 {\scriptsize $\pm$ 0.023} & - & - \\
+SEAL & \textbf{0.813 {\scriptsize $\pm$ 0.013}} & +1.1\% & - & \textbf{0.799 {\scriptsize $\pm$ 0.024}} & +0.1\% & - \\
        \midrule
UNIv2 & 0.840 {\scriptsize $\pm$ 0.035} & - & - & 0.820 {\scriptsize $\pm$ 0.029} & - & - \\
+SEAL & \textbf{0.847 {\scriptsize $\pm$ 0.065}} & +0.8\% & - & \textbf{0.823 {\scriptsize $\pm$ 0.033}} & +0.4\% & - \\
        \midrule
VIRCHOW2 & 0.822 {\scriptsize $\pm$ 0.040} & - & - & 0.794 {\scriptsize $\pm$ 0.031} & - & - \\
+SEAL & \textbf{0.829 {\scriptsize $\pm$ 0.027}} & +0.9\% & - & \textbf{0.806 {\scriptsize $\pm$ 0.032}} & +1.5\% & * \\
        \midrule
        \midrule[\heavyrulewidth]
Average & 0.815 {\scriptsize $\pm$ 0.037} & - & - & 0.794 {\scriptsize $\pm$ 0.029} & - & - \\
+SEAL & \textbf{0.825 {\scriptsize $\pm$ 0.031}} & +1.2\% & - & \textbf{0.800 {\scriptsize $\pm$ 0.027}} & +0.8\% & - \\
        \bottomrule
    \end{tabular}}
\end{table}

\begin{table}[htbp]
    \centering
    \scriptsize
    \setlength{\abovecaptionskip}{3pt}
    \setlength{\belowcaptionskip}{-9pt}
    \renewcommand{\arraystretch}{0.9}
    \caption{Performance comparison between SEAL fine-tuned encoders and vision-only baselines on the \textbf{MUT-HET-RCC} dataset, showing the \textbf{SETD2 mutation classification} task (n=1291). This is a slide-level task evaluated using \textbf{AUC} across 5 cross-validation folds. The best-performing model for each encoder backbone in \textbf{bold}.}
    \label{tab:mut-het-rcc_SETD2_mutation_results}
    \adjustbox{width=\textwidth,center}{
    \begin{tabular}{lrrrrrrr}
        \toprule
 & \multicolumn{3}{c}{\textbf{ABMIL}} & \multicolumn{3}{c}{\textbf{MeanMIL}} \\
\cmidrule(lr){2-4} \cmidrule(lr){5-7}
\textbf{Model} & \textbf{AUC (↑)} & \textbf{\% change} & \textbf{Sig.} & \textbf{AUC (↑)} & \textbf{\% change} & \textbf{Sig.} \\
        \midrule
CONCH & 0.697 {\scriptsize $\pm$ 0.028} & - & - & 0.712 {\scriptsize $\pm$ 0.027} & - & - \\
+SEAL & \textbf{0.724 {\scriptsize $\pm$ 0.019}} & +3.9\% & * & \textbf{0.721 {\scriptsize $\pm$ 0.019}} & +1.3\% & - \\
        \midrule
H0MINI & 0.690 {\scriptsize $\pm$ 0.026} & - & - & 0.691 {\scriptsize $\pm$ 0.035} & - & - \\
+SEAL & \textbf{0.742 {\scriptsize $\pm$ 0.027}} & +7.5\% & ** & \textbf{0.729 {\scriptsize $\pm$ 0.028}} & +5.5\% & * \\
        \midrule
PHIKON2 & 0.702 {\scriptsize $\pm$ 0.044} & - & - & 0.723 {\scriptsize $\pm$ 0.026} & - & - \\
+SEAL & \textbf{0.725 {\scriptsize $\pm$ 0.044}} & +3.3\% & - & \textbf{0.725 {\scriptsize $\pm$ 0.025}} & +0.3\% & - \\
        \midrule
UNIv2 & 0.713 {\scriptsize $\pm$ 0.034} & - & - & \textbf{0.711 {\scriptsize $\pm$ 0.030}} & - & - \\
+SEAL & \textbf{0.716 {\scriptsize $\pm$ 0.018}} & +0.4\% & - & 0.704 {\scriptsize $\pm$ 0.024} & -1.0\% & - \\
        \midrule
VIRCHOW2 & 0.718 {\scriptsize $\pm$ 0.016} & - & - & 0.700 {\scriptsize $\pm$ 0.021} & - & - \\
+SEAL & \textbf{0.750 {\scriptsize $\pm$ 0.014}} & +4.5\% & ** & \textbf{0.718 {\scriptsize $\pm$ 0.028}} & +2.6\% & - \\
        \midrule
        \midrule[\heavyrulewidth]
Average & 0.704 {\scriptsize $\pm$ 0.030} & - & - & 0.707 {\scriptsize $\pm$ 0.028} & - & - \\
+SEAL & \textbf{0.731 {\scriptsize $\pm$ 0.024}} & +3.9\% & - & \textbf{0.719 {\scriptsize $\pm$ 0.025}} & +1.7\% & - \\
        \bottomrule
    \end{tabular}}
\end{table}

\begin{table}[htbp]
    \centering
    \scriptsize
    \setlength{\abovecaptionskip}{3pt}
    \setlength{\belowcaptionskip}{-9pt}
    \renewcommand{\arraystretch}{0.9}
    \caption{Performance comparison between SEAL fine-tuned encoders and vision-only baselines on the \textbf{CPTAC COAD} dataset, showing the \textbf{Microsatellite Instability classification} task (n=97). This is a slide-level task evaluated using \textbf{AUC} across 50 cross-validation folds. The best-performing model for each encoder backbone in \textbf{bold}.}
    \label{tab:cptac_coad_MSI_H_results}
    \adjustbox{width=\textwidth,center}{
    \begin{tabular}{lrrrrrrr}
        \toprule
 & \multicolumn{3}{c}{\textbf{ABMIL}} & \multicolumn{3}{c}{\textbf{MeanMIL}} \\
\cmidrule(lr){2-4} \cmidrule(lr){5-7}
\textbf{Model} & \textbf{AUC (↑)} & \textbf{\% change} & \textbf{Sig.} & \textbf{AUC (↑)} & \textbf{\% change} & \textbf{Sig.} \\
        \midrule
CONCH & \textbf{0.805 {\scriptsize $\pm$ 0.109}} & - & - & \textbf{0.844 {\scriptsize $\pm$ 0.096}} & - & - \\
+SEAL & 0.798 {\scriptsize $\pm$ 0.125} & -0.9\% & - & 0.837 {\scriptsize $\pm$ 0.110} & -0.8\% & - \\
        \midrule
H0MINI & 0.877 {\scriptsize $\pm$ 0.081} & - & - & 0.905 {\scriptsize $\pm$ 0.079} & - & - \\
+SEAL & \textbf{0.886 {\scriptsize $\pm$ 0.079}} & +1.0\% & - & \textbf{0.908 {\scriptsize $\pm$ 0.085}} & +0.3\% & - \\
        \midrule
PHIKON2 & 0.808 {\scriptsize $\pm$ 0.098} & - & - & 0.805 {\scriptsize $\pm$ 0.122} & - & - \\
+SEAL & \textbf{0.814 {\scriptsize $\pm$ 0.099}} & +0.7\% & - & \textbf{0.818 {\scriptsize $\pm$ 0.119}} & +1.6\% & - \\
        \midrule
UNIv2 & \textbf{0.879 {\scriptsize $\pm$ 0.072}} & - & - & 0.883 {\scriptsize $\pm$ 0.097} & - & - \\
+SEAL & 0.877 {\scriptsize $\pm$ 0.082} & -0.2\% & - & 0.883 {\scriptsize $\pm$ 0.097} & +0.0\% & - \\
        \midrule
VIRCHOW2 & 0.831 {\scriptsize $\pm$ 0.135} & - & - & 0.858 {\scriptsize $\pm$ 0.092} & - & - \\
+SEAL & \textbf{0.894 {\scriptsize $\pm$ 0.072}} & +7.6\% & - & \textbf{0.901 {\scriptsize $\pm$ 0.066}} & +5.0\% & - \\
        \midrule
        \midrule[\heavyrulewidth]
Average & 0.840 {\scriptsize $\pm$ 0.099} & - & - & 0.859 {\scriptsize $\pm$ 0.097} & - & - \\
+SEAL & \textbf{0.854 {\scriptsize $\pm$ 0.091}} & +1.6\% & - & \textbf{0.869 {\scriptsize $\pm$ 0.095}} & +1.2\% & - \\
        \bottomrule
    \end{tabular}}
\end{table}

\begin{table}[htbp]
    \centering
    \scriptsize
    \setlength{\abovecaptionskip}{3pt}
    \setlength{\belowcaptionskip}{-9pt}
    \renewcommand{\arraystretch}{0.9}
    \caption{Performance comparison between SEAL fine-tuned encoders and vision-only baselines on the \textbf{BRCA neoadjuvant therapy response} dataset, showing the \textbf{subtype classification} task (n=166). This is a slide-level task evaluated using \textbf{AUC} across 50 cross-validation folds. The best-performing model for each encoder backbone in \textbf{bold}.}
    \label{tab:bc_therapy_grade_results}
    \adjustbox{width=\textwidth,center}{
    \begin{tabular}{lrrrrrrr}
        \toprule
 & \multicolumn{3}{c}{\textbf{ABMIL}} & \multicolumn{3}{c}{\textbf{MeanMIL}} \\
\cmidrule(lr){2-4} \cmidrule(lr){5-7}
\textbf{Model} & \textbf{AUC (↑)} & \textbf{\% change} & \textbf{Sig.} & \textbf{AUC (↑)} & \textbf{\% change} & \textbf{Sig.} \\
        \midrule
CONCH & 0.711 {\scriptsize $\pm$ 0.082} & - & - & 0.733 {\scriptsize $\pm$ 0.060} & - & - \\
+SEAL & \textbf{0.735 {\scriptsize $\pm$ 0.071}} & +3.4\% & ** & \textbf{0.743 {\scriptsize $\pm$ 0.062}} & +1.4\% & * \\
        \midrule
H0MINI & 0.737 {\scriptsize $\pm$ 0.085} & - & - & 0.718 {\scriptsize $\pm$ 0.061} & - & - \\
+SEAL & \textbf{0.757 {\scriptsize $\pm$ 0.078}} & +2.7\% & * & \textbf{0.727 {\scriptsize $\pm$ 0.062}} & +1.3\% & ** \\
        \midrule
PHIKON2 & 0.729 {\scriptsize $\pm$ 0.093} & - & - & 0.717 {\scriptsize $\pm$ 0.058} & - & - \\
+SEAL & \textbf{0.736 {\scriptsize $\pm$ 0.078}} & +1.0\% & - & \textbf{0.735 {\scriptsize $\pm$ 0.059}} & +2.5\% & - \\
        \midrule
UNIv2 & \textbf{0.770 {\scriptsize $\pm$ 0.066}} & - & - & \textbf{0.751 {\scriptsize $\pm$ 0.058}} & - & - \\
+SEAL & 0.768 {\scriptsize $\pm$ 0.078} & -0.3\% & - & 0.747 {\scriptsize $\pm$ 0.059} & -0.5\% & - \\
        \midrule
VIRCHOW2 & \textbf{0.749 {\scriptsize $\pm$ 0.074}} & - & - & 0.735 {\scriptsize $\pm$ 0.066} & - & - \\
+SEAL & 0.745 {\scriptsize $\pm$ 0.075} & -0.5\% & - & \textbf{0.742 {\scriptsize $\pm$ 0.063}} & +1.0\% & - \\
        \midrule
        \midrule[\heavyrulewidth]
Average & 0.739 {\scriptsize $\pm$ 0.080} & - & - & 0.731 {\scriptsize $\pm$ 0.061} & - & - \\
+SEAL & \textbf{0.748 {\scriptsize $\pm$ 0.076}} & +1.2\% & - & \textbf{0.739 {\scriptsize $\pm$ 0.061}} & +1.1\% & - \\
        \bottomrule
    \end{tabular}}
\end{table}

\begin{table}[htbp]
    \centering
    \scriptsize
    \setlength{\abovecaptionskip}{3pt}
    \setlength{\belowcaptionskip}{-9pt}
    \renewcommand{\arraystretch}{0.9}
    \caption{Performance comparison between SEAL fine-tuned encoders and vision-only baselines on the \textbf{NSCLC} dataset, showing the \textbf{PDL1 expression regression} task (n=217). This is a slide-level task evaluated using \textbf{KAPPA} across 5 cross-validation folds. The best-performing model for each encoder backbone in \textbf{bold}.}
    \label{tab:nsclc_PDL1_expression_results}
    \adjustbox{width=\textwidth,center}{
    \begin{tabular}{lrrrrrrr}
        \toprule
 & \multicolumn{3}{c}{\textbf{ABMIL}} & \multicolumn{3}{c}{\textbf{MeanMIL}} \\
\cmidrule(lr){2-4} \cmidrule(lr){5-7}
\textbf{Model} & \textbf{KAPPA (↑)} & \textbf{\% change} & \textbf{Sig.} & \textbf{KAPPA (↑)} & \textbf{\% change} & \textbf{Sig.} \\
        \midrule
CONCH & \textbf{0.823 {\scriptsize $\pm$ 0.051}} & - & - & 0.595 {\scriptsize $\pm$ 0.073} & - & - \\
+SEAL & 0.752 {\scriptsize $\pm$ 0.068} & -8.6\% & - & \textbf{0.662 {\scriptsize $\pm$ 0.089}} & +11.3\% & * \\
        \midrule
H0MINI & 0.773 {\scriptsize $\pm$ 0.068} & - & - & 0.616 {\scriptsize $\pm$ 0.147} & - & - \\
+SEAL & \textbf{0.802 {\scriptsize $\pm$ 0.063}} & +3.8\% & ** & \textbf{0.643 {\scriptsize $\pm$ 0.140}} & +4.4\% & - \\
        \midrule
PHIKON2 & \textbf{0.829 {\scriptsize $\pm$ 0.038}} & - & - & 0.617 {\scriptsize $\pm$ 0.059} & - & - \\
+SEAL & 0.815 {\scriptsize $\pm$ 0.058} & -1.7\% & - & \textbf{0.620 {\scriptsize $\pm$ 0.081}} & +0.5\% & - \\
        \midrule
UNIv2 & 0.865 {\scriptsize $\pm$ 0.042} & - & - & 0.578 {\scriptsize $\pm$ 0.098} & - & - \\
+SEAL & \textbf{0.869 {\scriptsize $\pm$ 0.058}} & +0.5\% & - & \textbf{0.622 {\scriptsize $\pm$ 0.092}} & +7.6\% & ** \\
        \midrule
VIRCHOW2 & \textbf{0.856 {\scriptsize $\pm$ 0.052}} & - & - & \textbf{0.701 {\scriptsize $\pm$ 0.124}} & - & - \\
+SEAL & 0.819 {\scriptsize $\pm$ 0.085} & -4.3\% & - & 0.625 {\scriptsize $\pm$ 0.087} & -10.8\% & - \\
        \midrule
        \midrule[\heavyrulewidth]
Average & \textbf{0.829 {\scriptsize $\pm$ 0.050}} & - & - & 0.621 {\scriptsize $\pm$ 0.100} & - & - \\
+SEAL & 0.811 {\scriptsize $\pm$ 0.066} & -2.1\% & - & \textbf{0.634 {\scriptsize $\pm$ 0.098}} & +2.1\% & - \\
        \bottomrule
    \end{tabular}}
\end{table}

\begin{table}[htbp]
    \centering
    \scriptsize
    \setlength{\abovecaptionskip}{3pt}
    \setlength{\belowcaptionskip}{-9pt}
    \renewcommand{\arraystretch}{0.9}
    \caption{Performance comparison between SEAL fine-tuned encoders and vision-only baselines on the \textbf{ovarian treatment response} dataset, showing the \textbf{treatment response classsification} task (n=85). This is a slide-level task evaluated using \textbf{AUC} across 10 cross-validation folds. The best-performing model for each encoder backbone in \textbf{bold}.}
    \label{tab:ovarian_response_results}
    \adjustbox{width=\textwidth,center}{
    \begin{tabular}{lrrrrrrr}
        \toprule
 & \multicolumn{3}{c}{\textbf{ABMIL}} & \multicolumn{3}{c}{\textbf{MeanMIL}} \\
\cmidrule(lr){2-4} \cmidrule(lr){5-7}
\textbf{Model} & \textbf{AUC (↑)} & \textbf{\% change} & \textbf{Sig.} & \textbf{AUC (↑)} & \textbf{\% change} & \textbf{Sig.} \\
        \midrule
CONCH & 0.700 {\scriptsize $\pm$ 0.245} & - & - & 0.667 {\scriptsize $\pm$ 0.183} & - & - \\
+SEAL & \textbf{0.717 {\scriptsize $\pm$ 0.252}} & +2.4\% & - & \textbf{0.783 {\scriptsize $\pm$ 0.183}} & +17.4\% & * \\
        \midrule
H0MINI & 0.367 {\scriptsize $\pm$ 0.194} & - & - & 0.500 {\scriptsize $\pm$ 0.211} & - & - \\
+SEAL & \textbf{0.450 {\scriptsize $\pm$ 0.269}} & +22.6\% & - & \textbf{0.667 {\scriptsize $\pm$ 0.197}} & +33.4\% & ** \\
        \midrule
PHIKON2 & 0.283 {\scriptsize $\pm$ 0.269} & - & - & 0.383 {\scriptsize $\pm$ 0.325} & - & - \\
+SEAL & \textbf{0.317 {\scriptsize $\pm$ 0.252}} & +12.0\% & - & \textbf{0.467 {\scriptsize $\pm$ 0.364}} & +21.9\% & - \\
        \midrule
UNIv2 & 0.433 {\scriptsize $\pm$ 0.226} & - & - & 0.650 {\scriptsize $\pm$ 0.273} & - & - \\
+SEAL & \textbf{0.617 {\scriptsize $\pm$ 0.211}} & +42.5\% & * & 0.650 {\scriptsize $\pm$ 0.157} & +0.0\% & - \\
        \midrule
VIRCHOW2 & 0.450 {\scriptsize $\pm$ 0.279} & - & - & 0.550 {\scriptsize $\pm$ 0.130} & - & - \\
+SEAL & \textbf{0.483 {\scriptsize $\pm$ 0.174}} & +7.3\% & - & \textbf{0.717 {\scriptsize $\pm$ 0.167}} & +30.4\% & ** \\
        \midrule
        \midrule[\heavyrulewidth]
Average & 0.447 {\scriptsize $\pm$ 0.243} & - & - & 0.550 {\scriptsize $\pm$ 0.224} & - & - \\
+SEAL & \textbf{0.517 {\scriptsize $\pm$ 0.232}} & +15.7\% & - & \textbf{0.657 {\scriptsize $\pm$ 0.214}} & +19.4\% & - \\
        \bottomrule
    \end{tabular}}
\end{table}

\begin{table}[htbp]
    \centering
    \scriptsize
    \setlength{\abovecaptionskip}{3pt}
    \setlength{\belowcaptionskip}{-9pt}
    \renewcommand{\arraystretch}{0.9}
    \caption{Performance comparison between SEAL fine-tuned encoders and vision-only baselines on the \textbf{SurGen} dataset, showing the \textbf{overall survival} task (n=387). This is a slide-level task evaluated using \textbf{CINDEX} across 5 cross-validation folds. The best-performing model for each encoder backbone in \textbf{bold}.}
    \label{tab:sr386__OS_results}
    \adjustbox{width=\textwidth,center}{
    \begin{tabular}{lrrrrrrr}
        \toprule
 & \multicolumn{3}{c}{\textbf{ABMIL}} & \multicolumn{3}{c}{\textbf{MeanMIL}} \\
\cmidrule(lr){2-4} \cmidrule(lr){5-7}
\textbf{Model} & \textbf{CINDEX (↑)} & \textbf{\% change} & \textbf{Sig.} & \textbf{CINDEX (↑)} & \textbf{\% change} & \textbf{Sig.} \\
        \midrule
CONCH & 0.622 {\scriptsize $\pm$ 0.074} & - & - & 0.651 {\scriptsize $\pm$ 0.046} & - & - \\
+SEAL & \textbf{0.678 {\scriptsize $\pm$ 0.051}} & +9.0\% & - & \textbf{0.660 {\scriptsize $\pm$ 0.068}} & +1.4\% & - \\
        \midrule
H0MINI & 0.577 {\scriptsize $\pm$ 0.083} & - & - & 0.639 {\scriptsize $\pm$ 0.085} & - & - \\
+SEAL & \textbf{0.609 {\scriptsize $\pm$ 0.040}} & +5.5\% & - & \textbf{0.659 {\scriptsize $\pm$ 0.078}} & +3.1\% & * \\
        \midrule
PHIKON2 & 0.630 {\scriptsize $\pm$ 0.065} & - & - & \textbf{0.640 {\scriptsize $\pm$ 0.053}} & - & - \\
+SEAL & \textbf{0.634 {\scriptsize $\pm$ 0.069}} & +0.6\% & - & 0.609 {\scriptsize $\pm$ 0.059} & -4.8\% & - \\
        \midrule
UNIv2 & 0.620 {\scriptsize $\pm$ 0.071} & - & - & 0.641 {\scriptsize $\pm$ 0.089} & - & - \\
+SEAL & \textbf{0.623 {\scriptsize $\pm$ 0.067}} & +0.5\% & - & \textbf{0.665 {\scriptsize $\pm$ 0.076}} & +3.7\% & - \\
        \midrule
VIRCHOW2 & 0.626 {\scriptsize $\pm$ 0.064} & - & - & 0.644 {\scriptsize $\pm$ 0.094} & - & - \\
+SEAL & \textbf{0.644 {\scriptsize $\pm$ 0.089}} & +2.9\% & - & \textbf{0.688 {\scriptsize $\pm$ 0.065}} & +6.8\% & * \\
        \midrule
        \midrule[\heavyrulewidth]
Average & 0.615 {\scriptsize $\pm$ 0.071} & - & - & 0.643 {\scriptsize $\pm$ 0.073} & - & - \\
+SEAL & \textbf{0.638 {\scriptsize $\pm$ 0.063}} & +3.7\% & - & \textbf{0.656 {\scriptsize $\pm$ 0.069}} & +2.1\% & - \\
        \bottomrule
    \end{tabular}}
\end{table}

\begin{table}[htbp]
    \centering
    \scriptsize
    \setlength{\abovecaptionskip}{3pt}
    \setlength{\belowcaptionskip}{-9pt}
    \renewcommand{\arraystretch}{0.9}
    \caption{Performance comparison between SEAL fine-tuned encoders and vision-only baselines on the \textbf{CPTAC BRCA} dataset, showing the \textbf{Epithelial-Mesenchymal Transition Pathway regression} task (n=112). This is a slide-level task evaluated using \textbf{PCC} across 10 cross-validation folds. The best-performing model for each encoder backbone in \textbf{bold}.}
    \label{tab:cptac_brca_HALLMARK_EPITHELIAL_MESENCHYMAL_TRANSITION_results}
    \adjustbox{width=\textwidth,center}{
    \begin{tabular}{lrrrrrrr}
        \toprule
 & \multicolumn{3}{c}{\textbf{ABMIL}} & \multicolumn{3}{c}{\textbf{MeanMIL}} \\
\cmidrule(lr){2-4} \cmidrule(lr){5-7}
\textbf{Model} & \textbf{PCC (↑)} & \textbf{\% change} & \textbf{Sig.} & \textbf{PCC (↑)} & \textbf{\% change} & \textbf{Sig.} \\
        \midrule
CONCH & \textbf{0.660 {\scriptsize $\pm$ 0.088}} & - & - & \textbf{-0.128 {\scriptsize $\pm$ 0.241}} & - & - \\
+SEAL & 0.578 {\scriptsize $\pm$ 0.130} & -12.4\% & - & -0.141 {\scriptsize $\pm$ 0.224} & +10.2\% & - \\
        \midrule
H0MINI & 0.648 {\scriptsize $\pm$ 0.135} & - & - & \textbf{0.025 {\scriptsize $\pm$ 0.309}} & - & - \\
+SEAL & \textbf{0.655 {\scriptsize $\pm$ 0.140}} & +1.1\% & - & 0.003 {\scriptsize $\pm$ 0.271} & -88.0\% & - \\
        \midrule
PHIKON2 & 0.357 {\scriptsize $\pm$ 0.258} & - & - & 0.052 {\scriptsize $\pm$ 0.208} & - & - \\
+SEAL & \textbf{0.542 {\scriptsize $\pm$ 0.198}} & +51.8\% & * & \textbf{0.073 {\scriptsize $\pm$ 0.220}} & +40.4\% & - \\
        \midrule
UNIv2 & 0.657 {\scriptsize $\pm$ 0.105} & - & - & -0.047 {\scriptsize $\pm$ 0.294} & - & - \\
+SEAL & \textbf{0.673 {\scriptsize $\pm$ 0.097}} & +2.4\% & - & \textbf{0.000 {\scriptsize $\pm$ 0.207}} & -100.0\% & - \\
        \midrule
VIRCHOW2 & 0.655 {\scriptsize $\pm$ 0.110} & - & - & \textbf{0.642 {\scriptsize $\pm$ 0.105}} & - & - \\
+SEAL & \textbf{0.723 {\scriptsize $\pm$ 0.067}} & +10.4\% & * & 0.485 {\scriptsize $\pm$ 0.151} & -24.5\% & - \\
        \midrule
        \midrule[\heavyrulewidth]
Average & 0.595 {\scriptsize $\pm$ 0.139} & - & - & \textbf{0.109 {\scriptsize $\pm$ 0.231}} & - & - \\
+SEAL & \textbf{0.634 {\scriptsize $\pm$ 0.126}} & +6.5\% & - & 0.084 {\scriptsize $\pm$ 0.215} & -22.8\% & - \\
        \bottomrule
    \end{tabular}}
\end{table}

\begin{table}[htbp]
    \centering
    \scriptsize
    \setlength{\abovecaptionskip}{3pt}
    \setlength{\belowcaptionskip}{-9pt}
    \renewcommand{\arraystretch}{0.9}
    \caption{Performance comparison between SEAL fine-tuned encoders and vision-only baselines on the \textbf{CPTAC BRCA} dataset, showing the \textbf{G2/M Checkpoint Pathway regression} task (n=112). This is a slide-level task evaluated using \textbf{PCC} across 10 cross-validation folds. The best-performing model for each encoder backbone in \textbf{bold}.}
    \label{tab:cptac_brca_HALLMARK_G2M_CHECKPOINT_results}
    \adjustbox{width=\textwidth,center}{
    \begin{tabular}{lrrrrrrr}
        \toprule
 & \multicolumn{3}{c}{\textbf{ABMIL}} & \multicolumn{3}{c}{\textbf{MeanMIL}} \\
\cmidrule(lr){2-4} \cmidrule(lr){5-7}
\textbf{Model} & \textbf{PCC (↑)} & \textbf{\% change} & \textbf{Sig.} & \textbf{PCC (↑)} & \textbf{\% change} & \textbf{Sig.} \\
        \midrule
CONCH & 0.580 {\scriptsize $\pm$ 0.143} & - & - & 0.087 {\scriptsize $\pm$ 0.317} & - & - \\
+SEAL & \textbf{0.668 {\scriptsize $\pm$ 0.105}} & +15.2\% & - & \textbf{0.218 {\scriptsize $\pm$ 0.311}} & +150.6\% & - \\
        \midrule
H0MINI & 0.606 {\scriptsize $\pm$ 0.118} & - & - & -0.094 {\scriptsize $\pm$ 0.233} & - & - \\
+SEAL & \textbf{0.649 {\scriptsize $\pm$ 0.093}} & +7.1\% & * & \textbf{-0.079 {\scriptsize $\pm$ 0.227}} & -16.0\% & - \\
        \midrule
PHIKON2 & 0.321 {\scriptsize $\pm$ 0.195} & - & - & \textbf{-0.027 {\scriptsize $\pm$ 0.348}} & - & - \\
+SEAL & \textbf{0.529 {\scriptsize $\pm$ 0.149}} & +64.8\% & - & -0.048 {\scriptsize $\pm$ 0.370} & +77.8\% & - \\
        \midrule
UNIv2 & 0.565 {\scriptsize $\pm$ 0.147} & - & - & \textbf{-0.020 {\scriptsize $\pm$ 0.285}} & - & - \\
+SEAL & \textbf{0.628 {\scriptsize $\pm$ 0.143}} & +11.2\% & ** & -0.030 {\scriptsize $\pm$ 0.290} & +50.0\% & - \\
        \midrule
VIRCHOW2 & 0.619 {\scriptsize $\pm$ 0.114} & - & - & \textbf{0.605 {\scriptsize $\pm$ 0.126}} & - & - \\
+SEAL & \textbf{0.678 {\scriptsize $\pm$ 0.081}} & +9.5\% & * & 0.601 {\scriptsize $\pm$ 0.140} & -0.7\% & - \\
        \midrule
        \midrule[\heavyrulewidth]
Average & 0.538 {\scriptsize $\pm$ 0.143} & - & - & 0.110 {\scriptsize $\pm$ 0.262} & - & - \\
+SEAL & \textbf{0.630 {\scriptsize $\pm$ 0.114}} & +17.1\% & - & \textbf{0.132 {\scriptsize $\pm$ 0.268}} & +20.1\% & - \\
        \bottomrule
    \end{tabular}}
\end{table}

\begin{table}[htbp]
    \centering
    \scriptsize
    \setlength{\abovecaptionskip}{3pt}
    \setlength{\belowcaptionskip}{-9pt}
    \renewcommand{\arraystretch}{0.9}
    \caption{Performance comparison between SEAL fine-tuned encoders and vision-only baselines on the \textbf{CPTAC BRCA} dataset, showing the \textbf{Upregulated genes in PI3K/AKT/mTOR signaling Pathway regression} task (n=112). This is a slide-level task evaluated using \textbf{PCC} across 10 cross-validation folds. The best-performing model for each encoder backbone in \textbf{bold}.}
    \label{tab:cptac_brca_HALLMARK_PI3K_AKT_MTOR_SIGNALING_results}
    \adjustbox{width=\textwidth,center}{
    \begin{tabular}{lrrrrrrr}
        \toprule
 & \multicolumn{3}{c}{\textbf{ABMIL}} & \multicolumn{3}{c}{\textbf{MeanMIL}} \\
\cmidrule(lr){2-4} \cmidrule(lr){5-7}
\textbf{Model} & \textbf{PCC (↑)} & \textbf{\% change} & \textbf{Sig.} & \textbf{PCC (↑)} & \textbf{\% change} & \textbf{Sig.} \\
        \midrule
CONCH & 0.332 {\scriptsize $\pm$ 0.220} & - & - & \textbf{0.019 {\scriptsize $\pm$ 0.280}} & - & - \\
+SEAL & \textbf{0.410 {\scriptsize $\pm$ 0.196}} & +23.5\% & * & -0.037 {\scriptsize $\pm$ 0.273} & -294.7\% & - \\
        \midrule
H0MINI & 0.279 {\scriptsize $\pm$ 0.210} & - & - & 0.018 {\scriptsize $\pm$ 0.195} & - & - \\
+SEAL & \textbf{0.348 {\scriptsize $\pm$ 0.229}} & +24.7\% & ** & \textbf{0.023 {\scriptsize $\pm$ 0.193}} & +27.8\% & - \\
        \midrule
PHIKON2 & \textbf{0.304 {\scriptsize $\pm$ 0.213}} & - & - & \textbf{0.009 {\scriptsize $\pm$ 0.297}} & - & - \\
+SEAL & 0.149 {\scriptsize $\pm$ 0.199} & -51.0\% & - & 0.004 {\scriptsize $\pm$ 0.285} & -55.6\% & - \\
        \midrule
UNIv2 & \textbf{0.416 {\scriptsize $\pm$ 0.105}} & - & - & 0.019 {\scriptsize $\pm$ 0.300} & - & - \\
+SEAL & 0.398 {\scriptsize $\pm$ 0.151} & -4.3\% & - & \textbf{0.059 {\scriptsize $\pm$ 0.342}} & +210.5\% & - \\
        \midrule
VIRCHOW2 & 0.392 {\scriptsize $\pm$ 0.145} & - & - & \textbf{0.355 {\scriptsize $\pm$ 0.131}} & - & - \\
+SEAL & \textbf{0.513 {\scriptsize $\pm$ 0.184}} & +30.9\% & ** & 0.332 {\scriptsize $\pm$ 0.232} & -6.5\% & - \\
        \midrule
        \midrule[\heavyrulewidth]
Average & 0.345 {\scriptsize $\pm$ 0.179} & - & - & \textbf{0.084 {\scriptsize $\pm$ 0.241}} & - & - \\
+SEAL & \textbf{0.364 {\scriptsize $\pm$ 0.192}} & +5.5\% & - & 0.076 {\scriptsize $\pm$ 0.265} & -9.3\% & - \\
        \bottomrule
    \end{tabular}}
\end{table}

\begin{table}[htbp]
    \centering
    \scriptsize
    \setlength{\abovecaptionskip}{3pt}
    \setlength{\belowcaptionskip}{-9pt}
    \renewcommand{\arraystretch}{0.9}
    \caption{Performance comparison between SEAL fine-tuned encoders and vision-only baselines on the \textbf{CPTAC COAD} dataset, showing the \textbf{Epithelial-Mesenchymal Transition Pathway regression} task (n=97). This is a slide-level task evaluated using \textbf{PCC} across 10 cross-validation folds. The best-performing model for each encoder backbone in \textbf{bold}.}
    \label{tab:cptac_coad_HALLMARK_EPITHELIAL_MESENCHYMAL_TRANSITION_results}
    \adjustbox{width=\textwidth,center}{
    \begin{tabular}{lrrrrrrr}
        \toprule
 & \multicolumn{3}{c}{\textbf{ABMIL}} & \multicolumn{3}{c}{\textbf{MeanMIL}} \\
\cmidrule(lr){2-4} \cmidrule(lr){5-7}
\textbf{Model} & \textbf{PCC (↑)} & \textbf{\% change} & \textbf{Sig.} & \textbf{PCC (↑)} & \textbf{\% change} & \textbf{Sig.} \\
        \midrule
CONCH & 0.740 {\scriptsize $\pm$ 0.081} & - & - & 0.731 {\scriptsize $\pm$ 0.100} & - & - \\
+SEAL & \textbf{0.778 {\scriptsize $\pm$ 0.065}} & +5.1\% & - & \textbf{0.752 {\scriptsize $\pm$ 0.054}} & +2.9\% & - \\
        \midrule
H0MINI & 0.725 {\scriptsize $\pm$ 0.112} & - & - & 0.767 {\scriptsize $\pm$ 0.087} & - & - \\
+SEAL & \textbf{0.750 {\scriptsize $\pm$ 0.099}} & +3.4\% & - & \textbf{0.778 {\scriptsize $\pm$ 0.091}} & +1.4\% & - \\
        \midrule
PHIKON2 & 0.293 {\scriptsize $\pm$ 0.241} & - & - & \textbf{-0.056 {\scriptsize $\pm$ 0.313}} & - & - \\
+SEAL & \textbf{0.488 {\scriptsize $\pm$ 0.213}} & +66.6\% & - & -0.070 {\scriptsize $\pm$ 0.327} & +25.0\% & - \\
        \midrule
UNIv2 & 0.767 {\scriptsize $\pm$ 0.072} & - & - & 0.751 {\scriptsize $\pm$ 0.076} & - & - \\
+SEAL & \textbf{0.771 {\scriptsize $\pm$ 0.072}} & +0.5\% & - & \textbf{0.768 {\scriptsize $\pm$ 0.085}} & +2.3\% & - \\
        \midrule
VIRCHOW2 & 0.734 {\scriptsize $\pm$ 0.108} & - & - & \textbf{0.765 {\scriptsize $\pm$ 0.096}} & - & - \\
+SEAL & \textbf{0.772 {\scriptsize $\pm$ 0.088}} & +5.2\% & - & 0.677 {\scriptsize $\pm$ 0.119} & -11.5\% & - \\
        \midrule
        \midrule[\heavyrulewidth]
Average & 0.652 {\scriptsize $\pm$ 0.123} & - & - & \textbf{0.592 {\scriptsize $\pm$ 0.134}} & - & - \\
+SEAL & \textbf{0.712 {\scriptsize $\pm$ 0.107}} & +9.2\% & - & 0.581 {\scriptsize $\pm$ 0.135} & -1.8\% & - \\
        \bottomrule
    \end{tabular}}
\end{table}

\begin{table}[htbp]
    \centering
    \scriptsize
    \setlength{\abovecaptionskip}{3pt}
    \setlength{\belowcaptionskip}{-9pt}
    \renewcommand{\arraystretch}{0.9}
    \caption{Performance comparison between SEAL fine-tuned encoders and vision-only baselines on the \textbf{CPTAC COAD} dataset, showing the \textbf{G2/M Checkpoint Pathway regression} task (n=97). This is a slide-level task evaluated using \textbf{PCC} across 10 cross-validation folds. The best-performing model for each encoder backbone in \textbf{bold}.}
    \label{tab:cptac_coad_HALLMARK_G2M_CHECKPOINT_results}
    \adjustbox{width=\textwidth,center}{
    \begin{tabular}{lrrrrrrr}
        \toprule
 & \multicolumn{3}{c}{\textbf{ABMIL}} & \multicolumn{3}{c}{\textbf{MeanMIL}} \\
\cmidrule(lr){2-4} \cmidrule(lr){5-7}
\textbf{Model} & \textbf{PCC (↑)} & \textbf{\% change} & \textbf{Sig.} & \textbf{PCC (↑)} & \textbf{\% change} & \textbf{Sig.} \\
        \midrule
CONCH & 0.201 {\scriptsize $\pm$ 0.147} & - & - & 0.169 {\scriptsize $\pm$ 0.200} & - & - \\
+SEAL & \textbf{0.301 {\scriptsize $\pm$ 0.206}} & +49.8\% & ** & \textbf{0.272 {\scriptsize $\pm$ 0.237}} & +60.9\% & * \\
        \midrule
H0MINI & 0.327 {\scriptsize $\pm$ 0.206} & - & - & 0.290 {\scriptsize $\pm$ 0.236} & - & - \\
+SEAL & \textbf{0.441 {\scriptsize $\pm$ 0.183}} & +34.9\% & ** & \textbf{0.316 {\scriptsize $\pm$ 0.261}} & +9.0\% & - \\
        \midrule
PHIKON2 & 0.045 {\scriptsize $\pm$ 0.183} & - & - & \textbf{-0.045 {\scriptsize $\pm$ 0.231}} & - & - \\
+SEAL & \textbf{0.076 {\scriptsize $\pm$ 0.153}} & +68.9\% & - & -0.061 {\scriptsize $\pm$ 0.208} & +35.6\% & - \\
        \midrule
UNIv2 & \textbf{0.339 {\scriptsize $\pm$ 0.155}} & - & - & 0.251 {\scriptsize $\pm$ 0.208} & - & - \\
+SEAL & 0.309 {\scriptsize $\pm$ 0.199} & -8.8\% & - & \textbf{0.309 {\scriptsize $\pm$ 0.212}} & +23.1\% & - \\
        \midrule
VIRCHOW2 & 0.436 {\scriptsize $\pm$ 0.154} & - & - & \textbf{0.423 {\scriptsize $\pm$ 0.178}} & - & - \\
+SEAL & \textbf{0.503 {\scriptsize $\pm$ 0.163}} & +15.4\% & * & 0.402 {\scriptsize $\pm$ 0.221} & -5.0\% & - \\
        \midrule
        \midrule[\heavyrulewidth]
Average & 0.270 {\scriptsize $\pm$ 0.169} & - & - & 0.218 {\scriptsize $\pm$ 0.211} & - & - \\
+SEAL & \textbf{0.326 {\scriptsize $\pm$ 0.181}} & +20.9\% & - & \textbf{0.248 {\scriptsize $\pm$ 0.228}} & +13.8\% & - \\
        \bottomrule
    \end{tabular}}
\end{table}

\begin{table}[htbp]
    \centering
    \scriptsize
    \setlength{\abovecaptionskip}{3pt}
    \setlength{\belowcaptionskip}{-9pt}
    \renewcommand{\arraystretch}{0.9}
    \caption{Performance comparison between SEAL fine-tuned encoders and vision-only baselines on the \textbf{CPTAC GBM} dataset, showing the \textbf{Epithelial-Mesenchymal Transition Pathway regression} task (n=243). This is a slide-level task evaluated using \textbf{PCC} across 10 cross-validation folds. The best-performing model for each encoder backbone in \textbf{bold}.}
    \label{tab:cptac_gbm_HALLMARK_EPITHELIAL_MESENCHYMAL_TRANSITION_results}
    \adjustbox{width=\textwidth,center}{
    \begin{tabular}{lrrrrrrr}
        \toprule
 & \multicolumn{3}{c}{\textbf{ABMIL}} & \multicolumn{3}{c}{\textbf{MeanMIL}} \\
\cmidrule(lr){2-4} \cmidrule(lr){5-7}
\textbf{Model} & \textbf{PCC (↑)} & \textbf{\% change} & \textbf{Sig.} & \textbf{PCC (↑)} & \textbf{\% change} & \textbf{Sig.} \\
        \midrule
CONCH & \textbf{0.505 {\scriptsize $\pm$ 0.146}} & - & - & \textbf{0.065 {\scriptsize $\pm$ 0.223}} & - & - \\
+SEAL & 0.485 {\scriptsize $\pm$ 0.181} & -4.0\% & - & 0.027 {\scriptsize $\pm$ 0.251} & -58.5\% & - \\
        \midrule
H0MINI & 0.480 {\scriptsize $\pm$ 0.247} & - & - & \textbf{0.525 {\scriptsize $\pm$ 0.177}} & - & - \\
+SEAL & \textbf{0.621 {\scriptsize $\pm$ 0.102}} & +29.4\% & * & 0.168 {\scriptsize $\pm$ 0.293} & -68.0\% & - \\
        \midrule
PHIKON2 & \textbf{0.481 {\scriptsize $\pm$ 0.198}} & - & - & \textbf{-0.080 {\scriptsize $\pm$ 0.252}} & - & - \\
+SEAL & 0.466 {\scriptsize $\pm$ 0.189} & -3.1\% & - & -0.097 {\scriptsize $\pm$ 0.277} & +21.3\% & - \\
        \midrule
UNIv2 & 0.453 {\scriptsize $\pm$ 0.151} & - & - & \textbf{-0.068 {\scriptsize $\pm$ 0.165}} & - & - \\
+SEAL & \textbf{0.552 {\scriptsize $\pm$ 0.153}} & +21.9\% & * & -0.092 {\scriptsize $\pm$ 0.234} & +35.3\% & - \\
        \midrule
VIRCHOW2 & 0.579 {\scriptsize $\pm$ 0.180} & - & - & 0.574 {\scriptsize $\pm$ 0.158} & - & - \\
+SEAL & \textbf{0.666 {\scriptsize $\pm$ 0.123}} & +15.0\% & * & \textbf{0.613 {\scriptsize $\pm$ 0.192}} & +6.8\% & - \\
        \midrule
        \midrule[\heavyrulewidth]
Average & 0.500 {\scriptsize $\pm$ 0.184} & - & - & \textbf{0.203 {\scriptsize $\pm$ 0.195}} & - & - \\
+SEAL & \textbf{0.558 {\scriptsize $\pm$ 0.150}} & +11.7\% & - & 0.124 {\scriptsize $\pm$ 0.249} & -39.1\% & - \\
        \bottomrule
    \end{tabular}}
\end{table}

\begin{table}[htbp]
    \centering
    \scriptsize
    \setlength{\abovecaptionskip}{3pt}
    \setlength{\belowcaptionskip}{-9pt}
    \renewcommand{\arraystretch}{0.9}
    \caption{Performance comparison between SEAL fine-tuned encoders and vision-only baselines on the \textbf{CPTAC GBM} dataset, showing the \textbf{G2/M Checkpoint Pathway regression} task (n=243). This is a slide-level task evaluated using \textbf{PCC} across 10 cross-validation folds. The best-performing model for each encoder backbone in \textbf{bold}.}
    \label{tab:cptac_gbm_HALLMARK_G2M_CHECKPOINT_results}
    \adjustbox{width=\textwidth,center}{
    \begin{tabular}{lrrrrrrr}
        \toprule
 & \multicolumn{3}{c}{\textbf{ABMIL}} & \multicolumn{3}{c}{\textbf{MeanMIL}} \\
\cmidrule(lr){2-4} \cmidrule(lr){5-7}
\textbf{Model} & \textbf{PCC (↑)} & \textbf{\% change} & \textbf{Sig.} & \textbf{PCC (↑)} & \textbf{\% change} & \textbf{Sig.} \\
        \midrule
CONCH & 0.454 {\scriptsize $\pm$ 0.158} & - & - & 0.034 {\scriptsize $\pm$ 0.267} & - & - \\
+SEAL & \textbf{0.504 {\scriptsize $\pm$ 0.213}} & +11.0\% & - & \textbf{0.061 {\scriptsize $\pm$ 0.234}} & +79.4\% & - \\
        \midrule
H0MINI & 0.473 {\scriptsize $\pm$ 0.196} & - & - & \textbf{0.436 {\scriptsize $\pm$ 0.158}} & - & - \\
+SEAL & \textbf{0.514 {\scriptsize $\pm$ 0.162}} & +8.7\% & - & -0.079 {\scriptsize $\pm$ 0.255} & -118.1\% & - \\
        \midrule
PHIKON2 & 0.156 {\scriptsize $\pm$ 0.241} & - & - & -0.117 {\scriptsize $\pm$ 0.249} & - & - \\
+SEAL & \textbf{0.304 {\scriptsize $\pm$ 0.195}} & +94.9\% & * & \textbf{-0.084 {\scriptsize $\pm$ 0.224}} & -28.2\% & - \\
        \midrule
UNIv2 & 0.381 {\scriptsize $\pm$ 0.174} & - & - & \textbf{0.054 {\scriptsize $\pm$ 0.235}} & - & - \\
+SEAL & \textbf{0.535 {\scriptsize $\pm$ 0.124}} & +40.4\% & - & 0.039 {\scriptsize $\pm$ 0.205} & -27.8\% & - \\
        \midrule
VIRCHOW2 & 0.454 {\scriptsize $\pm$ 0.149} & - & - & 0.356 {\scriptsize $\pm$ 0.201} & - & - \\
+SEAL & \textbf{0.500 {\scriptsize $\pm$ 0.119}} & +10.1\% & - & \textbf{0.441 {\scriptsize $\pm$ 0.170}} & +23.9\% & * \\
        \midrule
        \midrule[\heavyrulewidth]
Average & 0.384 {\scriptsize $\pm$ 0.184} & - & - & \textbf{0.153 {\scriptsize $\pm$ 0.222}} & - & - \\
+SEAL & \textbf{0.471 {\scriptsize $\pm$ 0.163}} & +22.9\% & - & 0.076 {\scriptsize $\pm$ 0.218} & -50.5\% & - \\
        \bottomrule
    \end{tabular}}
\end{table}

\begin{table}[htbp]
    \centering
    \scriptsize
    \setlength{\abovecaptionskip}{3pt}
    \setlength{\belowcaptionskip}{-9pt}
    \renewcommand{\arraystretch}{0.9}
    \caption{Performance comparison between SEAL fine-tuned encoders and vision-only baselines on the \textbf{CPTAC GBM} dataset, showing the \textbf{Upregulated genes in PI3K/AKT/mTOR signaling Pathway regression} task (n=243). This is a slide-level task evaluated using \textbf{PCC} across 10 cross-validation folds. The best-performing model for each encoder backbone in \textbf{bold}.}
    \label{tab:cptac_gbm_HALLMARK_PI3K_AKT_MTOR_SIGNALING_results}
    \adjustbox{width=\textwidth,center}{
    \begin{tabular}{lrrrrrrr}
        \toprule
 & \multicolumn{3}{c}{\textbf{ABMIL}} & \multicolumn{3}{c}{\textbf{MeanMIL}} \\
\cmidrule(lr){2-4} \cmidrule(lr){5-7}
\textbf{Model} & \textbf{PCC (↑)} & \textbf{\% change} & \textbf{Sig.} & \textbf{PCC (↑)} & \textbf{\% change} & \textbf{Sig.} \\
        \midrule
CONCH & 0.277 {\scriptsize $\pm$ 0.151} & - & - & \textbf{0.133 {\scriptsize $\pm$ 0.269}} & - & - \\
+SEAL & \textbf{0.493 {\scriptsize $\pm$ 0.115}} & +78.0\% & - & 0.101 {\scriptsize $\pm$ 0.212} & -24.1\% & - \\
        \midrule
H0MINI & 0.238 {\scriptsize $\pm$ 0.133} & - & - & 0.188 {\scriptsize $\pm$ 0.150} & - & - \\
+SEAL & \textbf{0.378 {\scriptsize $\pm$ 0.119}} & +58.8\% & - & \textbf{0.236 {\scriptsize $\pm$ 0.259}} & +25.5\% & - \\
        \midrule
PHIKON2 & 0.196 {\scriptsize $\pm$ 0.134} & - & - & \textbf{0.003 {\scriptsize $\pm$ 0.275}} & - & - \\
+SEAL & \textbf{0.197 {\scriptsize $\pm$ 0.131}} & +0.5\% & - & -0.029 {\scriptsize $\pm$ 0.289} & -1066.7\% & - \\
        \midrule
UNIv2 & 0.118 {\scriptsize $\pm$ 0.161} & - & - & \textbf{-0.004 {\scriptsize $\pm$ 0.258}} & - & - \\
+SEAL & \textbf{0.193 {\scriptsize $\pm$ 0.182}} & +63.6\% & * & -0.023 {\scriptsize $\pm$ 0.247} & +475.0\% & - \\
        \midrule
VIRCHOW2 & 0.346 {\scriptsize $\pm$ 0.153} & - & - & \textbf{0.335 {\scriptsize $\pm$ 0.198}} & - & - \\
+SEAL & \textbf{0.350 {\scriptsize $\pm$ 0.146}} & +1.2\% & - & 0.289 {\scriptsize $\pm$ 0.198} & -13.7\% & - \\
        \midrule
        \midrule[\heavyrulewidth]
Average & 0.235 {\scriptsize $\pm$ 0.146} & - & - & \textbf{0.131 {\scriptsize $\pm$ 0.230}} & - & - \\
+SEAL & \textbf{0.322 {\scriptsize $\pm$ 0.139}} & +37.1\% & - & 0.115 {\scriptsize $\pm$ 0.241} & -12.4\% & - \\
        \bottomrule
    \end{tabular}}
\end{table}

\begin{table}[htbp]
    \centering
    \scriptsize
    \setlength{\abovecaptionskip}{3pt}
    \setlength{\belowcaptionskip}{-9pt}
    \renewcommand{\arraystretch}{0.9}
    \caption{Performance comparison between SEAL fine-tuned encoders and vision-only baselines on the \textbf{CPTAC LUAD} dataset, showing the \textbf{Epithelial-Mesenchymal Transition Pathway regression} task (n=324). This is a slide-level task evaluated using \textbf{PCC} across 10 cross-validation folds. The best-performing model for each encoder backbone in \textbf{bold}.}
    \label{tab:cptac_luad_HALLMARK_EPITHELIAL_MESENCHYMAL_TRANSITION_results}
    \adjustbox{width=\textwidth,center}{
    \begin{tabular}{lrrrrrrr}
        \toprule
 & \multicolumn{3}{c}{\textbf{ABMIL}} & \multicolumn{3}{c}{\textbf{MeanMIL}} \\
\cmidrule(lr){2-4} \cmidrule(lr){5-7}
\textbf{Model} & \textbf{PCC (↑)} & \textbf{\% change} & \textbf{Sig.} & \textbf{PCC (↑)} & \textbf{\% change} & \textbf{Sig.} \\
        \midrule
CONCH & 0.435 {\scriptsize $\pm$ 0.121} & - & - & 0.328 {\scriptsize $\pm$ 0.124} & - & - \\
+SEAL & \textbf{0.577 {\scriptsize $\pm$ 0.121}} & +32.6\% & * & \textbf{0.452 {\scriptsize $\pm$ 0.126}} & +37.8\% & ** \\
        \midrule
H0MINI & 0.451 {\scriptsize $\pm$ 0.112} & - & - & 0.262 {\scriptsize $\pm$ 0.186} & - & - \\
+SEAL & \textbf{0.473 {\scriptsize $\pm$ 0.096}} & +4.9\% & - & \textbf{0.415 {\scriptsize $\pm$ 0.136}} & +58.4\% & * \\
        \midrule
PHIKON2 & \textbf{0.200 {\scriptsize $\pm$ 0.177}} & - & - & -0.052 {\scriptsize $\pm$ 0.215} & - & - \\
+SEAL & 0.188 {\scriptsize $\pm$ 0.176} & -6.0\% & - & \textbf{-0.018 {\scriptsize $\pm$ 0.230}} & -65.4\% & - \\
        \midrule
UNIv2 & 0.238 {\scriptsize $\pm$ 0.135} & - & - & 0.007 {\scriptsize $\pm$ 0.158} & - & - \\
+SEAL & \textbf{0.405 {\scriptsize $\pm$ 0.165}} & +70.2\% & ** & \textbf{0.329 {\scriptsize $\pm$ 0.180}} & +4600.0\% & - \\
        \midrule
VIRCHOW2 & 0.488 {\scriptsize $\pm$ 0.190} & - & - & \textbf{0.483 {\scriptsize $\pm$ 0.155}} & - & - \\
+SEAL & \textbf{0.509 {\scriptsize $\pm$ 0.152}} & +4.3\% & - & 0.363 {\scriptsize $\pm$ 0.151} & -24.8\% & - \\
        \midrule
        \midrule[\heavyrulewidth]
Average & 0.362 {\scriptsize $\pm$ 0.147} & - & - & 0.206 {\scriptsize $\pm$ 0.168} & - & - \\
+SEAL & \textbf{0.430 {\scriptsize $\pm$ 0.142}} & +18.8\% & - & \textbf{0.308 {\scriptsize $\pm$ 0.165}} & +49.9\% & - \\
        \bottomrule
    \end{tabular}}
\end{table}

\begin{table}[htbp]
    \centering
    \scriptsize
    \setlength{\abovecaptionskip}{3pt}
    \setlength{\belowcaptionskip}{-9pt}
    \renewcommand{\arraystretch}{0.9}
    \caption{Performance comparison between SEAL fine-tuned encoders and vision-only baselines on the \textbf{CPTAC LUAD} dataset, showing the \textbf{G2/M Checkpoint Pathway regression} task (n=324). This is a slide-level task evaluated using \textbf{PCC} across 10 cross-validation folds. The best-performing model for each encoder backbone in \textbf{bold}.}
    \label{tab:cptac_luad_HALLMARK_G2M_CHECKPOINT_results}
    \adjustbox{width=\textwidth,center}{
    \begin{tabular}{lrrrrrrr}
        \toprule
 & \multicolumn{3}{c}{\textbf{ABMIL}} & \multicolumn{3}{c}{\textbf{MeanMIL}} \\
\cmidrule(lr){2-4} \cmidrule(lr){5-7}
\textbf{Model} & \textbf{PCC (↑)} & \textbf{\% change} & \textbf{Sig.} & \textbf{PCC (↑)} & \textbf{\% change} & \textbf{Sig.} \\
        \midrule
CONCH & 0.736 {\scriptsize $\pm$ 0.067} & - & - & 0.660 {\scriptsize $\pm$ 0.143} & - & - \\
+SEAL & \textbf{0.742 {\scriptsize $\pm$ 0.064}} & +0.8\% & - & \textbf{0.715 {\scriptsize $\pm$ 0.118}} & +8.3\% & * \\
        \midrule
H0MINI & 0.670 {\scriptsize $\pm$ 0.134} & - & - & 0.497 {\scriptsize $\pm$ 0.168} & - & - \\
+SEAL & \textbf{0.773 {\scriptsize $\pm$ 0.056}} & +15.4\% & ** & \textbf{0.738 {\scriptsize $\pm$ 0.086}} & +48.5\% & - \\
        \midrule
PHIKON2 & 0.616 {\scriptsize $\pm$ 0.075} & - & - & -0.110 {\scriptsize $\pm$ 0.314} & - & - \\
+SEAL & \textbf{0.619 {\scriptsize $\pm$ 0.135}} & +0.5\% & - & \textbf{-0.081 {\scriptsize $\pm$ 0.359}} & -26.4\% & - \\
        \midrule
UNIv2 & 0.626 {\scriptsize $\pm$ 0.149} & - & - & 0.325 {\scriptsize $\pm$ 0.182} & - & - \\
+SEAL & \textbf{0.761 {\scriptsize $\pm$ 0.073}} & +21.6\% & ** & \textbf{0.680 {\scriptsize $\pm$ 0.139}} & +109.2\% & - \\
        \midrule
VIRCHOW2 & 0.721 {\scriptsize $\pm$ 0.109} & - & - & \textbf{0.725 {\scriptsize $\pm$ 0.088}} & - & - \\
+SEAL & \textbf{0.762 {\scriptsize $\pm$ 0.088}} & +5.7\% & * & 0.692 {\scriptsize $\pm$ 0.082} & -4.6\% & - \\
        \midrule
        \midrule[\heavyrulewidth]
Average & 0.674 {\scriptsize $\pm$ 0.107} & - & - & 0.419 {\scriptsize $\pm$ 0.179} & - & - \\
+SEAL & \textbf{0.731 {\scriptsize $\pm$ 0.083}} & +8.5\% & - & \textbf{0.549 {\scriptsize $\pm$ 0.157}} & +30.9\% & - \\
        \bottomrule
    \end{tabular}}
\end{table}

\begin{table}[htbp]
    \centering
    \scriptsize
    \setlength{\abovecaptionskip}{3pt}
    \setlength{\belowcaptionskip}{-9pt}
    \renewcommand{\arraystretch}{0.9}
    \caption{Performance comparison between SEAL fine-tuned encoders and vision-only baselines on the \textbf{CPTAC LUAD} dataset, showing the \textbf{Upregulated genes in PI3K/AKT/mTOR signaling Pathway regression} task (n=324). This is a slide-level task evaluated using \textbf{PCC} across 10 cross-validation folds. The best-performing model for each encoder backbone in \textbf{bold}.}
    \label{tab:cptac_luad_HALLMARK_PI3K_AKT_MTOR_SIGNALING_results}
    \adjustbox{width=\textwidth,center}{
    \begin{tabular}{lrrrrrrr}
        \toprule
 & \multicolumn{3}{c}{\textbf{ABMIL}} & \multicolumn{3}{c}{\textbf{MeanMIL}} \\
\cmidrule(lr){2-4} \cmidrule(lr){5-7}
\textbf{Model} & \textbf{PCC (↑)} & \textbf{\% change} & \textbf{Sig.} & \textbf{PCC (↑)} & \textbf{\% change} & \textbf{Sig.} \\
        \midrule
CONCH & 0.217 {\scriptsize $\pm$ 0.128} & - & - & 0.112 {\scriptsize $\pm$ 0.099} & - & - \\
+SEAL & \textbf{0.392 {\scriptsize $\pm$ 0.147}} & +80.6\% & - & \textbf{0.397 {\scriptsize $\pm$ 0.123}} & +254.5\% & - \\
        \midrule
H0MINI & 0.270 {\scriptsize $\pm$ 0.198} & - & - & 0.077 {\scriptsize $\pm$ 0.270} & - & - \\
+SEAL & \textbf{0.344 {\scriptsize $\pm$ 0.150}} & +27.4\% & * & \textbf{0.318 {\scriptsize $\pm$ 0.146}} & +313.0\% & ** \\
        \midrule
PHIKON2 & 0.105 {\scriptsize $\pm$ 0.139} & - & - & -0.115 {\scriptsize $\pm$ 0.266} & - & - \\
+SEAL & \textbf{0.199 {\scriptsize $\pm$ 0.145}} & +89.5\% & - & \textbf{-0.111 {\scriptsize $\pm$ 0.265}} & -3.5\% & - \\
        \midrule
UNIv2 & 0.241 {\scriptsize $\pm$ 0.174} & - & - & -0.110 {\scriptsize $\pm$ 0.214} & - & - \\
+SEAL & \textbf{0.330 {\scriptsize $\pm$ 0.125}} & +36.9\% & - & \textbf{0.212 {\scriptsize $\pm$ 0.099}} & -292.7\% & - \\
        \midrule
VIRCHOW2 & 0.353 {\scriptsize $\pm$ 0.141} & - & - & \textbf{0.417 {\scriptsize $\pm$ 0.144}} & - & - \\
+SEAL & \textbf{0.368 {\scriptsize $\pm$ 0.153}} & +4.2\% & - & 0.270 {\scriptsize $\pm$ 0.156} & -35.3\% & - \\
        \midrule
        \midrule[\heavyrulewidth]
Average & 0.237 {\scriptsize $\pm$ 0.156} & - & - & 0.076 {\scriptsize $\pm$ 0.199} & - & - \\
+SEAL & \textbf{0.327 {\scriptsize $\pm$ 0.144}} & +37.7\% & - & \textbf{0.217 {\scriptsize $\pm$ 0.158}} & +185.0\% & - \\
        \bottomrule
    \end{tabular}}
\end{table}

\begin{table}[htbp]
    \centering
    \scriptsize
    \setlength{\abovecaptionskip}{3pt}
    \setlength{\belowcaptionskip}{-9pt}
    \renewcommand{\arraystretch}{0.9}
    \caption{Performance comparison between SEAL fine-tuned encoders and vision-only baselines on the \textbf{CPTAC PDA} dataset, showing the \textbf{Epithelial-Mesenchymal Transition Pathway regression} task (n=242). This is a slide-level task evaluated using \textbf{PCC} across 10 cross-validation folds. The best-performing model for each encoder backbone in \textbf{bold}.}
    \label{tab:cptac_pda_HALLMARK_EPITHELIAL_MESENCHYMAL_TRANSITION_results}
    \adjustbox{width=\textwidth,center}{
    \begin{tabular}{lrrrrrrr}
        \toprule
 & \multicolumn{3}{c}{\textbf{ABMIL}} & \multicolumn{3}{c}{\textbf{MeanMIL}} \\
\cmidrule(lr){2-4} \cmidrule(lr){5-7}
\textbf{Model} & \textbf{PCC (↑)} & \textbf{\% change} & \textbf{Sig.} & \textbf{PCC (↑)} & \textbf{\% change} & \textbf{Sig.} \\
        \midrule
CONCH & 0.357 {\scriptsize $\pm$ 0.138} & - & - & 0.359 {\scriptsize $\pm$ 0.132} & - & - \\
+SEAL & \textbf{0.413 {\scriptsize $\pm$ 0.152}} & +15.7\% & - & \textbf{0.386 {\scriptsize $\pm$ 0.146}} & +7.5\% & - \\
        \midrule
H0MINI & 0.341 {\scriptsize $\pm$ 0.081} & - & - & 0.236 {\scriptsize $\pm$ 0.209} & - & - \\
+SEAL & \textbf{0.380 {\scriptsize $\pm$ 0.113}} & +11.4\% & - & \textbf{0.422 {\scriptsize $\pm$ 0.124}} & +78.8\% & * \\
        \midrule
PHIKON2 & 0.103 {\scriptsize $\pm$ 0.156} & - & - & \textbf{-0.039 {\scriptsize $\pm$ 0.221}} & - & - \\
+SEAL & \textbf{0.209 {\scriptsize $\pm$ 0.257}} & +102.9\% & - & -0.042 {\scriptsize $\pm$ 0.232} & +7.7\% & - \\
        \midrule
UNIv2 & 0.087 {\scriptsize $\pm$ 0.156} & - & - & -0.014 {\scriptsize $\pm$ 0.177} & - & - \\
+SEAL & \textbf{0.217 {\scriptsize $\pm$ 0.171}} & +149.4\% & * & \textbf{0.001 {\scriptsize $\pm$ 0.177}} & -107.1\% & - \\
        \midrule
VIRCHOW2 & 0.441 {\scriptsize $\pm$ 0.128} & - & - & 0.407 {\scriptsize $\pm$ 0.125} & - & - \\
+SEAL & \textbf{0.462 {\scriptsize $\pm$ 0.089}} & +4.8\% & - & \textbf{0.411 {\scriptsize $\pm$ 0.096}} & +1.0\% & - \\
        \midrule
        \midrule[\heavyrulewidth]
Average & 0.266 {\scriptsize $\pm$ 0.132} & - & - & 0.190 {\scriptsize $\pm$ 0.173} & - & - \\
+SEAL & \textbf{0.336 {\scriptsize $\pm$ 0.156}} & +26.5\% & - & \textbf{0.236 {\scriptsize $\pm$ 0.155}} & +24.1\% & - \\
        \bottomrule
    \end{tabular}}
\end{table}

\clearpage

\begin{table}[htbp]
    \centering
    \scriptsize
    \setlength{\abovecaptionskip}{3pt}
    \setlength{\belowcaptionskip}{-9pt}
    \renewcommand{\arraystretch}{0.9}
    \caption{Performance comparison between SEAL fine-tuned encoders and vision-only baselines on the \textbf{CPTAC PDA} dataset, showing the \textbf{G2/M Checkpoint Pathway regression} task (n=242). This is a slide-level task evaluated using \textbf{PCC} across 10 cross-validation folds. The best-performing model for each encoder backbone in \textbf{bold}.}
    \label{tab:cptac_pda_HALLMARK_G2M_CHECKPOINT_results}
    \adjustbox{width=\textwidth,center}{
    \begin{tabular}{lrrrrrrr}
        \toprule
 & \multicolumn{3}{c}{\textbf{ABMIL}} & \multicolumn{3}{c}{\textbf{MeanMIL}} \\
\cmidrule(lr){2-4} \cmidrule(lr){5-7}
\textbf{Model} & \textbf{PCC (↑)} & \textbf{\% change} & \textbf{Sig.} & \textbf{PCC (↑)} & \textbf{\% change} & \textbf{Sig.} \\
        \midrule
CONCH & 0.495 {\scriptsize $\pm$ 0.160} & - & - & 0.430 {\scriptsize $\pm$ 0.152} & - & - \\
+SEAL & \textbf{0.624 {\scriptsize $\pm$ 0.111}} & +26.1\% & ** & \textbf{0.648 {\scriptsize $\pm$ 0.112}} & +50.7\% & - \\
        \midrule
H0MINI & 0.387 {\scriptsize $\pm$ 0.238} & - & - & 0.078 {\scriptsize $\pm$ 0.141} & - & - \\
+SEAL & \textbf{0.521 {\scriptsize $\pm$ 0.193}} & +34.6\% & ** & \textbf{0.537 {\scriptsize $\pm$ 0.179}} & +588.5\% & - \\
        \midrule
PHIKON2 & 0.010 {\scriptsize $\pm$ 0.184} & - & - & -0.025 {\scriptsize $\pm$ 0.150} & - & - \\
+SEAL & \textbf{0.207 {\scriptsize $\pm$ 0.256}} & +1970.0\% & * & \textbf{0.012 {\scriptsize $\pm$ 0.213}} & -148.0\% & - \\
        \midrule
UNIv2 & 0.242 {\scriptsize $\pm$ 0.164} & - & - & 0.193 {\scriptsize $\pm$ 0.220} & - & - \\
+SEAL & \textbf{0.463 {\scriptsize $\pm$ 0.073}} & +91.3\% & ** & \textbf{0.324 {\scriptsize $\pm$ 0.208}} & +67.9\% & ** \\
        \midrule
VIRCHOW2 & 0.467 {\scriptsize $\pm$ 0.170} & - & - & 0.498 {\scriptsize $\pm$ 0.174} & - & - \\
+SEAL & \textbf{0.682 {\scriptsize $\pm$ 0.129}} & +46.0\% & - & \textbf{0.544 {\scriptsize $\pm$ 0.137}} & +9.2\% & - \\
        \midrule
        \midrule[\heavyrulewidth]
Average & 0.320 {\scriptsize $\pm$ 0.183} & - & - & 0.235 {\scriptsize $\pm$ 0.167} & - & - \\
+SEAL & \textbf{0.499 {\scriptsize $\pm$ 0.152}} & +56.0\% & - & \textbf{0.413 {\scriptsize $\pm$ 0.170}} & +75.9\% & - \\
        \bottomrule
    \end{tabular}}
\end{table}

\clearpage

\begin{table}[htbp]
    \centering
    \scriptsize
    \setlength{\abovecaptionskip}{3pt}
    \setlength{\belowcaptionskip}{-9pt}
    \renewcommand{\arraystretch}{0.9}
    \caption{Performance comparison between SEAL fine-tuned encoders and vision-only baselines on the \textbf{CPTAC PDA} dataset, showing the \textbf{Upregulated genes in PI3K/AKT/mTOR signaling Pathway regression} task (n=242). This is a slide-level task evaluated using \textbf{PCC} across 10 cross-validation folds. The best-performing model for each encoder backbone in \textbf{bold}.}
    \label{tab:cptac_pda_HALLMARK_PI3K_AKT_MTOR_SIGNALING_results}
    \adjustbox{width=\textwidth,center}{
    \begin{tabular}{lrrrrrrr}
        \toprule
 & \multicolumn{3}{c}{\textbf{ABMIL}} & \multicolumn{3}{c}{\textbf{MeanMIL}} \\
\cmidrule(lr){2-4} \cmidrule(lr){5-7}
\textbf{Model} & \textbf{PCC (↑)} & \textbf{\% change} & \textbf{Sig.} & \textbf{PCC (↑)} & \textbf{\% change} & \textbf{Sig.} \\
        \midrule
CONCH & 0.143 {\scriptsize $\pm$ 0.186} & - & - & 0.094 {\scriptsize $\pm$ 0.167} & - & - \\
+SEAL & \textbf{0.238 {\scriptsize $\pm$ 0.162}} & +66.4\% & - & \textbf{0.241 {\scriptsize $\pm$ 0.168}} & +156.4\% & * \\
        \midrule
H0MINI & 0.060 {\scriptsize $\pm$ 0.174} & - & - & 0.032 {\scriptsize $\pm$ 0.141} & - & - \\
+SEAL & \textbf{0.186 {\scriptsize $\pm$ 0.134}} & +210.0\% & * & \textbf{0.206 {\scriptsize $\pm$ 0.121}} & +543.7\% & ** \\
        \midrule
PHIKON2 & \textbf{0.060 {\scriptsize $\pm$ 0.192}} & - & - & \textbf{-0.017 {\scriptsize $\pm$ 0.222}} & - & - \\
+SEAL & -0.175 {\scriptsize $\pm$ 0.227} & -391.7\% & - & -0.051 {\scriptsize $\pm$ 0.226} & +200.0\% & - \\
        \midrule
UNIv2 & 0.161 {\scriptsize $\pm$ 0.233} & - & - & \textbf{0.211 {\scriptsize $\pm$ 0.167}} & - & - \\
+SEAL & \textbf{0.207 {\scriptsize $\pm$ 0.236}} & +28.6\% & - & 0.201 {\scriptsize $\pm$ 0.183} & -4.7\% & - \\
        \midrule
VIRCHOW2 & \textbf{0.188 {\scriptsize $\pm$ 0.087}} & - & - & 0.175 {\scriptsize $\pm$ 0.081} & - & - \\
+SEAL & 0.175 {\scriptsize $\pm$ 0.190} & -6.9\% & - & \textbf{0.250 {\scriptsize $\pm$ 0.159}} & +42.9\% & - \\
        \midrule
        \midrule[\heavyrulewidth]
Average & 0.122 {\scriptsize $\pm$ 0.174} & - & - & 0.099 {\scriptsize $\pm$ 0.156} & - & - \\
+SEAL & \textbf{0.126 {\scriptsize $\pm$ 0.190}} & +3.1\% & - & \textbf{0.169 {\scriptsize $\pm$ 0.171}} & +71.1\% & - \\
        \bottomrule
    \end{tabular}}
\end{table}